\documentclass[draft]{agujournal2019}
\usepackage{url} %this package should fix any errors with URLs in refs.
\usepackage{lineno}
\usepackage[inline]{trackchanges} %for better track changes. finalnew option will compile document with changes incorporated.
\usepackage{soul}
\linenumbers

\usepackage{amsmath} % include the amstext, amsbsy, and amsopn packages
\usepackage{amssymb,stmaryrd} % extra symbols
\usepackage{amsfonts}% extra fonts and symbols, including boldface 
\usepackage{amscd}   % CD environment for simple commutative diagrams 
\usepackage{array} 
\usepackage{algorithm}
\usepackage{algcompatible}
\usepackage{subfigure}
\nolinenumbers
\usepackage{graphicx}
\usepackage{xcolor}

\usepackage{xr}

\makeatletter
\newcommand*{\addFileDependency}[1]{% argument=file name and extension
  \typeout{(#1)}
  \@addtofilelist{#1}
  \IfFileExists{#1}{}{\typeout{No file #1.}}
}
\makeatother

\newcommand*{\myexternaldocument}[1]{%
    \externaldocument{#1}%
    \addFileDependency{#1.tex}%
    \addFileDependency{#1.aux}%
}

\myexternaldocument{Supplement}

\draftfalse
\journalname{JGR: Solid Earth}

\begin{document}

\title{SeismoGen: Seismic Waveform Synthesis Using Generative Adversarial Networks}

\authors{Tiantong Wang\affil{1, 2},
Daniel Trugman\affil{3}, and
Youzuo Lin\affil{1}}

\affiliation{1}{Geophysics Group, Earth and Environment Science Division, Los Alamos National Laboratory, \\ Los Alamos, NM 87545, USA.}
\affiliation{2}{School of Information Sciences, University of Pittsburgh,\\ Pittsburgh, PA 15260, USA.}
\affiliation{3}{Department of Geological Sciences, Jackson School of Geosciences, The University of Texas at Austin,\\ Austin, TX 78712, USA. }
%(repeat as many times as is necessary)

\correspondingauthor{Youzuo Lin}{ylin@lanl.gov}

% OLD: TOO LONG
% \begin{keypoints}
% \item We develop a generative adversarial neural network model to generate synthetic 3-component seismic waveforms of both noise and event classes. 
% \item We validate the quality of the synthetic waveforms qualitatively and quantitatively through application in training a machine-learning based earthquake classifier.  
% \item We further demonstrate that these high-quality synthetic waveforms can be combined with real seismic data to enhance the performance of the existing machine-learning based earthquake detection methods.
% \end{keypoints}

% NEW: 140 characters (with spaces, no abbreviations)
\begin{keypoints}
\item We develop a generative adversarial neural network model to generate synthetic 3-component waveforms of both event and noise classes.
\item We validate the synthetic waveforms both visually and quantitatively through use of a machine-learning based earthquake classifier. 
\item We demonstrate that our synthetic waveforms can augment real seismic data to improve machine learning-based earthquake detection methods. 
\end{keypoints}

\begin{abstract}
Detecting earthquake arrivals within seismic time series can be a challenging task. Visual, human detection has long been considered the gold standard but requires intensive manual labor that scales poorly to large data sets. In recent years, automatic detection methods based on machine learning have been developed to improve the accuracy and efficiency. However, accuracy of those methods rely on access to a sufficient amount of high-quality labeled training data, often tens of thousands of records or more. This paper aims to resolve this dilemma by answering two questions: (1) Provided with a limited amount of reliable labeled data, can we use them to generate additional, realistic synthetic data? and (2) Can we use those synthetic datasets to further hone our detection algorithms? To address these questions, we use a generative adversarial network (GAN), a type of machine learning model which has shown supreme capability in generating high-quality synthetic samples in multiple domains. Once trained, our GAN model is capable of producing realistic seismic waveforms of both noise and event classes. Applied to real-Earth seismic datasets in Oklahoma, we show that data augmentation from our GAN-generated synthetic waveforms can be used to improve earthquake detection algorithms in instances when only small amounts of labeled training data are available.\end{abstract}

% \section*{Plain Language Summary}
% [ enter your Plain Language Summary here or delete this section]

%\textcolor{red}{test color}

\section{Introduction}
\label{sec:introduction}

Detection of earthquake events within seismic time series records plays a fundamental role in seismology. However, such a task can in practice be challenging. Seismic waveforms have unique characteristics compared to time series from other physics domains, and require intensive training and domain knowledge to manually recognize and characterize them. Automated seismic detection methods have been deployed for decades, with the most popular methods including short-time-average/long-time-average \cite{allen1978automatic} and waveform correlation approaches \cite{Gibbons-2006-Gibbons}. However, these more conventional detection methods may sometimes generate too false positives, can fail in situations with low signal-to-noise ratio, and often suffer from expensive computational costs \cite{Earthquake-2015-Yoon}.

In recent years, with the volume of seismic data increasing significantly, automatic and efficient earthquake detection methods are needed.  Machine learning methods using deep neural network~(DNN) architectures have been successful in object detection to identify patterns. Of these, convolutional neural networks~(CNN)  have achieved promising results in computer vision, image analysis, and many other domains due to the significantly improved computational power. In 2012, AlexNet won the ImageNet competition~\cite{krizhevsky2012imagenet}, with a design incorporating fully connected layers and max-pooling layers to outperform other methods. After that, a sequence of different structures such as VGGNet~\cite{VGG}, ResNet~\cite{he2016deep}, GoogleNet~\cite{szegedy2017inception}, and DenseNet~\cite{huang2017densely} were introduced.  

Meanwhile, researchers in seismology have also started developing CNN-based earthquake detection methods. \citeA{Convolutional-2018-Perol} introduced a CNN network architecture (``ConvNetQuake'') to study the induced seismicity in Oklahoma. \citeA{ross_generalized_2018} leveraged the vast labeled datasets of the Southern California Seismic Network archive to develop the Generalized Phase Detection algorithm using CNNs, while \citeA{zhu_phasenet_2019} developed a similar approach called PhaseNet using datasets from northern California. Taking advantage of the temporal structure of seismic waveforms, \citeA{mousavi_cred_2019} used a hybrid convolutional and reccurrent neural network architecture in devising the CRED algorithm. Several other studies have built on and modified these approaches, applying them to various problems across seismology \cite{dokht_seismic_2019,kriegerowski_deep_2019,tibi_classification_2019,linville_deep_2019,lomax_investigation_2019,meier_reliable_2019}, see \citeA{bergen_machine_2019} and \citeA{kong_machine_2019} for recent reviews. 

In this article, we advance the ``DeepDetect'' detection method \cite{DeepDetect-2019-Wu}, which is a cascaded region-based convolutional neural network designed to capture earthquake events in different sizes while incorporating contextual information to enrich features for each proposal, and the work of  \citeA{zhang2019adaptive}, which implemented a deep learning based earthquake/non-earthquake classification model with an adaptive threshold frequency filtering module to achieve superior performance. 

All of the  aforementioned neural networks are supervised, meaning that they all require an iterative training procedure to learn the characteristic patterns of seismic waveforms from labeled datasets.  Training these models requires a sufficient amount of labeled data, tens of thousands or even millions of records in many cases \cite{ross_generalized_2018}. However, in many earthquake detection problems, labeled data at this scale is simply not available, and would require thousands of hours of human labor from trained seismic analysts to produce. In order to resolve this dilemma, we develop a generative model to synthesize realistic, labeled waveform data, and use them to augment real world training data.

We use a generative adversarial network (GAN), which is a type of generative model based on an adversarial min-max game between two networks, generator and discriminator~\cite{GAN}. The role of the generator is to synthesize realistic data by sampling from a simple distribution like Gaussian and learning to map to the data domain using a neural network as a universal function approximator. The discriminator, in contrast, is trained to distinguish this type of synthetic data from real data samples. This is achieved by adversarial training of these two networks. Researchers have successfully applied GAN to image synthesis \cite{GAN, creswell2018generative}, audio waveform generation~\cite{engel2019gansynth, yang2017midinet, chen2017deep}, and speech synthesis~\cite{pascual2017segan, saito2017statistical, kaneko2017generative}. In seismology, researchers have also applied GAN to several existing problems. In \citeA{Li-2018-Machine}, GAN is first used to extract a compact and effective representation of seismic waveforms. Once fully trained, a random forest classifier is built on the discriminator to distinguish between earthquake events and noise. In their work, only single component of the waveform data is considered. GAN has also been proved to be effective in other geophysical applications such as inversion~\cite{Zhang-2020-Data,Zhong-2020-Inversion}, processing~\cite{Picetti-2019-Seismic}, and interpretation~\cite{Lu-2018-generative}. 

There are multiple variants of GAN, the most important for this article being the conditional GAN \cite{CondGAN}, which turns the traditional GAN into a conditional model, which allows the user to customize the category of the generated samples by  an additional label information as input. In this paper, we developed a generative model based on conditional GAN that can produce synthetic seismic time series. While GAN models have been used previously in data augmentation tasks \cite{perez2017effectiveness}, to our knowledge GAN generated synthetic data has not been applied to data augmentation problems for 1D time series or seismic event detection tasks. We validate the quality of synthetic seismic events visually and quantitatively.  With the promise of our high-quality synthetic seismic samples, we further explore the feasibility of augmenting limited data sets with our synthetic samples on a earthquake detection problem in Oklahoma. 

The layout of this article is as follows. In Section~\ref{section: model}, we describe the fundamentals of GAN models and their variants. In Section~\ref{section:data}, we provide details on the field data and preprocessing techniques.  We then develop and discuss our model in Section~\ref{sec:model verification}. Section~\ref{sec:experiment} describes experimental results. Finally, in Sections 6 and 7, we discuss model limitations, future work, and present concluding remarks.

\section{Theory}
\label{section: model}

\subsection{Generative Adversarial Networks}

Generative adversarial networks (GAN) are a family of deep-learning-based generative models that can be used to learn a distribution and produce realistic synthetic samples. A typical GAN consists of two feed-forward neural networks: a generator and a discriminator. The generator learns a function that maps a prior vector to a realistic synthetic sample, while the discriminator reads in both real and synthetic samples and learns to distinguish between them. Training a GAN model can be usually expressed in terms of the optimization of a value function of the form:
\begin{equation} \label{eq:gan loss}
\underset{G}{\mathrm{min}}\,\underset{D}{\mathrm{max}}\,~V(D, G, x, z) = \mathbb{E}_{x \sim p_{\mathrm{data}} }[\log{(D(x))}] + \mathbb{E}_{z \sim p_{z}}[\log{(1-D(G(z)))}],
\end{equation}
where $G(\cdot)$ is the generator and $D(\cdot)$ is the discriminator. The random vector of $z$ follows $p_{z}$, which usually is a multi-dimensional Gaussian distribution and $x$ is sampled from the real distribution of  $p_{\mathrm{data}}$. $G(\cdot)$ produces a synthetic sample $\hat{x}=G(z)$. The discriminator $D(\cdot)$ reads in a sample (either $x$ and $\hat{x}$) and outputs a scalar value known as a critic.  The generator is trained to produce a synthetic sample $\hat{x}$ similar to real samples $x$, while the discriminator is trained to distinguish $\hat{x}$ and $x$ by yielding a lower scalar critic related to $\hat{x}$ and higher scalar critic to the $x$. 

Training GAN is an alternative min-max game between discriminator and generator. It is adversarial in that  the discriminator learns to better distinguish the synthetic samples from the real ones while the generator learns to produce more realistic samples by improving the approximation to the real sample distribution. The competition and cooperation between discriminator and generator will promote the closeness of the the generative distribution to the real sample distribution. A generic structure of GAN can be illustrated in Figure \ref{fig:gan}. 

Well-designed GAN models produce realistic samples. However, the value function developed in Eq.~\eqref{eq:gan loss} can be limited when applied to categorized data sets where the inputs multiple classes. The generator will learn the overall distribution of the whole dataset, while the label of a synthetic sample will be randomly specified. Hence, for problems like earthquake detection, where there are multiple classes of data -- earthquakes, noise, etc -- there is a need to incorporate label information to the GAN. 

\subsection{Conditional GAN}
\label{sec:conditional gan}

\begin{figure}
\centering
\subfigure[]{
 \includegraphics[width=0.40\columnwidth]{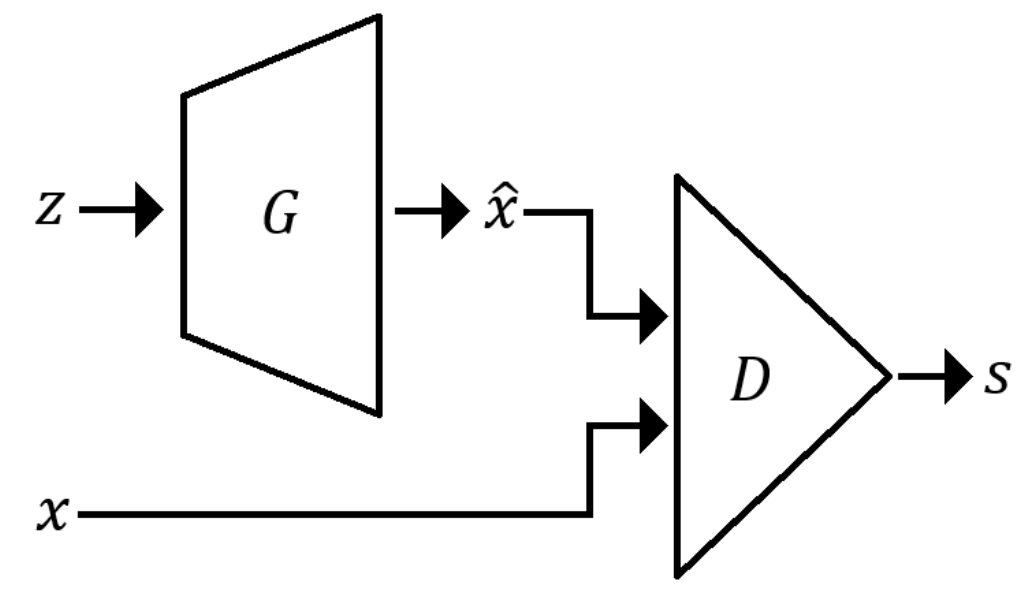}
 \label{fig:gan}}
\,\,\,\,\,\,
\subfigure[]{
 \includegraphics[width=0.40\columnwidth]{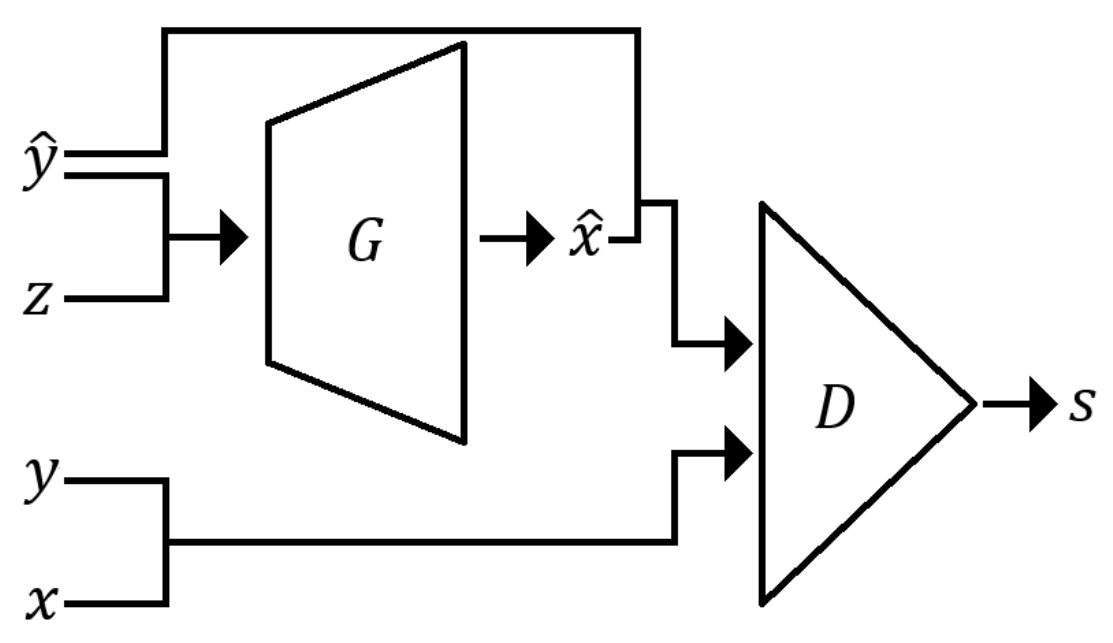}
 \label{fig:cond gan}}
\caption{An illustration of the structure of (a) GAN and (b) conditional GAN models. (a) The generator $G$ transforms an input Gaussian noise vector into a synthetic data sample $\hat{x}$. The discriminator $D$ distinguishes between real data $x$ and synthetic data $\hat{x}$ through its critic score $s$. (b) In a conditional GAN, class labels $y$ are incorporated into both the generator and discriminator.}
\end{figure}

With an input of a label information $y$ to both generator and discriminator, a traditional GAN can be turned into a conditional GAN~\cite{CondGAN}. The structure of conditional GAN can be illustrated in Figure~\ref{fig:cond gan}. It allows the generator to produce samples that belong to given categories. The dynamics of the value function of conditional GAN can be written as
\begin{equation} \label{eq:cond gan}
\underset{G}{\mathrm{min}}\,\underset{D}{\mathrm{max}}\,~V(D, G, x, z) = \mathbb{E}_{x \sim p_{\mathrm{data}} }[\log{(D(x|y))}] + \mathbb{E}_{z \sim p_{z}}[\log{(1-D(G(z|\hat{y})|\hat{y}))}], 
\end{equation}
where  $y$ is the label of real sample $x$, and $\hat{y}$ is the targeted label.

In conditional GAN~(Figure~\ref{fig:cond gan}), the generator, $G$, reads in the prior-label pair of  $(z, ~\hat{y})$, where $\hat{y}$ is the targeted label of the synthetic sample $\hat{x}$. The discriminator, $D$, reads in the sample-label pairs ($(x,~y)$ or $(\hat{x},~\hat{y})$), and yields a scalar critic for each pair. Besides evaluating the sample itself, $D$ will also justify whether the sample matches its label. A synthetic sample-label pair $(\hat{x},~\hat{y})$ can only achieve a high value of scalar critic from $D$ when both $\hat{x}$ is realistic enough and $\hat{x}$ belongs to the targeted label of $\hat{y}$. Consequently, $G$ is forced to generate high-quality synthetic sample $\hat{x}$ that will match  the targeted label $\hat{y}$.

\section{Data Description and Pre-processing}
\label{section:data}

\subsection{Raw Seismic Waveform Time Series} 
\label{sec:earthquake event}

The scientific focus of this paper is on the earthquake detection problem. Broadband seismometers are highly sensitive instruments that are capable of recording small earthquakes. This sensitivity comes with a tradeoff, as they will also record background noise and other non-earthquake signals. Earthquake detection can be posed mathematically as a classification problem, where the objective is to partition the observed waveforms into different classes. In the simplest case, which we adopt in this work, there are two classes of interest: earthquake and non-earthquake (or noise). 

The duration and characteristics of earthquake waveforms may vary significantly from event to event, depending on the source duration and mechanism, the source-receiver distance, and attenuation along the raypath and in the shallow subsurface. However, all earthquake waveforms exhibit a universal set of features governed by underlying geophysical constraints. The physics of seismic wave propagation imposes temporal and polarization structure on earthquake waveforms. For example, P-waves arrive before S-waves and are typically of lower amplitude and more visible on vertical-component sensors. Any machine learning algorithm meant to synthesize realistic earthquake waveforms will need to account for these physical constraints in their model, either explicitly or implicitly. 

\subsection{Dataset Generation}
\label{sec:dataset}

\begin{figure}[ht]
	\begin{center}
	\centering
    \includegraphics[width=0.9\columnwidth]{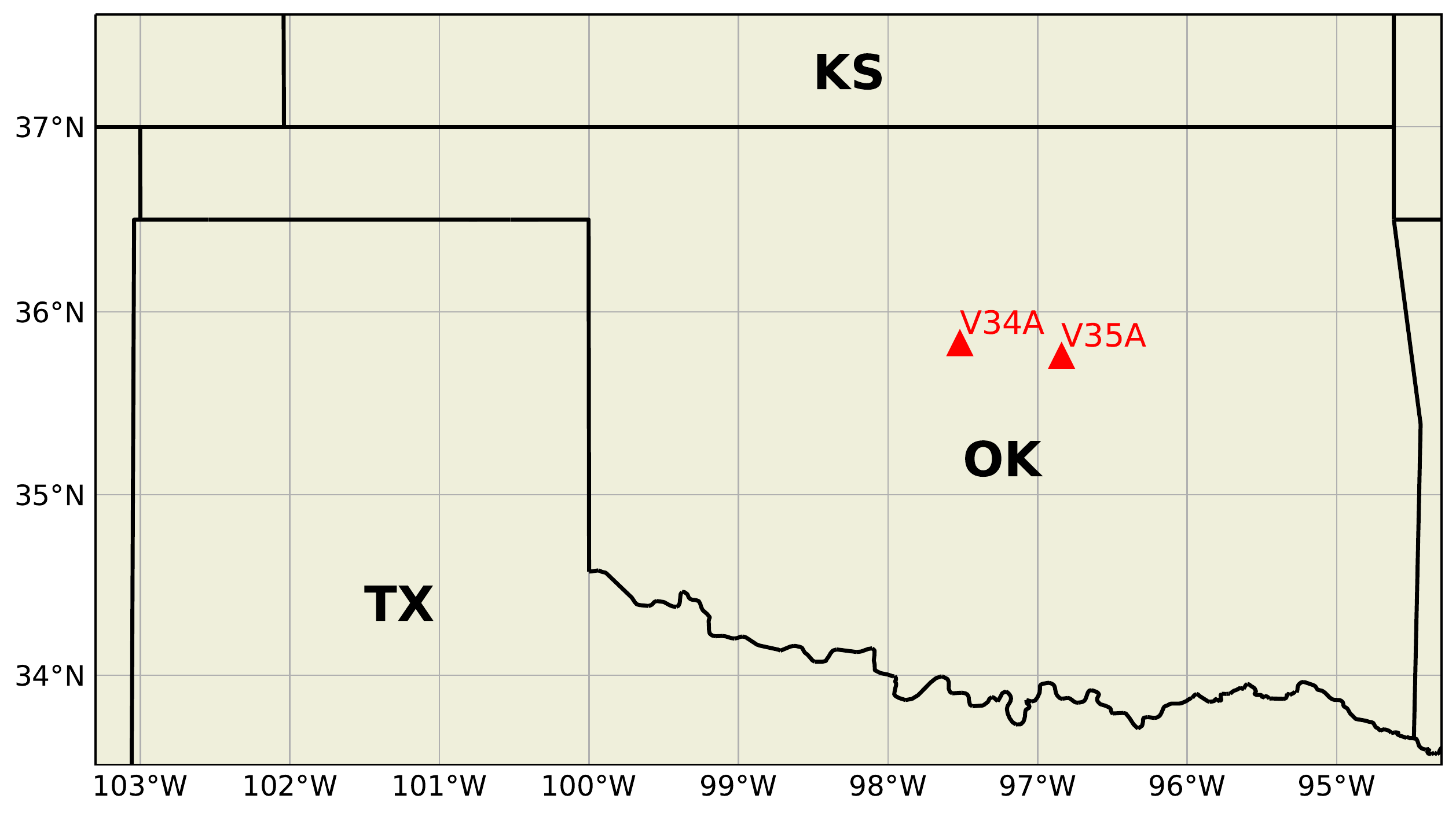}
	\end{center}
	\caption{Stations V34A and V35A are TA network seismometers that were located in the state of Oklahoma, USA, during a time period from 2009 to 2011. Waveform data acquired from these stations are used to test the performance of our model.}
    \label{fig:station}
\end{figure}

We use two field datasets to validate the performance of our model. Each dataset is processed from raw waveforms data acquired at two stations from the Transportable Array (network code TA): V34A and V35A. Station V34A and V35A are located in the state of Oklahoma, approximately 60 -- 80 km away from the Oklahoma City, as shown in Figure~\ref{fig:station}. Station V34A operated at its Oklahoma site from Nov 1st 2009, 21:59:18 to Sep 3rd 2011, 13:55:28, while station V35A operated at its Oklahoma site from Mar 14th 2010, 18:47:42 to Sep 4th 2011, 23:59:58. Both stations are three-component low-broadband seismometers (channel codes BHE, BHN, BHZ) operated at sampling rate of 40~Hz.

% To compile an earthquake catalog, we use a modified version of the earthquake detection method described in \citeA{Tidal-2017-Delorey}. In this method, patterns of arrivals are first identified by an autonomous seismic tremor detection method across a seismic array that emanate from local earthquakes, and then each identified arrival is examined by an expert so that false alarms are precluded. 

\begin{figure}[ht]
\centering
\subfigure[]{\includegraphics[width=0.4\columnwidth]{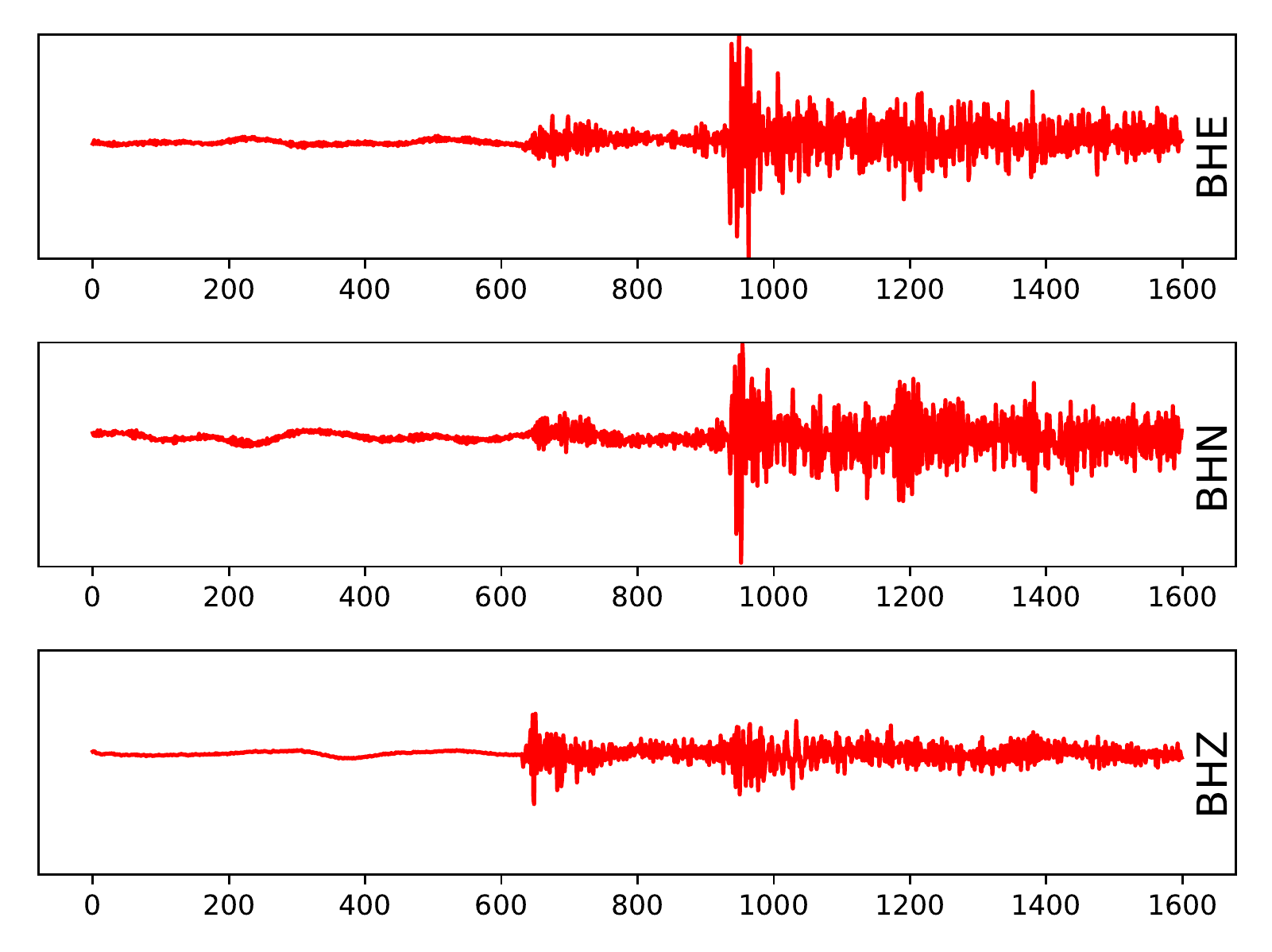}
 \label{fig:real sample anno pos raw 1}}
\,\,\,\,\,\,\,
\subfigure[]{\includegraphics[width=0.4\columnwidth]{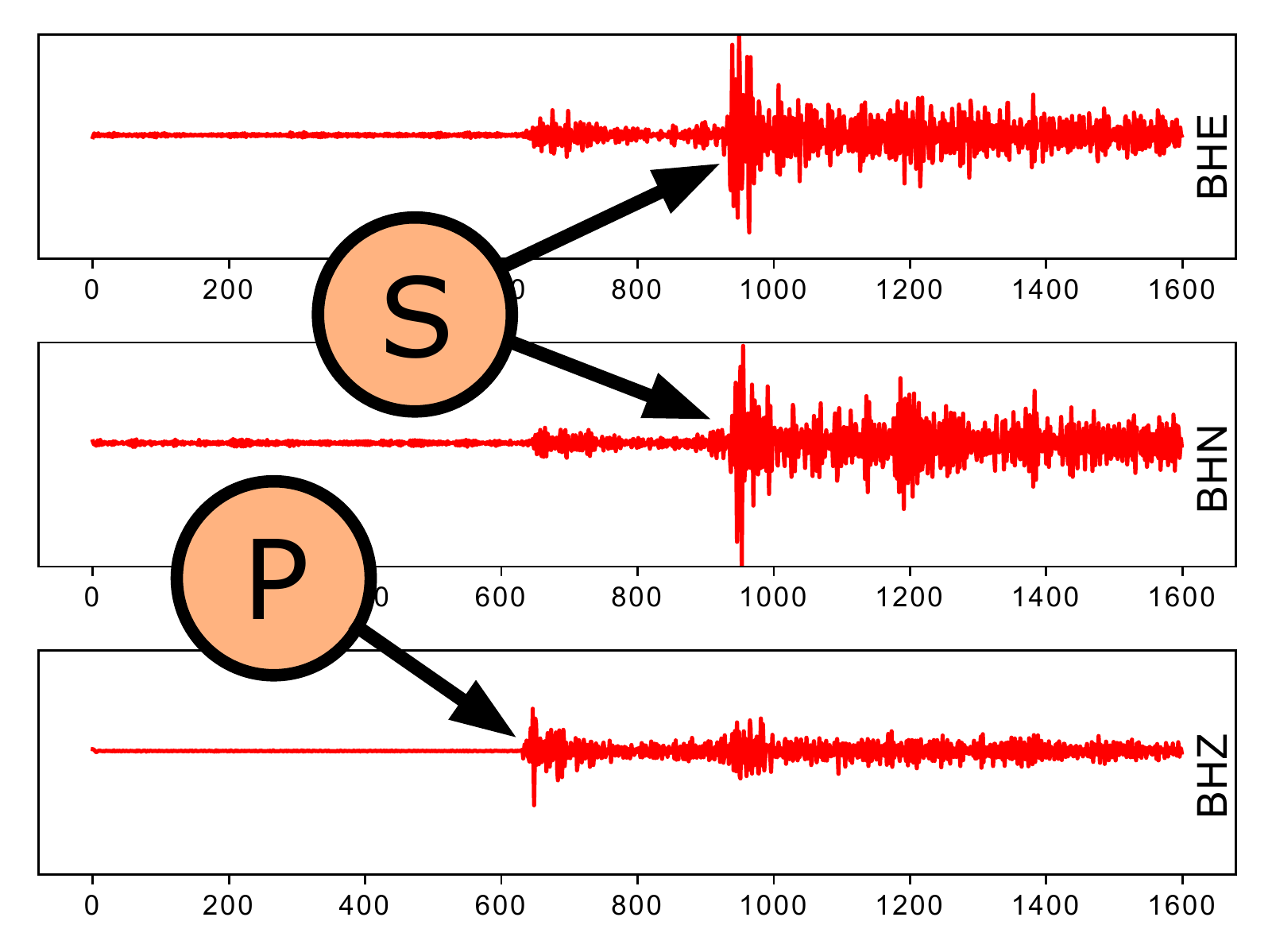}
 \label{fig:real sample anno pos high 1}}
\caption{Illustration of a 3-component real earthquake waveform event obtained from the seismic station V34A. We include both the raw waveform in (a) and the filtered waveform in (b). The three rows in each figure show components of BHE, BHN and BHZ, respectively.  In (b), we indicate the arrival time of P and S waves.}
\label{fig:Earthquake}
\end{figure}

To compile an earthquake catalog, we use a slightly modified catalog data obtained from Oklahoma Geological Survey (OGS). The original catalog from OGS can be obtained from \cite{OGS}. In our catalog, we have $1,025$ earthquakes from station V34A, and $1,120$ earthquakes from station V35A during the time of operation in our study area. An example of an earthquake detection is shown in Figure~\ref{fig:Earthquake}.

In designing a machine learning based detection algorithm, the maximum duration of an earthquake waveform is an important parameter to decide. We apply a consistent window size to all earthquakes in this work. We find that a window size of $40$ seconds (1600 time steps) to be a good option in that it is large enough to cover any individual earthquakes while small enough to facilitate an efficient training~\cite{zhang2019adaptive}. Our algorithm is thus designed to operate on time series samples defined as $3$-component vectors of length $1,600$. We provide both positive and negative seismic samples based on whether or not there is an earthquake event included in the time series. This parameterization is sufficient for our purposes, as earthquakes in our datasets are relatively sparse in occurrence over time. We find that the duration between any two neighboring earthquakes in our catalog is never less than $3,200$ time steps, so that any two consecutive earthquakes will not be included in the same positive sample of length $1,600$ time steps. 

With all the aforementioned details on our raw seismic waveform, we build our dataset guided by four rules:
\begin{enumerate}
\item Each positive sample shall cover a single earthquake;
\item Negative samples shall not cover any earthquake;
\item  Positive and negative samples shall not overlap with each other;
\item The number of positive and negative samples shall be balanced.
\end{enumerate}

\begin{figure}[ht]
\centering
\subfigure[]{\includegraphics[width=0.30\columnwidth]{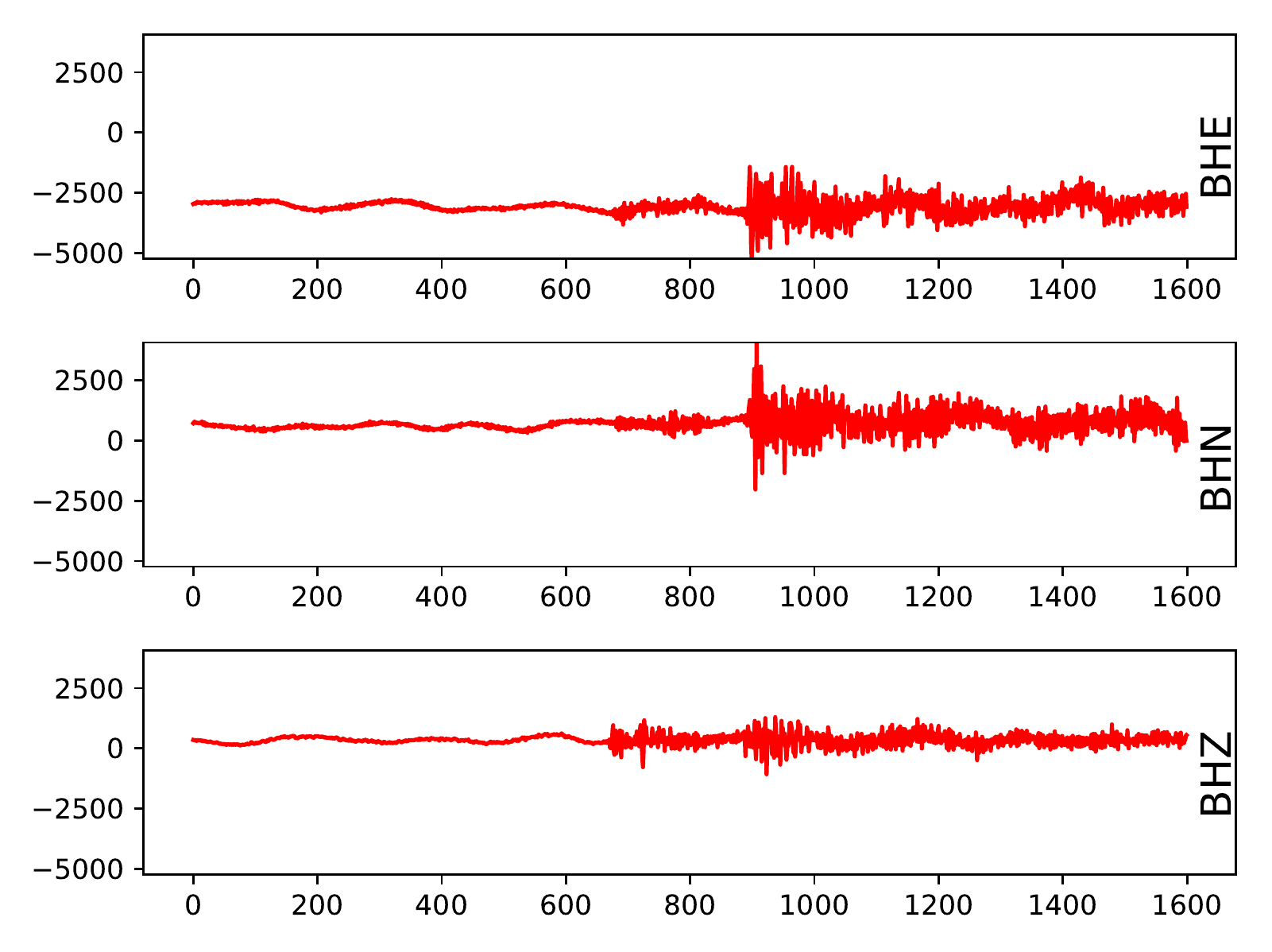}
 \label{fig:real sample pos raw 1}}
\subfigure[]{\includegraphics[width=0.30\columnwidth]{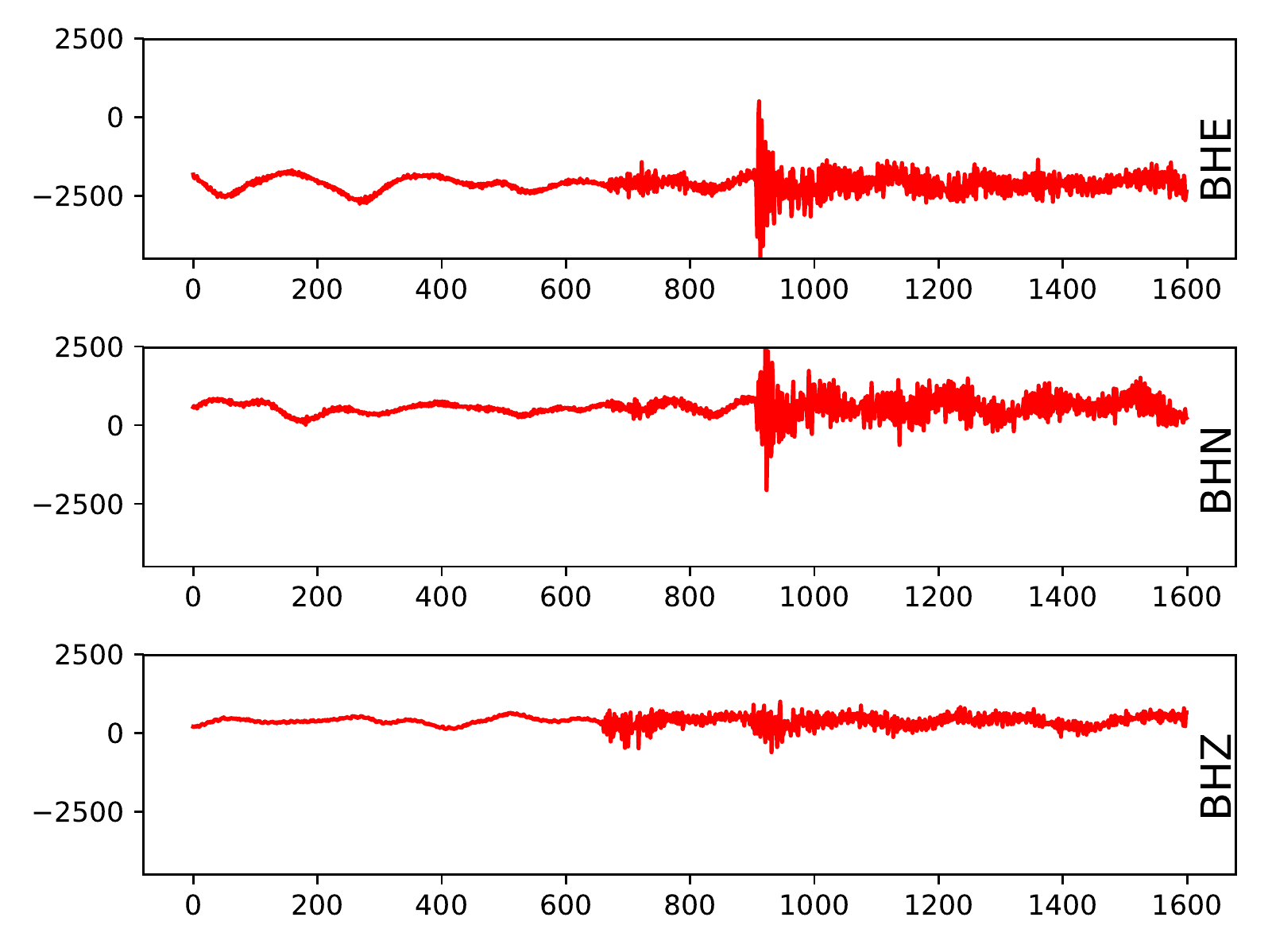}
 \label{fig:real sample pos raw 2}}
\subfigure[]{\includegraphics[width=0.30\columnwidth]{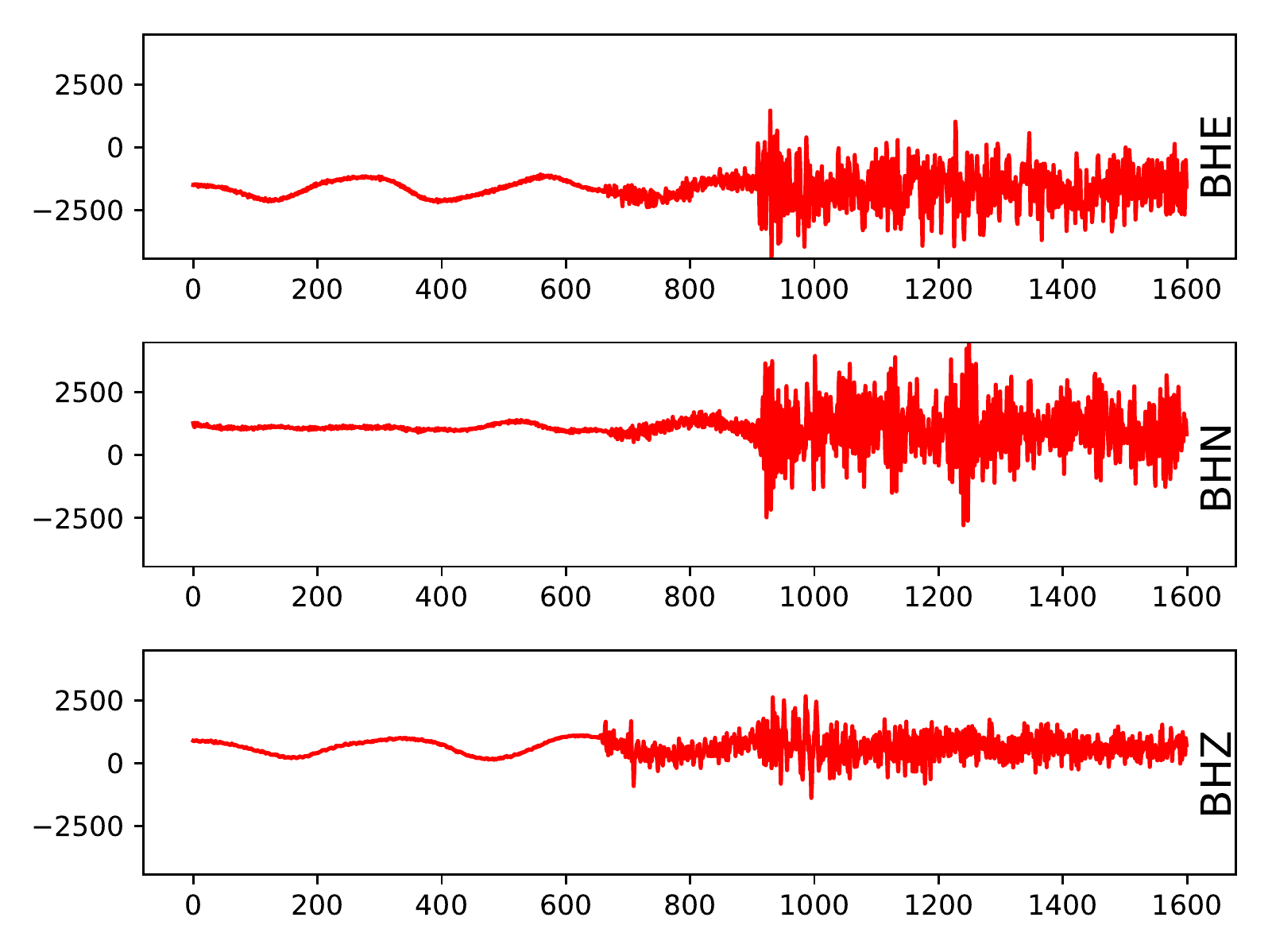}
 \label{fig:real sample pos raw 3}}
\vspace{-4mm}
\subfigure[]{\includegraphics[width=0.30\columnwidth]{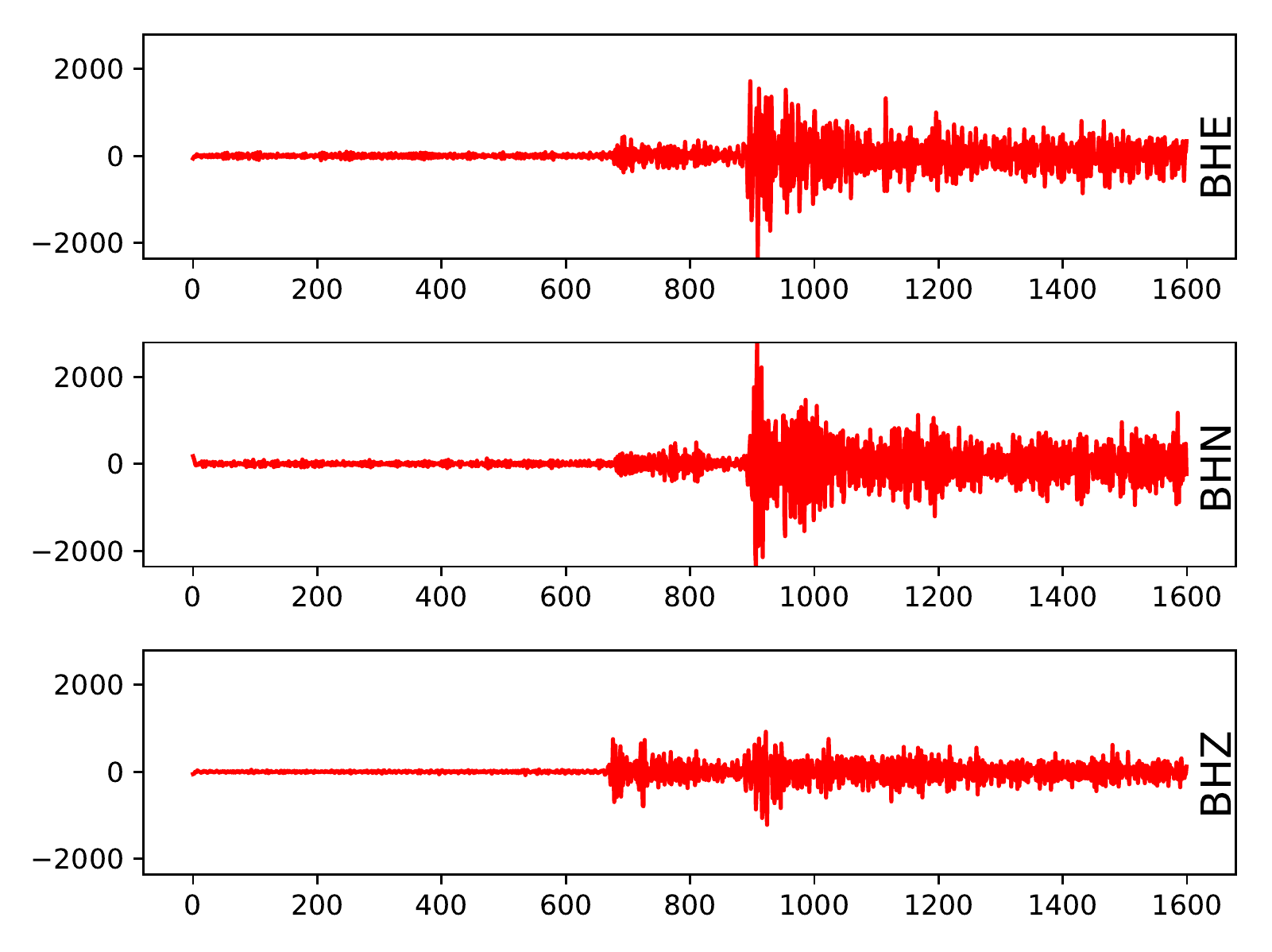}
 \label{fig:real sample pos high 1}}
\subfigure[]{\includegraphics[width=0.30\columnwidth]{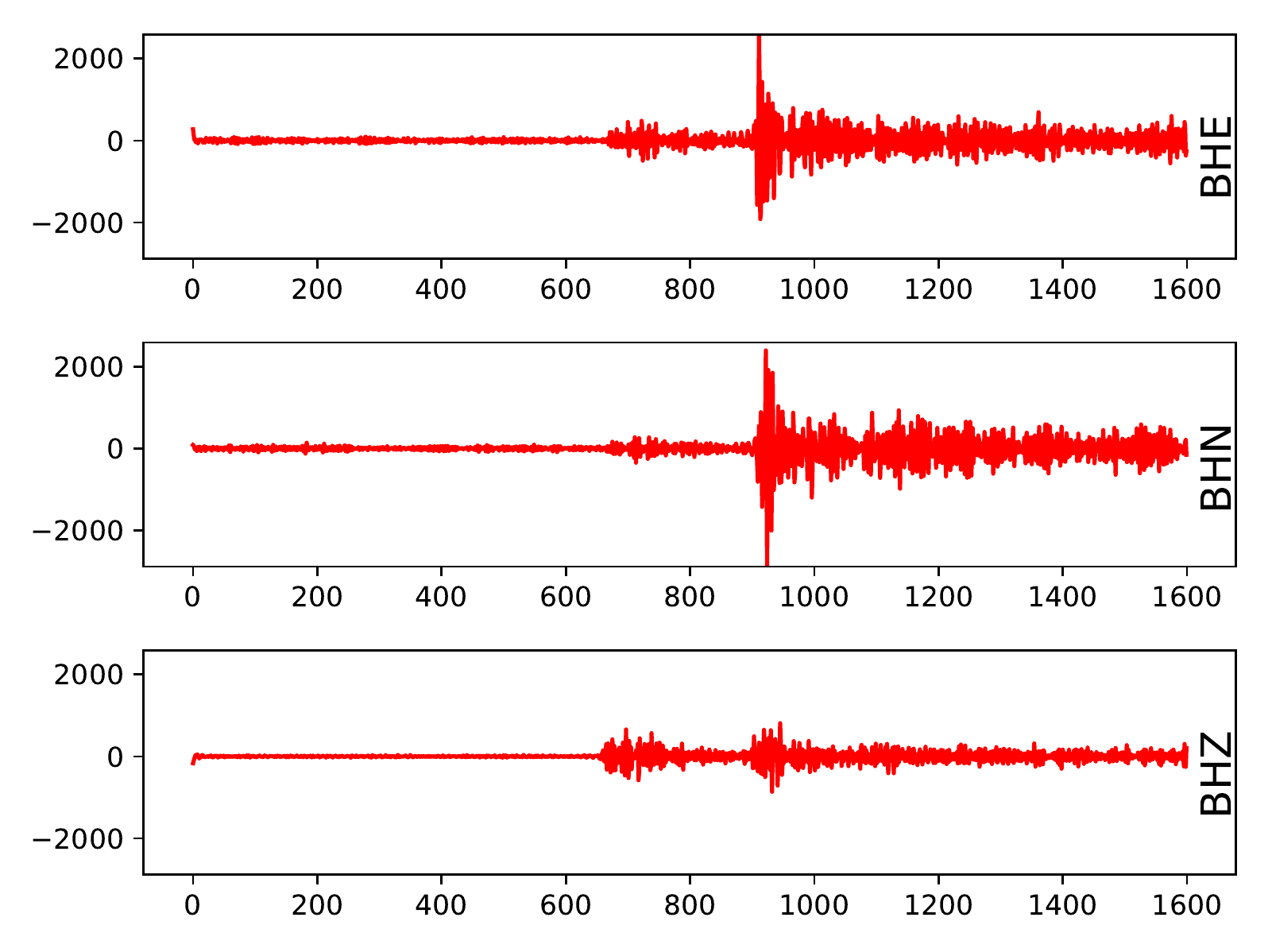}
 \label{fig:real sample pos high 2}}
\subfigure[]{\includegraphics[width=0.30\columnwidth]{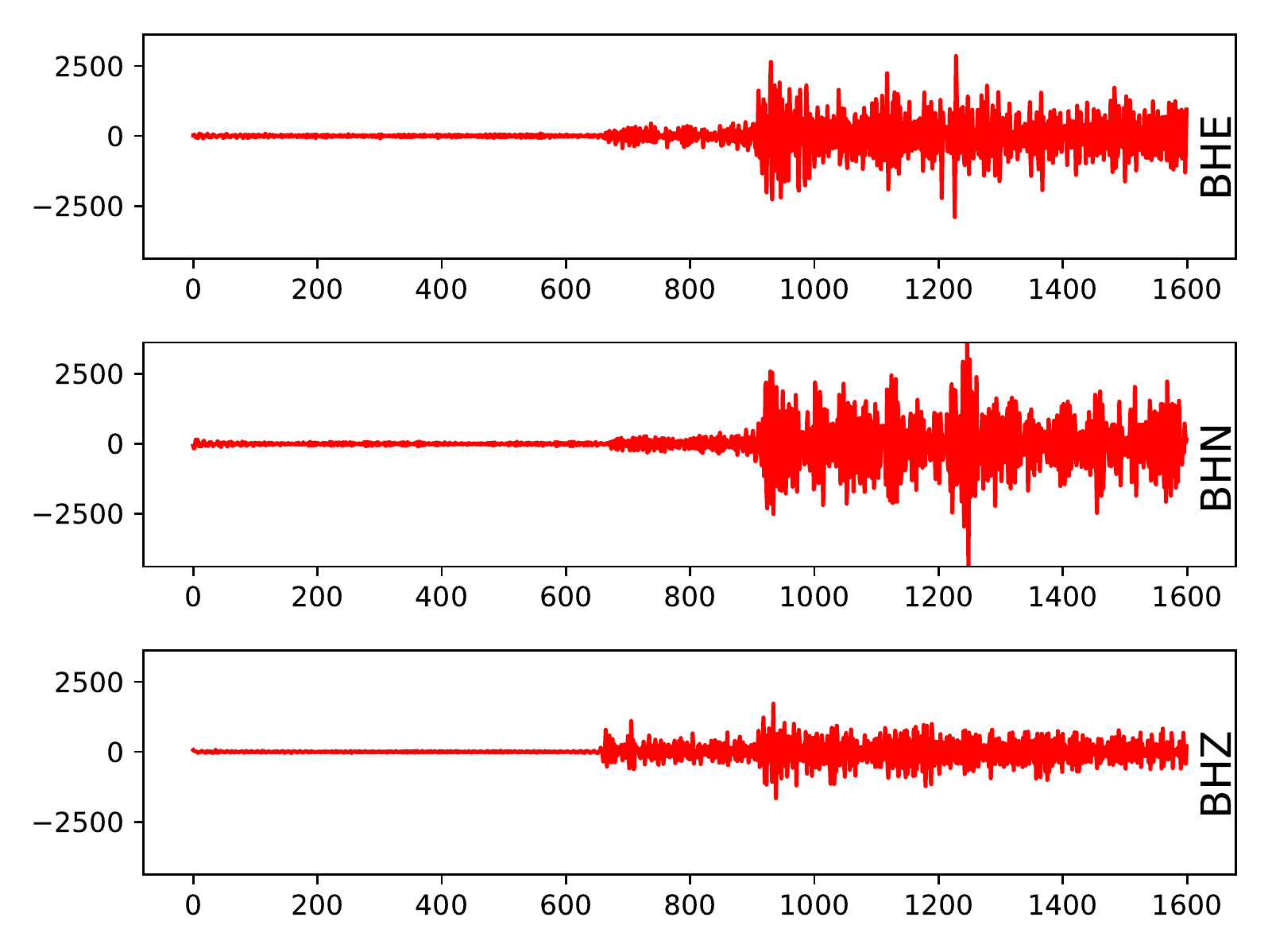}
 \label{fig:real sample pos high 3}}
\caption{Illustration of three positive waveform samples (\ref{fig:real sample pos raw 1}, \ref{fig:real sample pos raw 2} and \ref{fig:real sample pos raw 3}) and their corresponding filtered waveforms (\ref{fig:real sample pos high 1}, \ref{fig:real sample pos high 2} and \ref{fig:real sample pos high 3}). Each sample consists of a 40-second period of seismic waveform from station V34A with a sampling rate of $40$ Hz. Row~1 shows the raw waveforms of the positive samples, and Row.~2 shows their filtered waveforms.}
\label{fig:real sample Pos}
\end{figure}

\begin{figure}[ht]
\centering
\subfigure[]{\includegraphics[width=0.30\columnwidth]{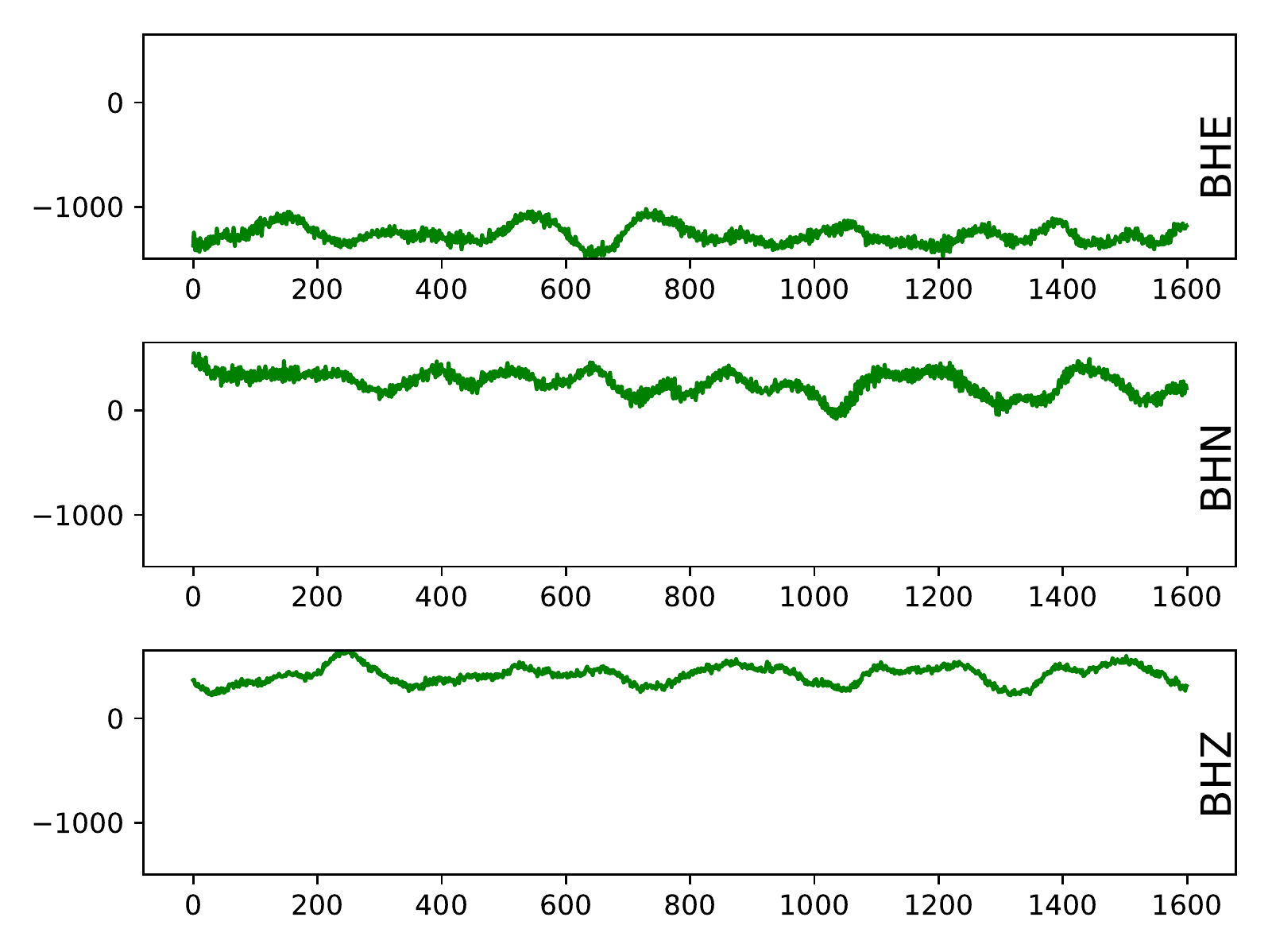}
 \label{fig:real sample neg raw 1}}
\subfigure[]{\includegraphics[width=0.30\columnwidth]{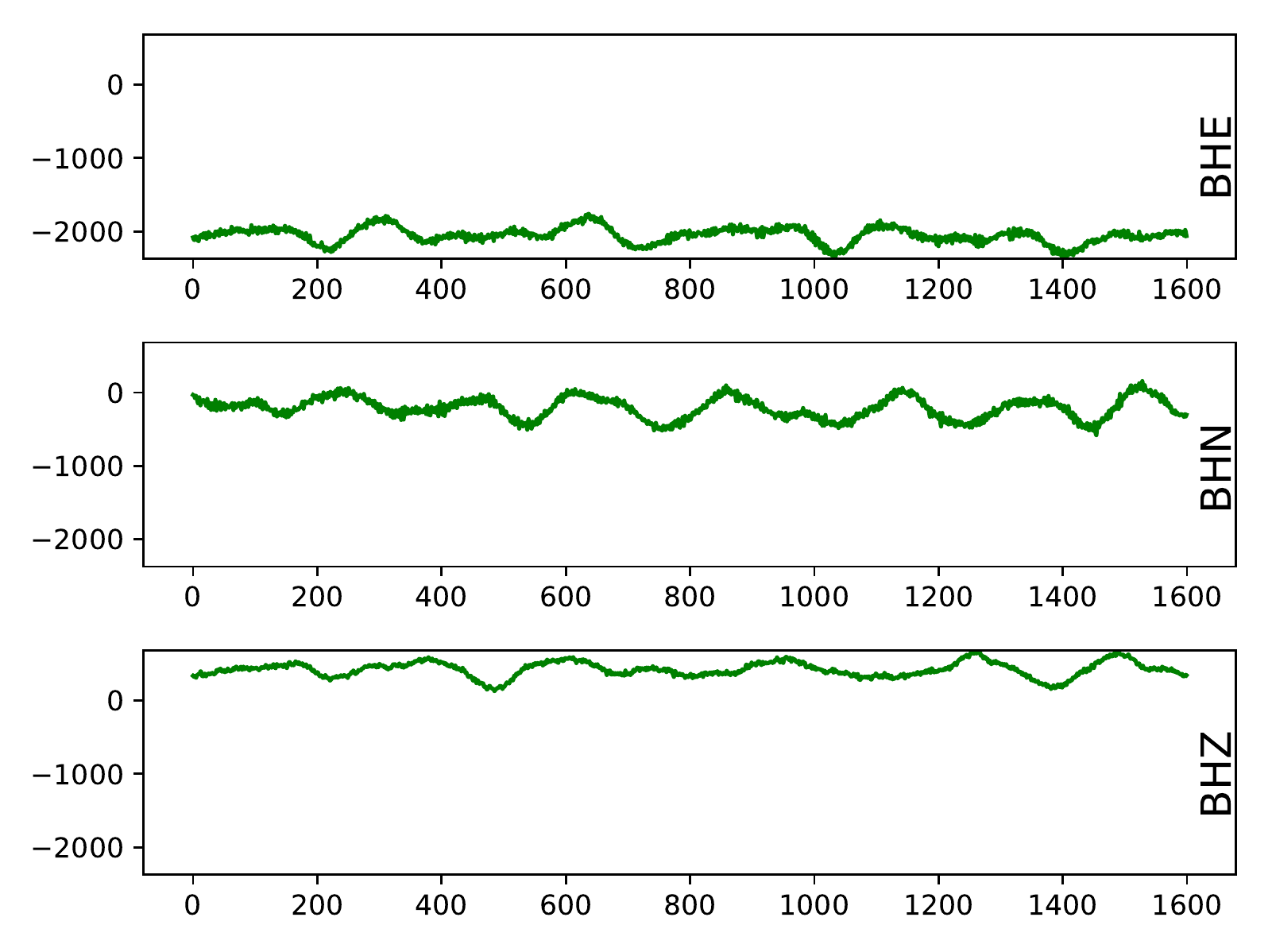}
 \label{fig:real sample neg raw 2}}
\subfigure[]{\includegraphics[width=0.30\columnwidth]{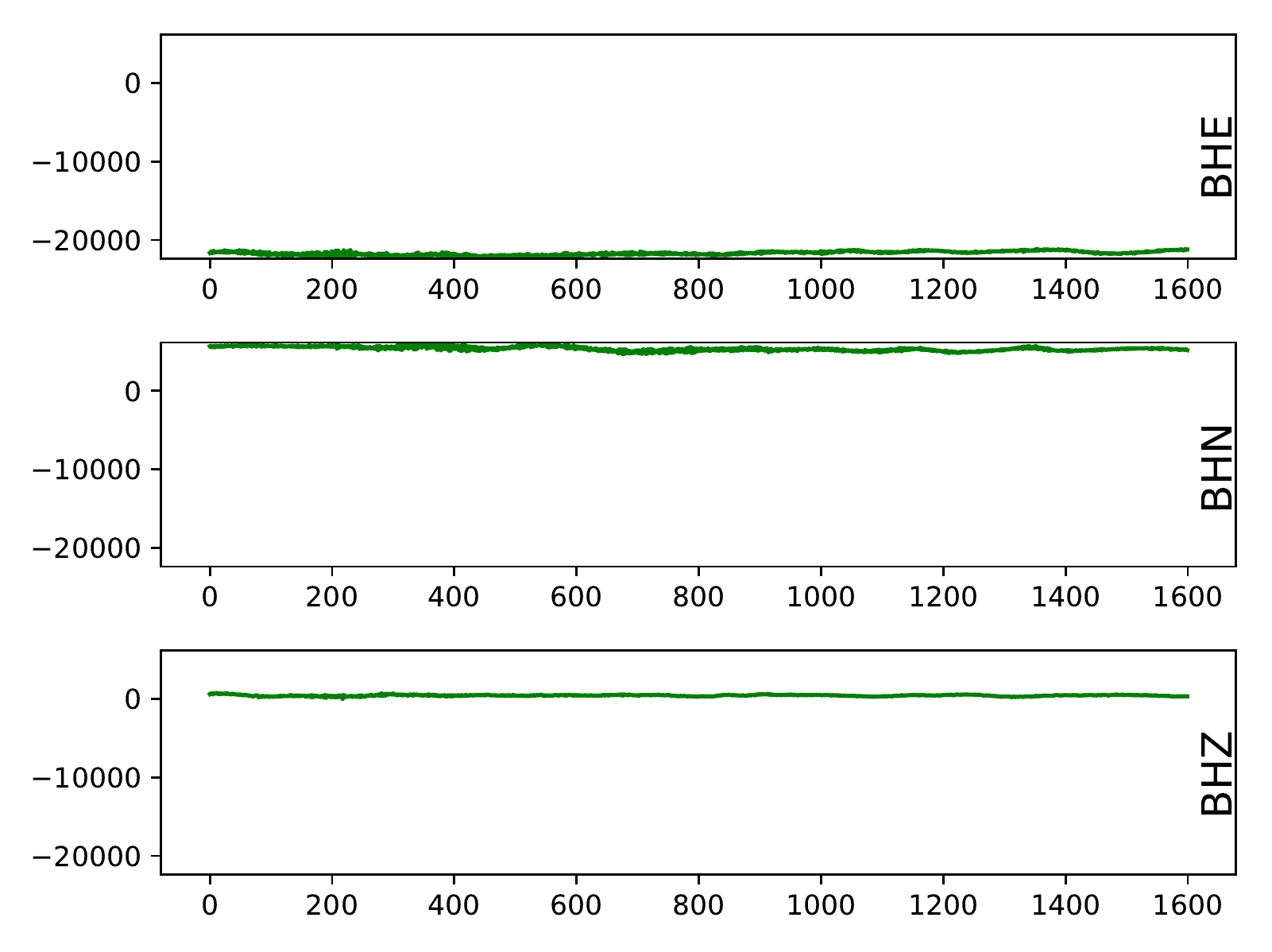}
 \label{fig:real sample neg raw 3}}
\vspace{-4mm}
\subfigure[]{\includegraphics[width=0.30\columnwidth]{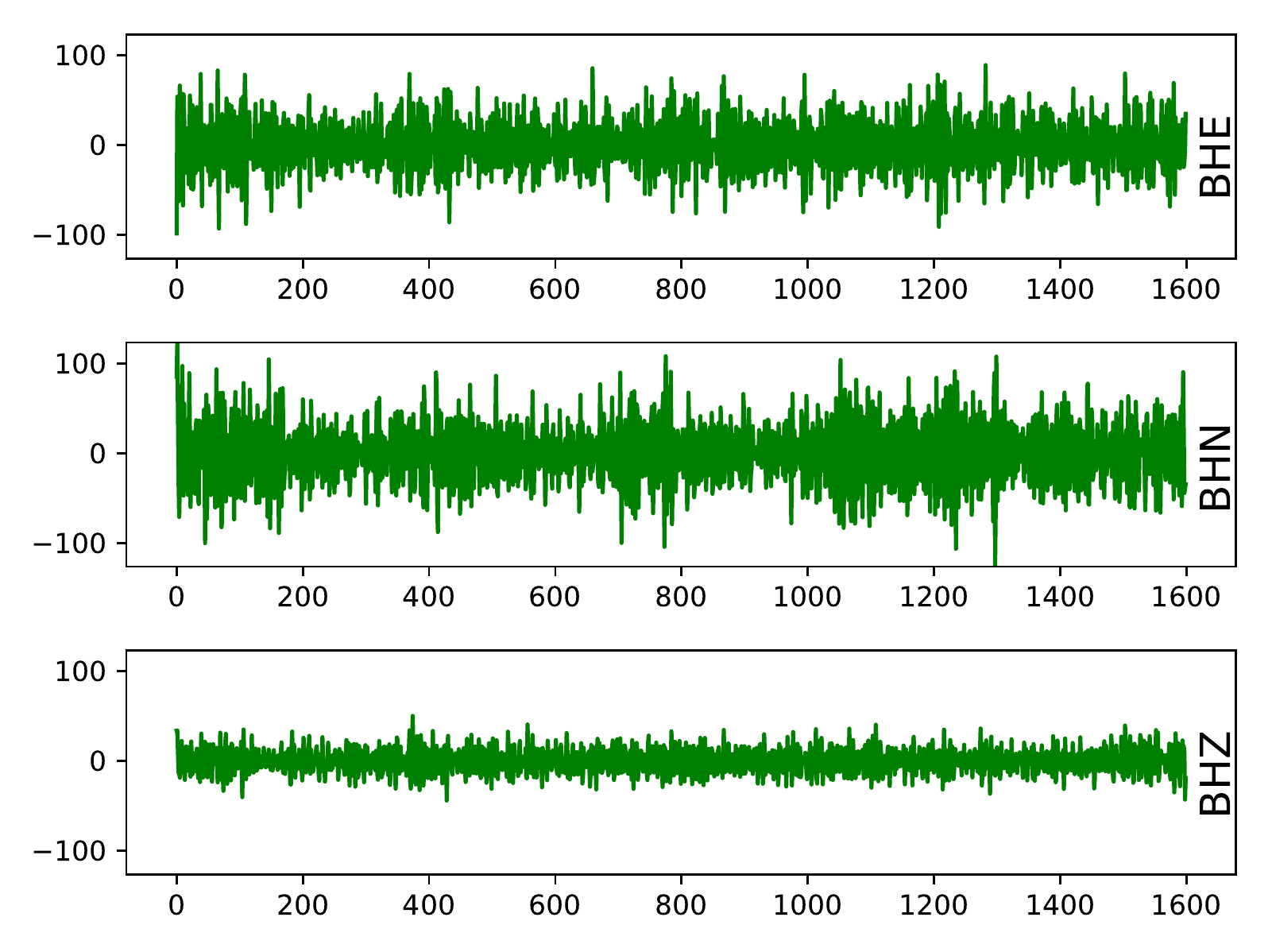}
 \label{fig:real sample neg high 1}}
\subfigure[]{\includegraphics[width=0.30\columnwidth]{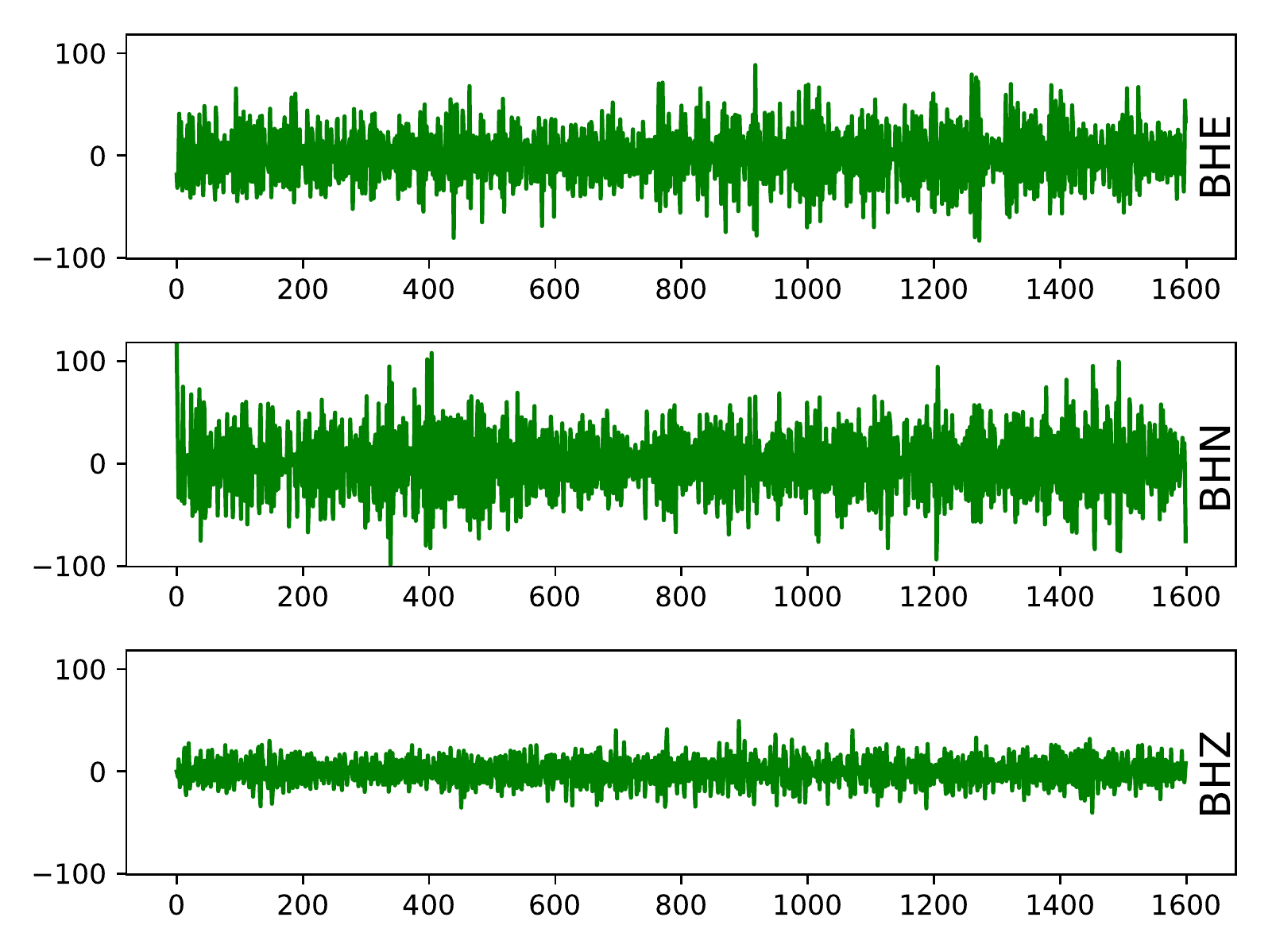}
 \label{fig:real sample neg high 2}}
\subfigure[]{\includegraphics[width=0.30\columnwidth]{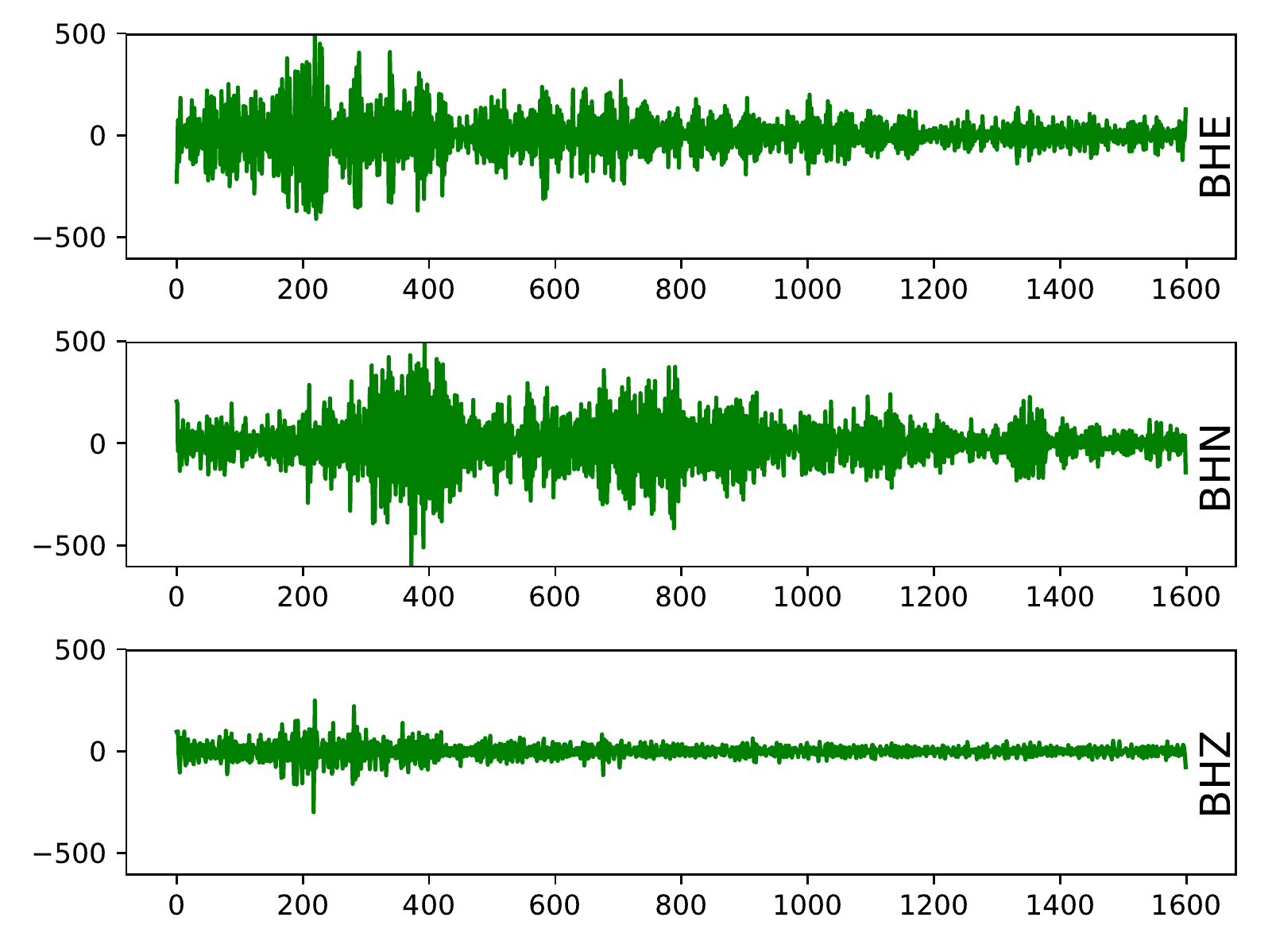}
 \label{fig:real sample neg high 3}}
\caption{Illustration of three negative waveform samples (\ref{fig:real sample neg raw 1}, \ref{fig:real sample neg raw 2} and \ref{fig:real sample neg raw 3}) and their corresponding filtered waveforms (\ref{fig:real sample neg high 1}, \ref{fig:real sample neg high 2} and \ref{fig:real sample neg high 3}). Each sample consists of a 40-second period of seismic waveform from station V34A with a sampling rate of $40$ Hz. Row~1 shows the raw waveforms of the negative samples, and Row.~2 shows their filtered waveforms.}
\label{fig:real sample Neg}
\end{figure}

We use the station V34A as an example to demonstrate this procedure. For each seismic event located at a time stamp~$t$, we firstly sample three offsets $o_{1}$, $o_{2}$ and $o_{3}$ from a discrete uniform distribution of $\mathit{Unif}[ -600,~600]$.  We then create three positive samples by segmenting three intervals length of $1,600$ centered at $t + o_{1}$, $t + o_{2}$ and $t + o_{3}$ on the raw waveform data. We repeat this procedure for each of the 1,025 events detected on V34A, providing us a total of $1,025\times3=3,075$ positive samples. We balance these positive samples by randomly selecting a total of $3,075$ time segments with a length of  $1,600$ from the remainder of the raw seismic waveform. Eventually, the positive and negative samples together will result in a total data size of $6,150$ for station V34A. Similar procedures can be applied to station V35A, and that will provide us with a dataset of size $6,432$ which consists of $3,216$ positive samples and $3,216$ negative samples. Figures~\ref{fig:real sample Pos} and ~\ref{fig:real sample Neg} compare waveforms from positive and negative samples on all three components.

% \begin{table}[h!]
% \centering
%  \begin{tabular}{| c || c | c |} 
%  \hline
%  Station ID & Positive (Earthquake) & Negative (Non-Earthquake)  \\
% \hline
% V34A & 3,075    & 3,075    \\ 
%  \hline
% V35A & 3,216    & 3,216  \\
%   \hline
% \end{tabular}
% \caption{Summary of positive (earthquake) and negative (non-earthquake) samples acquired from stations of V34A and V35A.}
% \label{tab:DataSamples}
% \end{table}

\subsection{Normalization}

In raw seismic time series, the digitized values logged by the seismic stations are spread over a range of $ \sim \pm 10^{7}$ counts. To effectively learn the features of the seismic waveforms, the dataset needs to appropriately normalized. In particular, for a $3$-component, $1,600$-length raw seismic time series of $[e, n, v]$, we subtract the mean and normalize each by their respective standard deviations:
% \begin{equation}
% [\hat{e},\hat{n},\hat{v}] = \frac{[e - \bar{e}, n - \bar{n}, v - \bar{v}]}{\sigma([e - \bar{e}, n - \bar{n}, v - \bar{v}])},
% \label{eq:normalization}
% \end{equation}
\begin{equation}
[\hat{e},\hat{n},\hat{z}] = \left[ \frac{e - \bar{e}}{\sigma_e},\frac{n - \bar{n}}{\sigma_n}, \frac{z - \bar{z}}{\sigma_z} \right],
\label{eq:normalization}
\end{equation}
where $e$, $n$, and $v$ stand for the raw measurements of velocity values in three components of BHE, BHN and BHZ, respectively. Through comparison to other normalization schemes~\cite{zhang2019adaptive}, the one in Eq.~\eqref{eq:normalization} yields the best results.

\section{Model Design}
\label{sec:model verification}

\subsection{Model}

\begin{figure*}
\centering
\subfigure[]{
 \includegraphics[width=0.70\columnwidth]{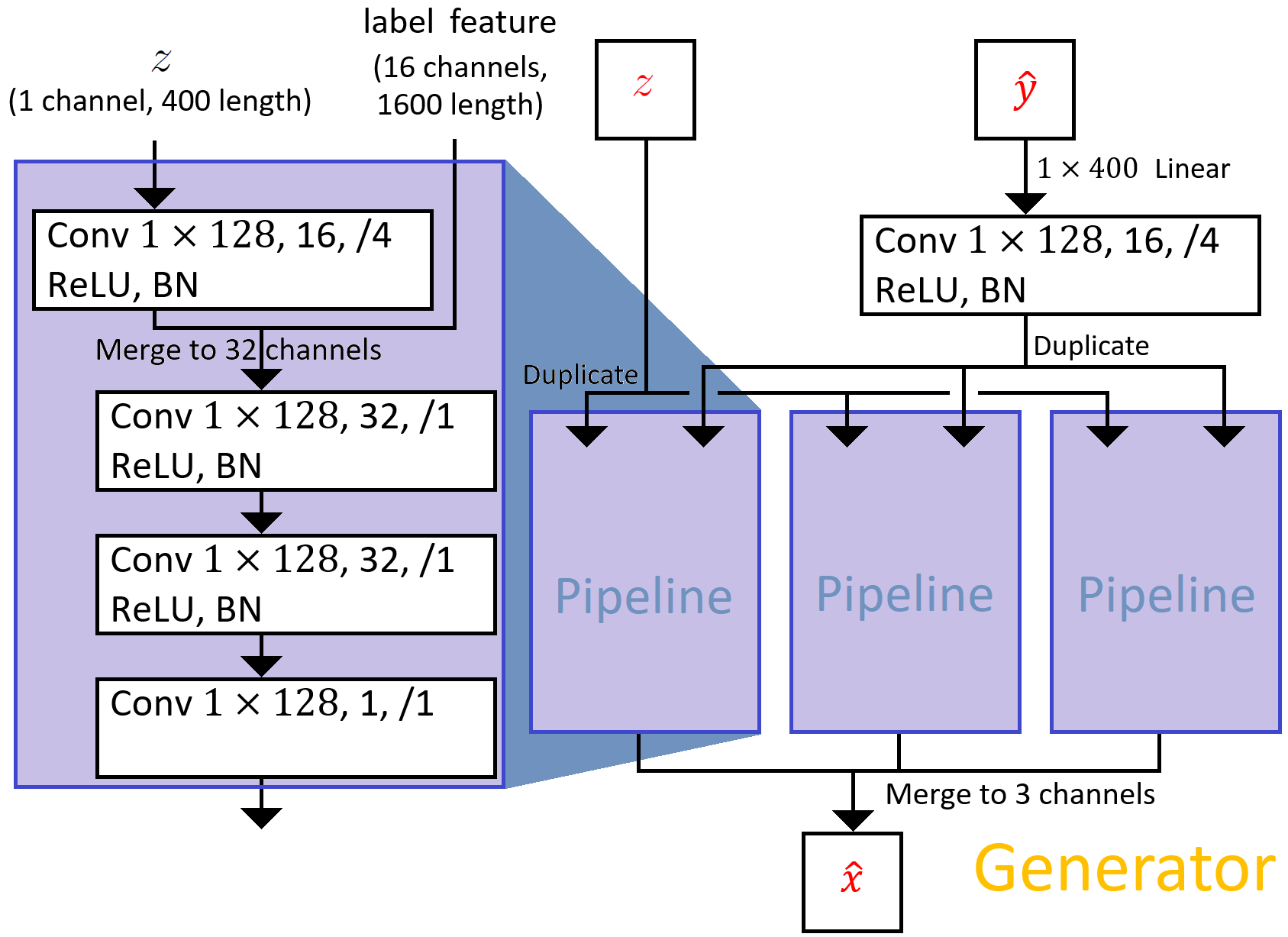}
 \label{fig:generator}}
\,\,\,\,\,\,
\subfigure[]{
 \includegraphics[width=0.70\columnwidth]{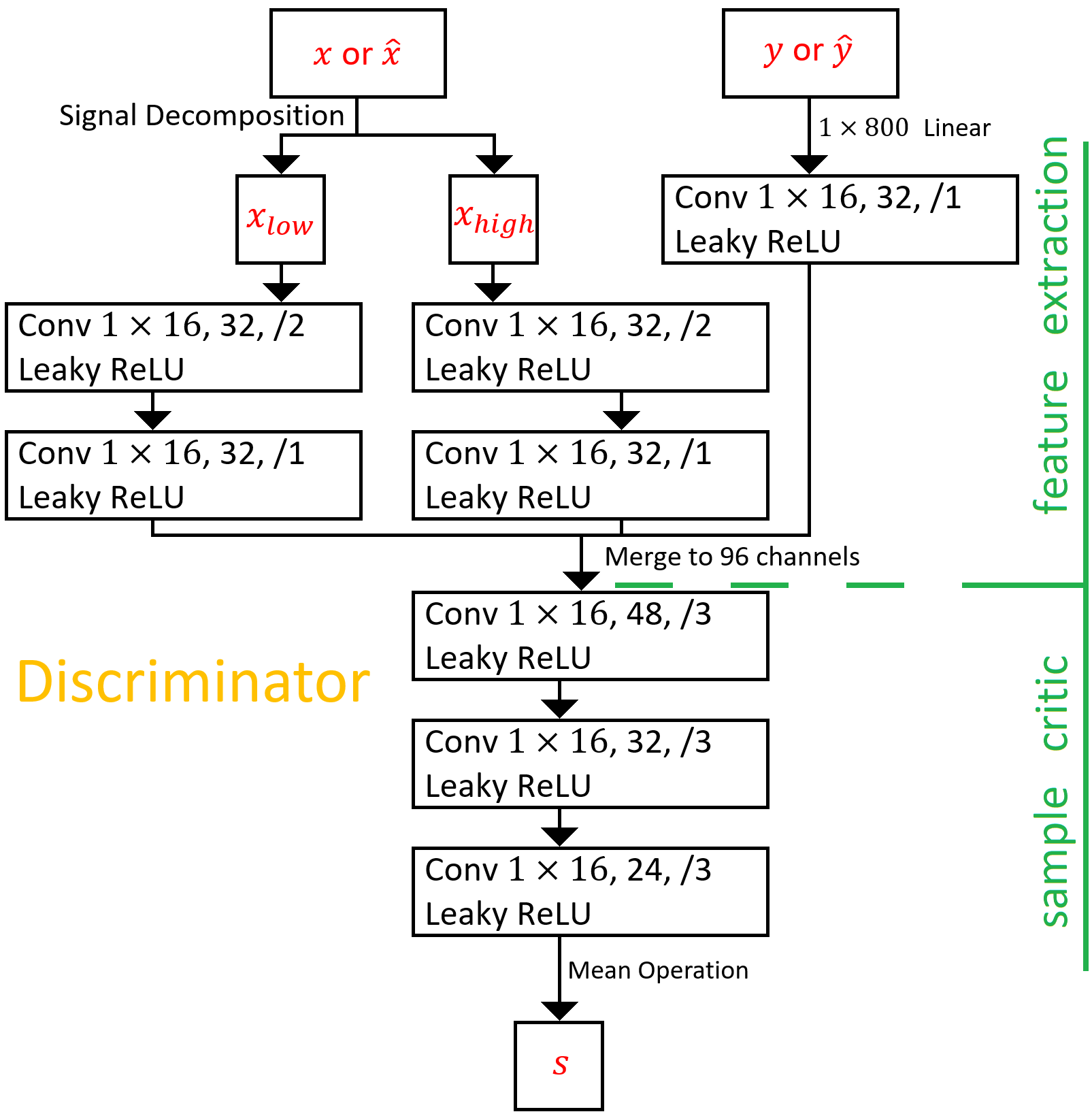}
 \label{fig:discriminator}}
\caption{An illustration of the network structure of our model: generator (a) and discriminator (b). The data dimensions and mathematical operations for each layer are listed in each panel.  Each long box in the figure represents a layer in our model. For example, ``Conv $1 \times 128$, 16, /4, ReLU, BN'' describes a 1D convolutional layer using 16 kernels with kernel size $1 \times 128$, stride 4, ReLU activation function, and batch normalization.}
\label{fig:network structure}
\end{figure*}

Our model is based on the structure of a conditional GAN \cite{CondGAN}. The main structure of our model is illustrated in Figure~\ref{fig:network structure}, which consists of two networks: the generator~(Figure~\ref{fig:generator}) and the discriminator~(Figure~\ref{fig:discriminator}). To increase the quality of the synthesized waveforms from different seismic stations, we train a separate GAN model for each station but using the same network structure. 

\subsubsection{Generator Structure}

We design our generator to comprise of three pipelines to synthesize each component of the data individually. All three pipelines share the same input and follow an identical network structure as shown in Figure~\ref{fig:generator}, but otherwise do not interact or share trainable parameters like weights. Each pipeline is a four-layer convolutional network. As shown in Figure~\ref{fig:generator}, the input vector $z$ is a Gaussian noise vector of length $400$, while $\hat{y}$ is a binary scalar with $\hat{y}=1$ representing positive event.  With both $z$ and $\hat{y}$ becoming available, we will pass $z$ through a transposed 1D convolution layer to obtain an augmented 1D feature vector of length $1,600$. In parallel, the scalar input of $\hat{y}$ will be augmented to be length of $1,600$, and then further concatenated with augmented $z$ vector.  Similar to the conventional DCGAN, we use an additional 3 layers of 1D convolution layers to synthesize one component data in the seismic sample of $\hat{x}$. Similarly, we can obtain the other two components of the synthetic seismic sample of $\hat{x}$ through two other pipelines.

\subsubsection{Discriminator Structure}

The discriminator is used to evaluate the quality of input samples, real or synthetic.  The discriminator first learns features representative of seismic signals, including both earthquake and non-earthquake events, and further provides critics based on the features learned. The design of our discriminator includes two sequential modules: ``feature extraction'' and ``sample critic''. The feature extraction module learns a feature vector that efficiently characterizes the waveforms. The feature vector is then passed onto the sample critic module for evaluation.

Based on conditional GAN, our discriminator receives two inputs: the sample and the label information. In particular, the sample and label come in as data pair, either $(x,~y)$ for real data, or $(\hat{x},~\hat{y})$ for synthetic data. 

In the feature extraction module, we characterize the seismic time series by first computing the frequency domain representation $\{X_k\}:=X_0,X_1,...,X_{N-1}$  of the temporal signal $\{x_n\}:=x_0,x_1,...,x_{N-1}$ by
\begin{equation}
    X_k = \sum^{N-1}_{n=0}x_n\cdot e^{-\frac{2\pi i}{N}kn}.
    \label{eq:DFT_1}
\end{equation}
Denoting the discrete Fourier transform (DFT) as $\mathbb{F}$, Eq.~\eqref{eq:DFT_1} can be written as 
\begin{equation}
    X = \mathbb{F}(x).
    \label{eq: DFT_2}
\end{equation}
With the full spectrum of $X$ obtained in Eq.~\eqref{eq: DFT_2}, we further decompose it into a low-frequency component, $X_\mathrm{low}$, and a high frequency component, $X_\mathrm{high}$, using a learnable cutoff threshold of $T$,
\begin{equation}
X_{\mathrm{low}} = 
\left\{
    \begin{array}{lr}
    X_k, & k \leq T \\
    0, & k > T
    \end{array}
\right.
\label{eq: low_freq_component}
\end{equation}
and
\begin{equation}
X_{\mathrm{high}} = 
\left\{
    \begin{array}{lr}
    0, & k \leq T \\
    X_k, & k > T
    \end{array}
\right.
\label{eq: high_freq_component}
\end{equation}

The basic motivation behind this frequency domain decomposition is that the earthquake and non-earthquake events are known to be characterized by different frequency content, and thus incorporating this physical intuition and domain knowledge directly into our model can be advantageous. The corresponding filtered low- and high-frequency signal in time domain can be calculated by
\begin{equation}
    x_\mathrm{low} = \mathbb{F}^{-1}(X_\mathrm{low}),
    \label{eq:low_temporal}
\end{equation}
and
\begin{equation}
    x_\mathrm{high} = \mathbb{F}^{-1}(X_\mathrm{high}),
    \label{eq:high_temporal}
\end{equation}
where $\mathbb{F}^{-1}$ represents the operator of inverse discrete Fourier transform (IDFT) and $x_\mathrm{low}$ and $x_\mathrm{high}$ means the filtered signal in time domain. 

The hyper-parameter of $T$ plays an important role in separating earthquake events from non-earthquake events.  An inappropriate selection of $T$ may confuse the discriminator in learning the feature representations of earthquake and non-earthquake waveforms. In this work, the hyper-parameter of $T$ is a learnable parameter, meaning that instead of using pre-determined fixed value, we use the training data to obtain an appropriate value through learning. The benefit of using a learnable parameter comparing to a pre-determined value is its adapability to small earthquake events, which could be challenging to separate from background noise. A more detailed discussion of this adaptive filtering techniques can be found in our recent work in \citeA{zhang2019adaptive}.

We next pass $x_{\mathrm{low}}$ and $x_{\mathrm{high}}$ obtained through Eqs.~\eqref{eq:low_temporal} and \eqref{eq:high_temporal} through two identical pipelines. Each pipeline consists of two convolution layers. As shown in Figure~\ref{fig:discriminator},  another input to the discriminator is a binary label, $y$. Similar to the generator, we firstly augment $y$ to be a vector of dimension $1 \times 800$ with a linear layer. To match the dimension of feature vector from sample, we further enlarge the $1 \times 800$ vector to be dimension of $32 \times 800$ with a convolution layer.  With three feature vectors learned from  $x_{\mathrm{low}}$, $x_{\mathrm{high}}$, and $y$, we combine them to obtain a feature vector of dimension $96 \times 800$. 

In the sample critic module, the discriminator uses the output vector from the feature extraction module to determine the quality of the input data. Specifically, we design a network of three convolutional layers with stride $3$ and followed by a mean operator as illustrated in Figure~\ref{fig:discriminator}. The output of the discriminator is a scalar value $s$, which can be any positive real number, with a higher values indicating higher quality (i.e, more realistic and appropriately labeled) input data pairs. 

\subsubsection{Value Function}

An improved  value function of Wasserstein GAN has been developed and shown to be effective in providing a more stable convergence during training \cite{gulrajani2017improved}. We therefore apply similar value function to our problem. In particular, the value function of generator and discriminator can be written as
\begin{equation} \label{eq:loss g}
L_{g} = -\underset{z \sim \mathcal{N}(0,1)}{\mathbb{E}}D(G(z)),
\end{equation}
and
\begin{equation} \label{eq:loss d}
L_{d} = \underset{z \sim \mathcal{N}(0,1)}{\mathbb{E}}D(G(z)) - \underset{x \sim \mathbb{P}_{r}}{\mathbb{E}}D(x) + \lambda\underset{z \sim \mathcal{N}(0,1)}{\mathbb{E}}[(\Vert{D(G(z))}\Vert_{2} - 1)^{2}],
\end{equation}
where $G(\cdot)$ represents the generator, and $D(\cdot)$ represents the discriminator. $\mathbb{P}_{r}$ represents the distribution of real samples.  $z$ represents a Gaussian noise vector. $\lambda$ is a hyper-parameter, that is set to be $10$ in our experiments according to \citeA{gulrajani2017improved}.

\section{Experiment}
\label{sec:experiment}

In this section, we design four tests to validate the performance of our generative model. In Test~1, we first provide a performance comparison of our model versus baseline models via visualization of the synthetic samples. In Test~2, we further evaluate the quality of our synthetic samples via a classification task. In Test ~3, we further study the robustness of our model under limited training sets.  Finally, in Test~, we apply our generative model on a data augmentation task.

\subsection{Test 1: Synthetic Earthquake Evaluation via Visualization}
\label{sec:visual evaluation}

In this test, we visually verify the synthetic results of our model and the baseline models. Visual similarity between synthetic and real waveforms is an important first test of our model, as traditional earthquake detection and classification techniques hinge on visual appearance. However, visual similarity is not by itself a sufficient metric to judge the quality of our model, and hence we dig deeper in the sections that follow. All the generative models in this section are trained on the full dataset from V35A, which contains $6,432$ real samples with positive versus negative ratio as 1 : 1. 

\subsubsection{Visual Appearance}

\begin{figure}[ht]
\centering{
\subfigure[]{\includegraphics[width=0.30\columnwidth]{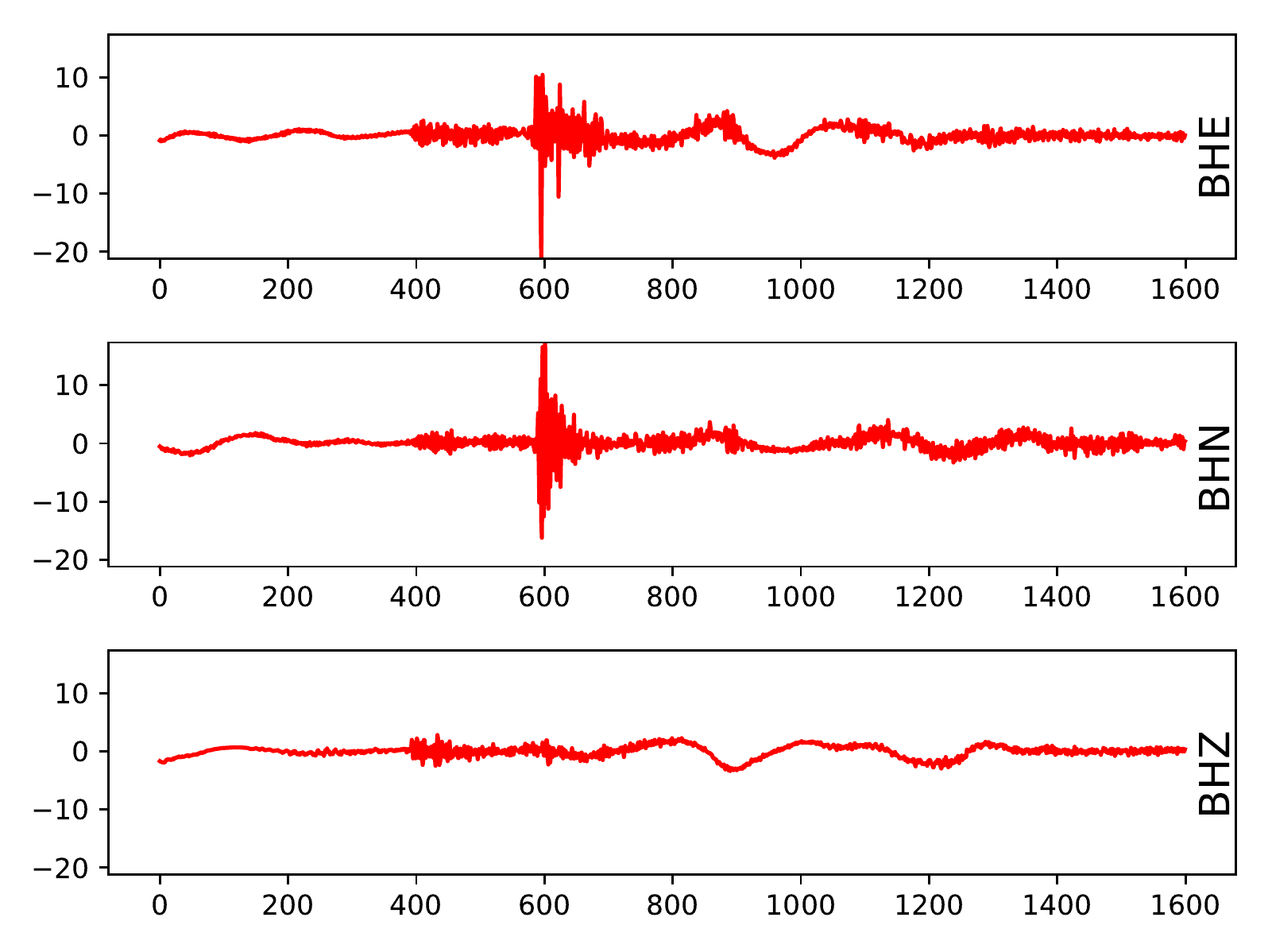}}
\subfigure[]{\includegraphics[width=0.30\columnwidth]{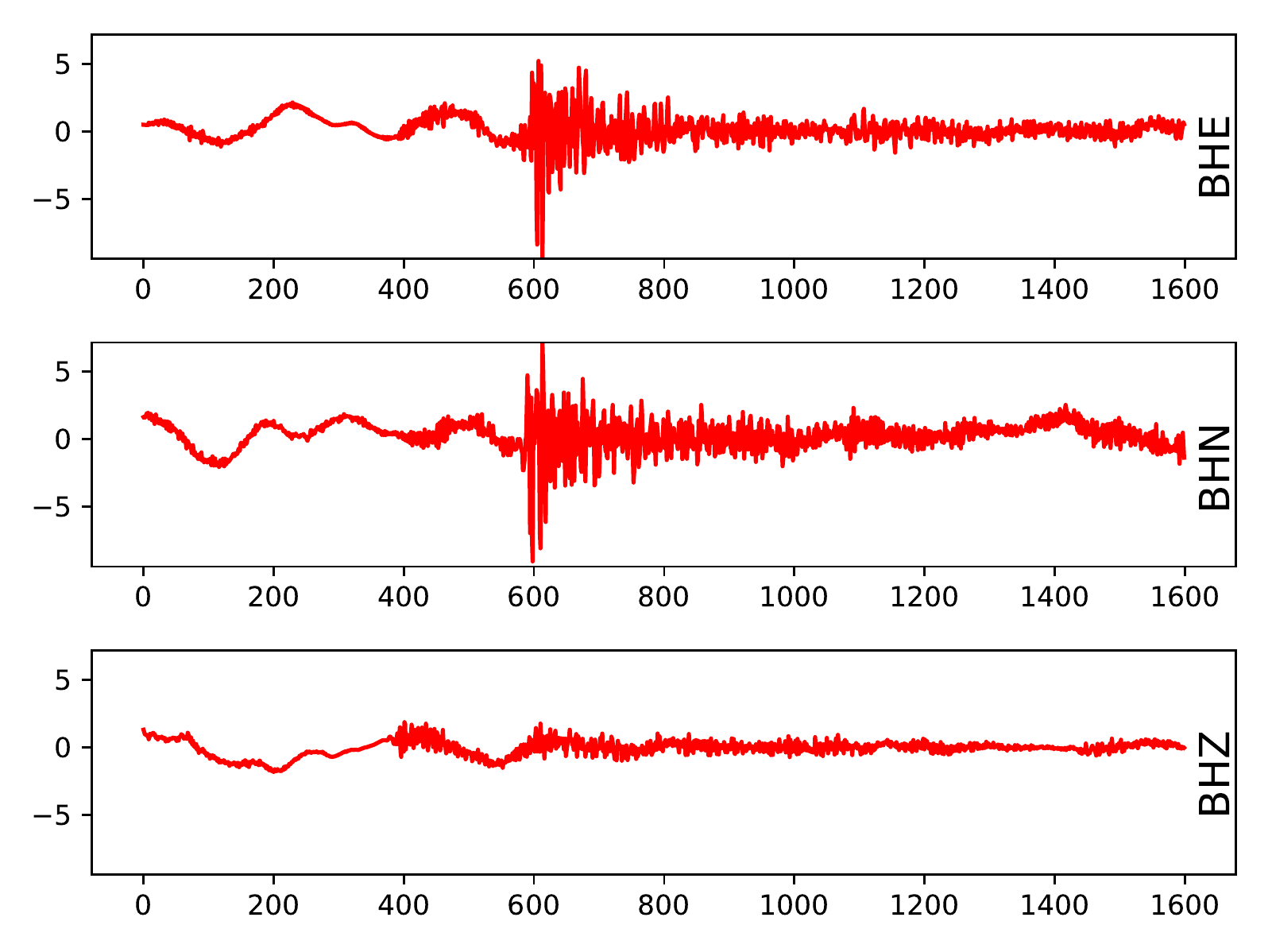}}
\subfigure[]{\includegraphics[width=0.30\columnwidth]{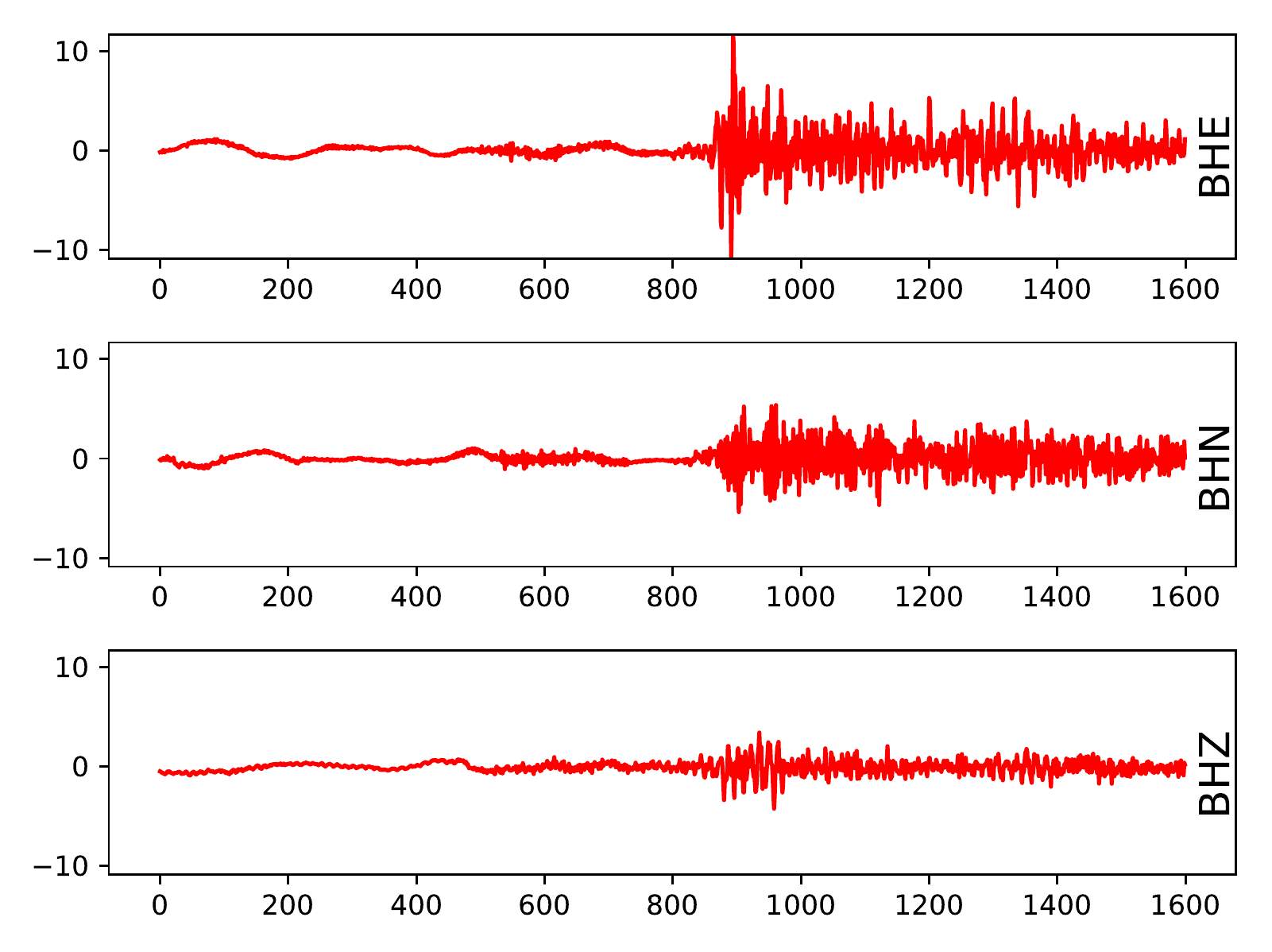}}}
\centering{
\subfigure[]{\includegraphics[width=0.30\columnwidth]{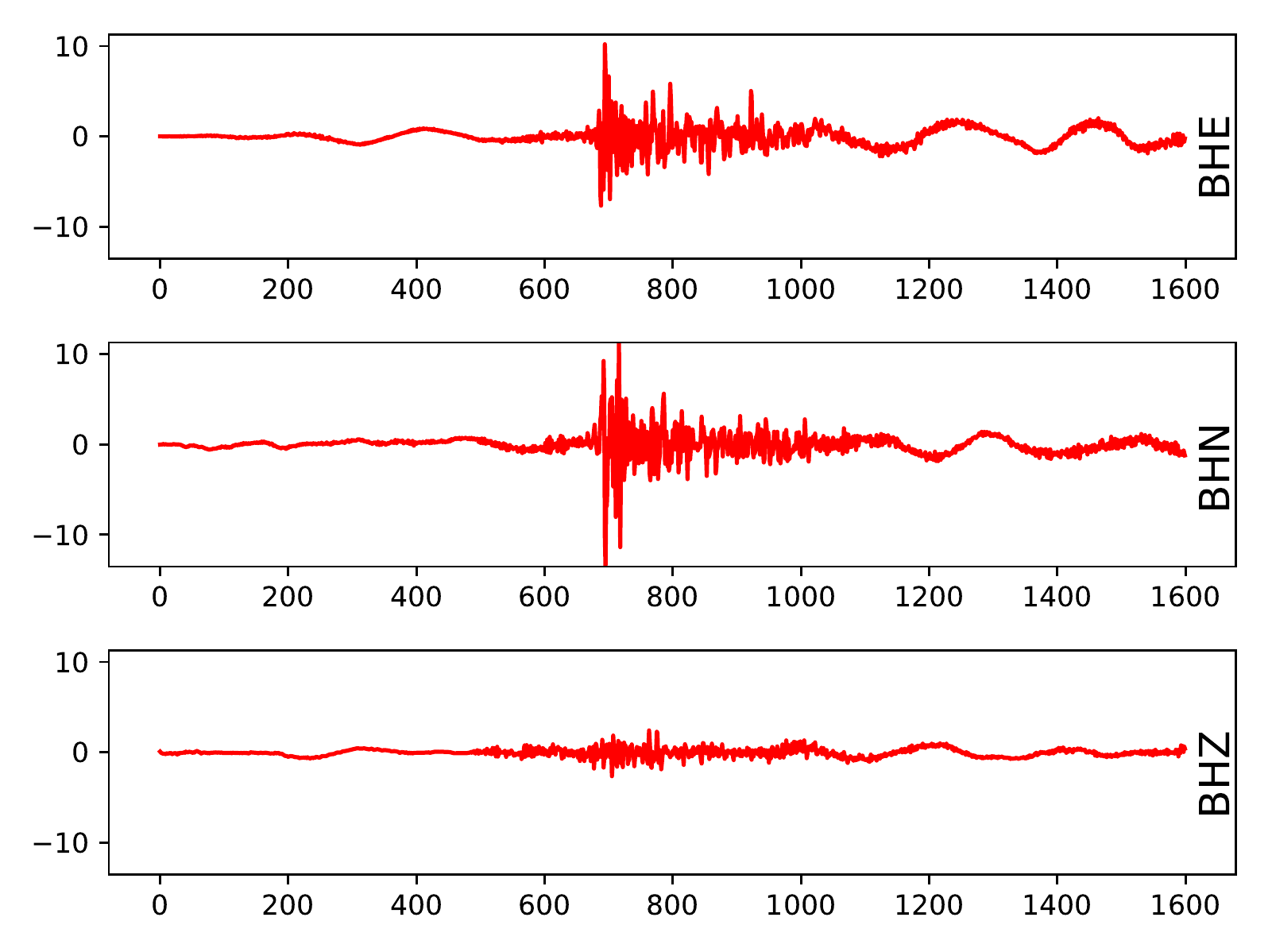}}
\subfigure[]{\includegraphics[width=0.30\columnwidth]{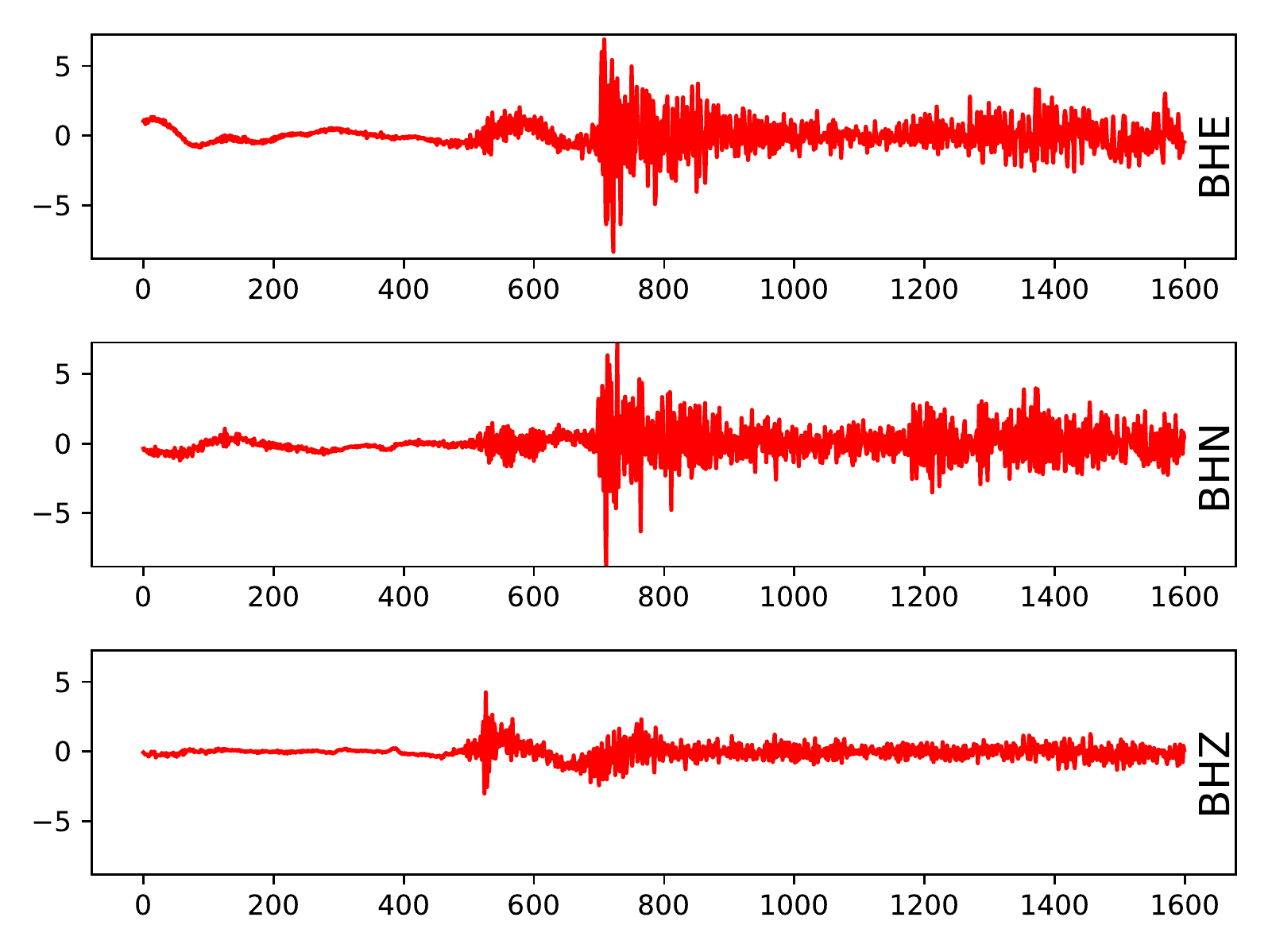}}
\subfigure[]{\includegraphics[width=0.30\columnwidth]{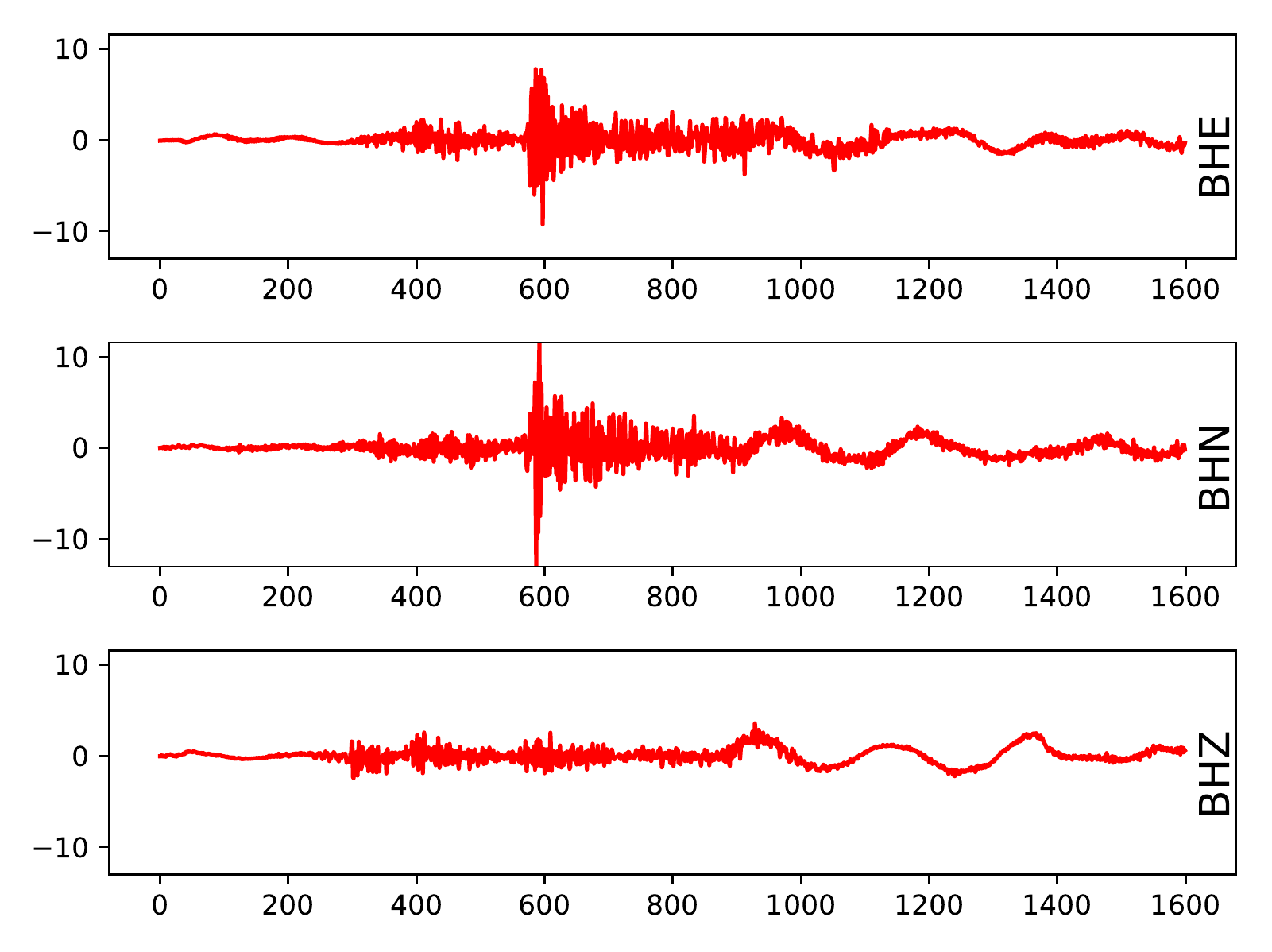}}}
\caption{Illustration of six synthetic raw positive samples generated by our model.}
\label{fig:generatedsample_pos_raw}
\end{figure}

\begin{figure}[ht]
\centering{
\subfigure[]{\includegraphics[width=0.30\columnwidth]{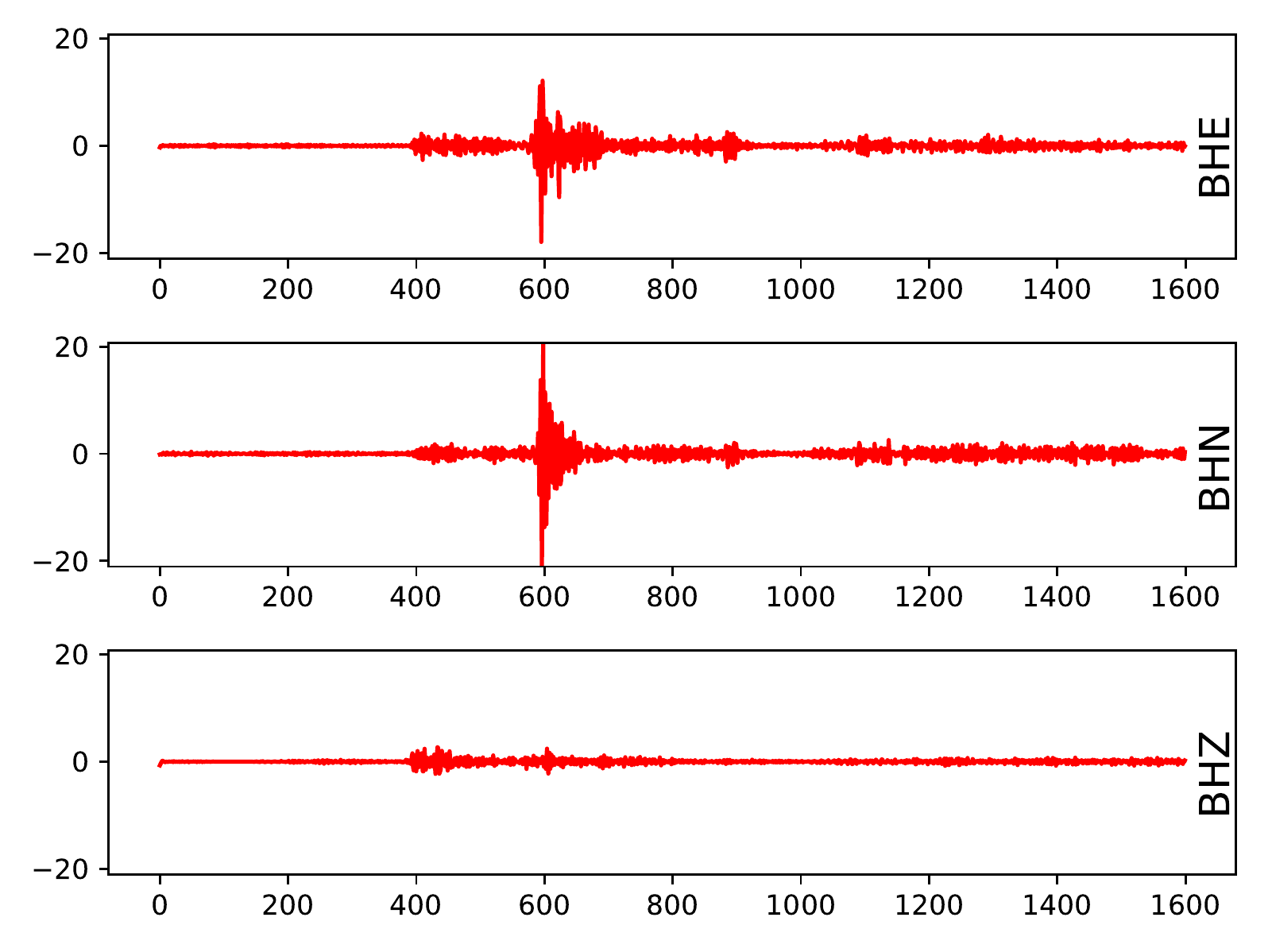}}
\subfigure[]{\includegraphics[width=0.30\columnwidth]{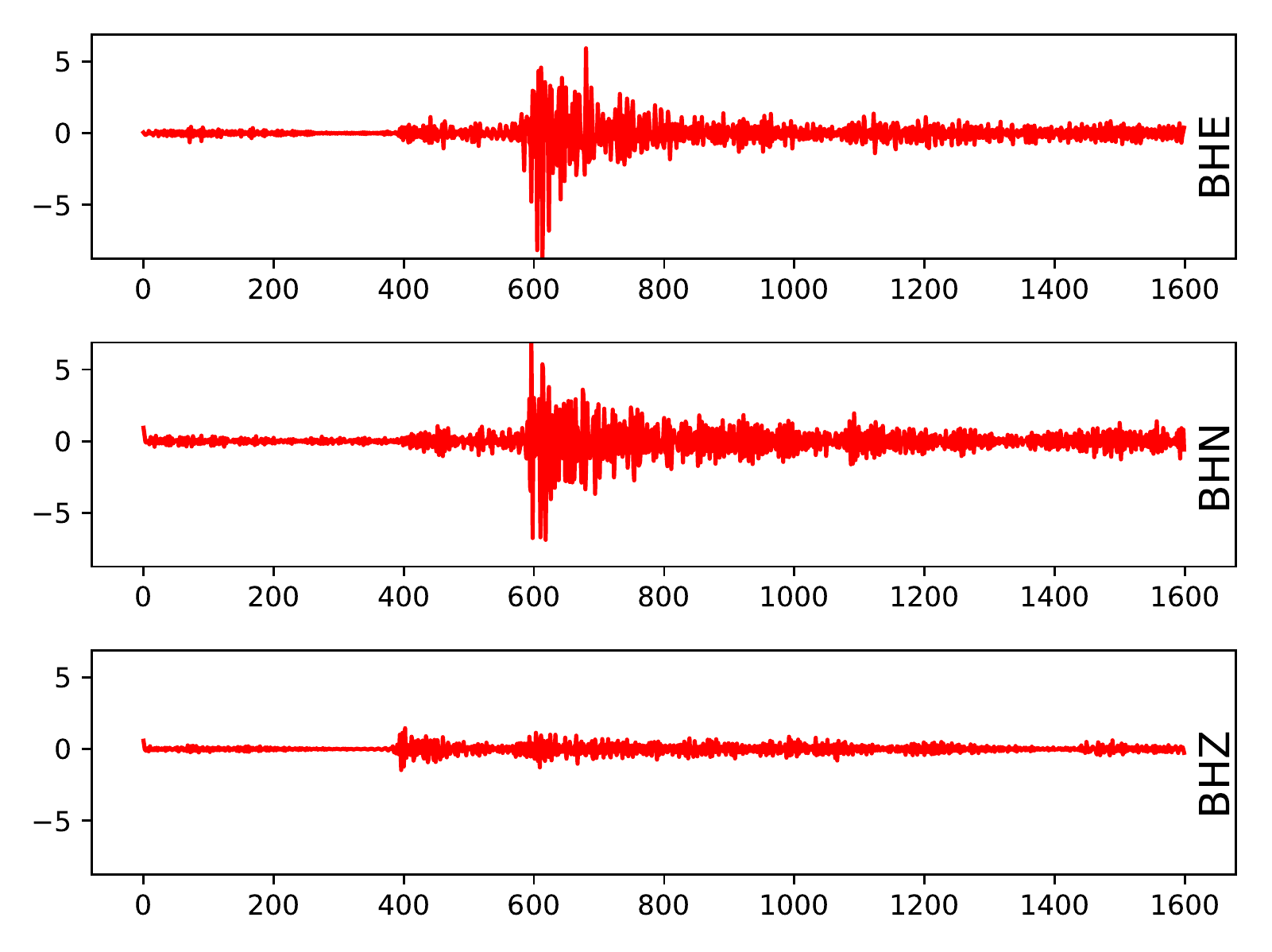}}
\subfigure[]{\includegraphics[width=0.30\columnwidth]{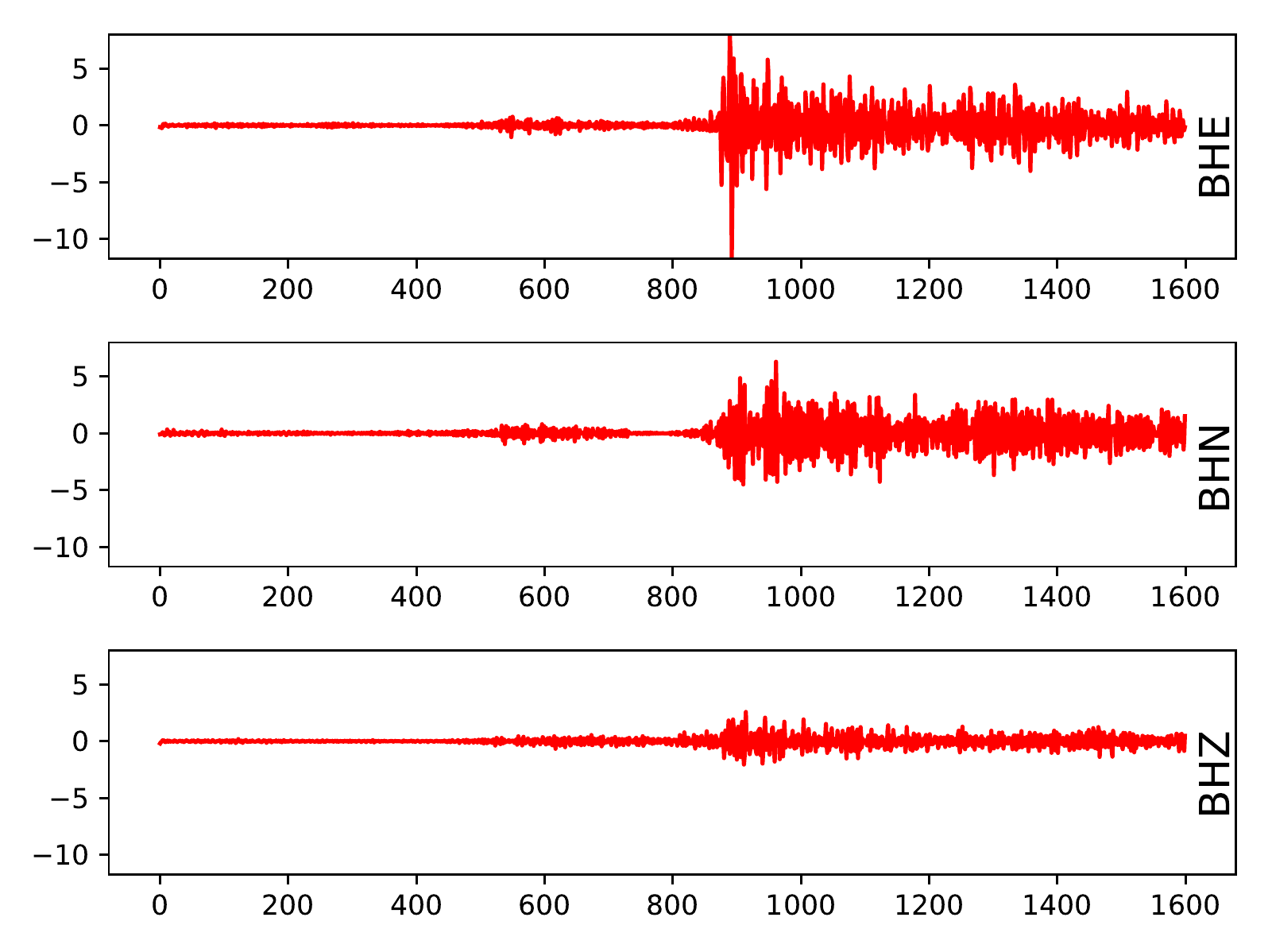}}}
\centering{
\subfigure[]{\includegraphics[width=0.30\columnwidth]{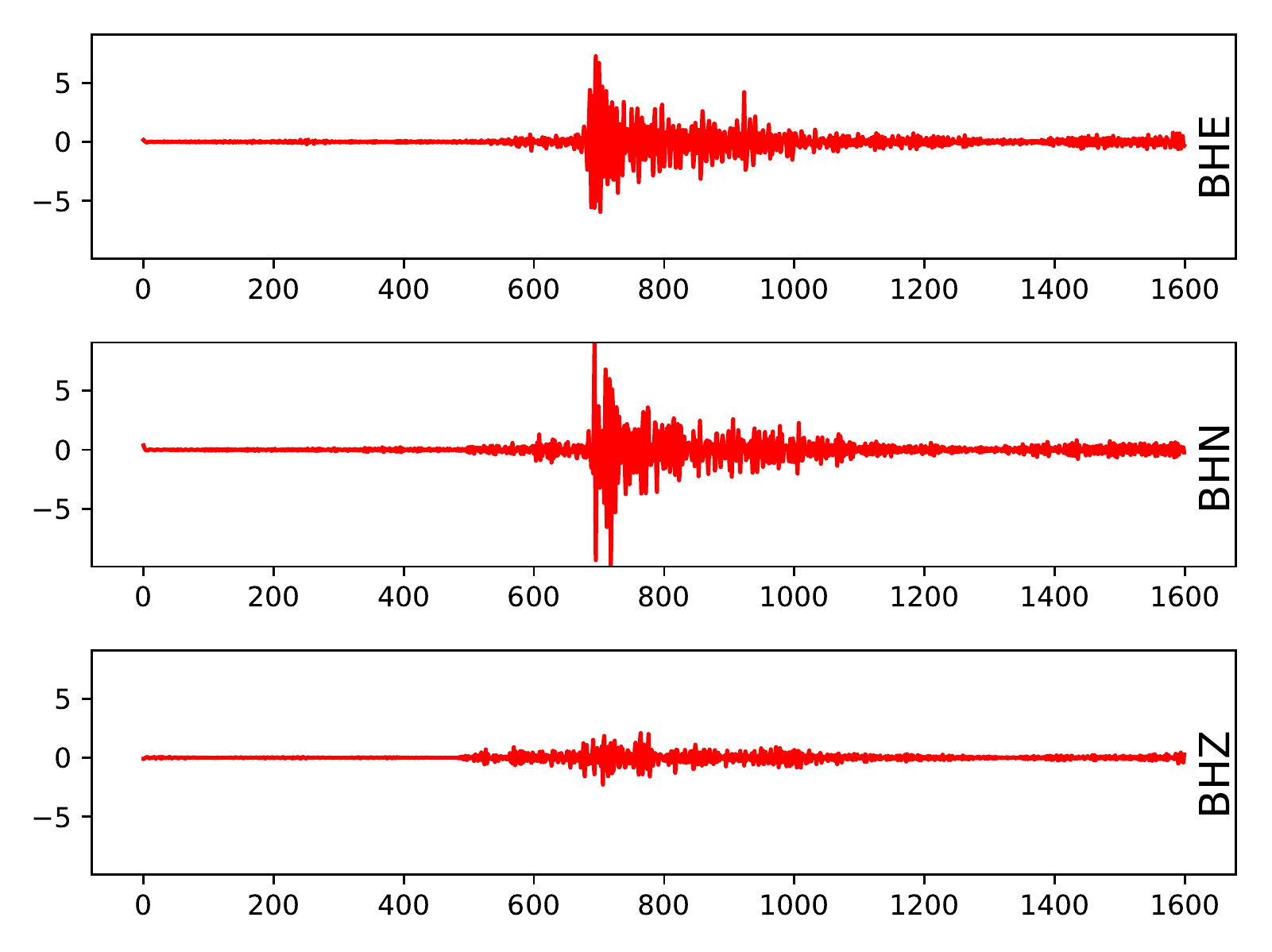}}
\subfigure[]{\includegraphics[width=0.30\columnwidth]{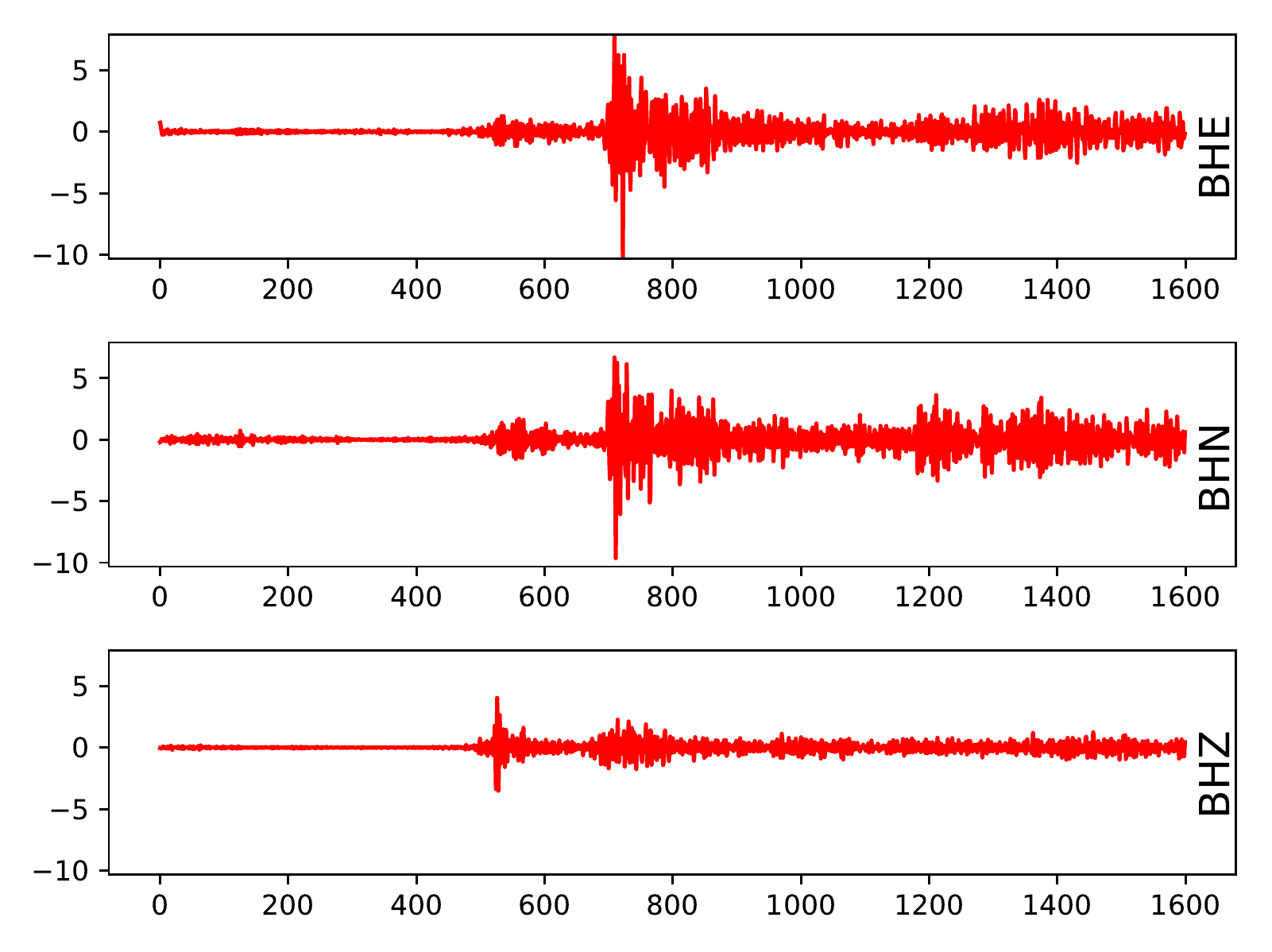}}
\subfigure[]{\includegraphics[width=0.30\columnwidth]{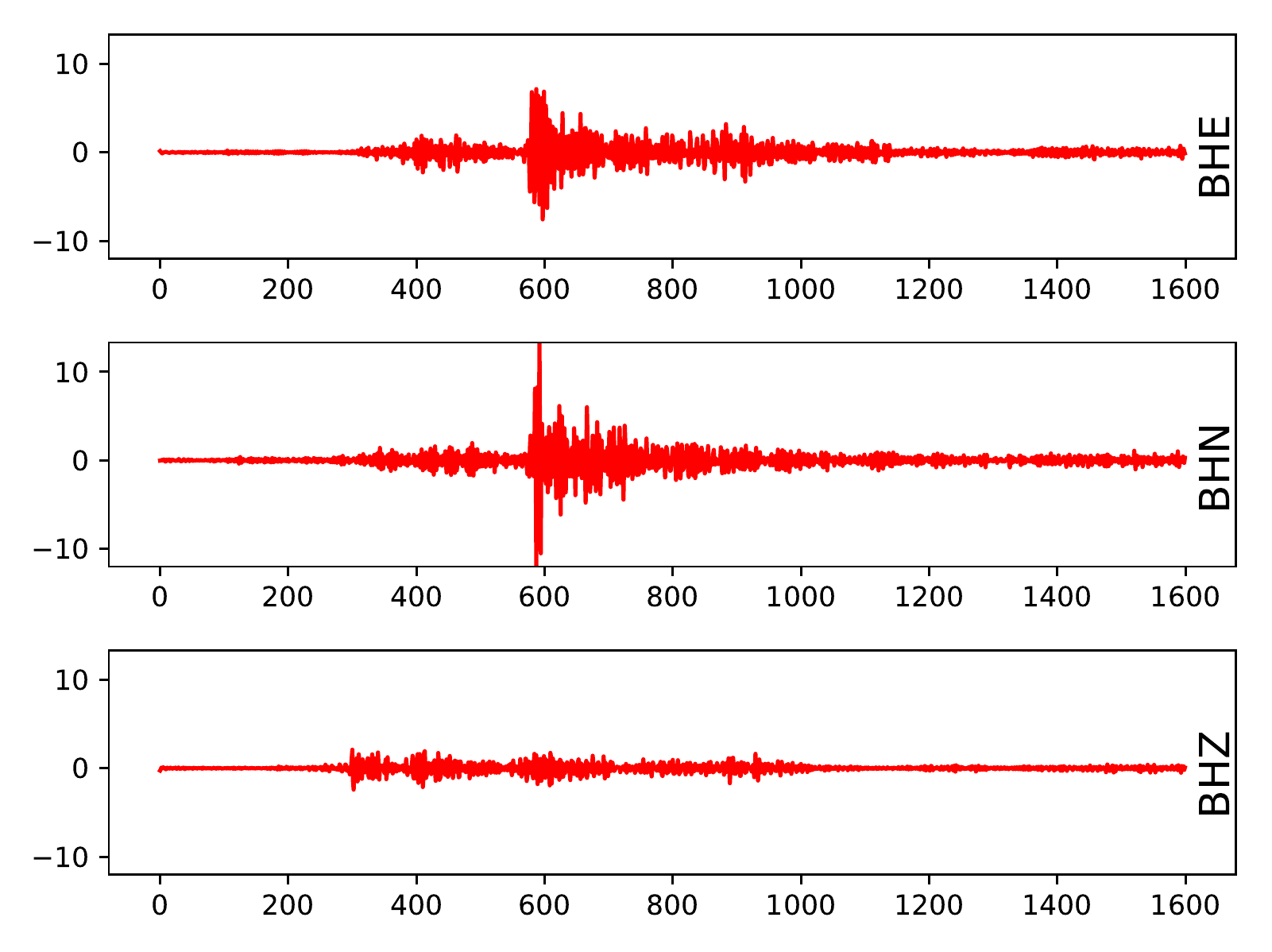}}}
\caption{Illustration of six synthetic filtered positive samples generated by our model.}
\label{fig:generatedsample_pos_filtered}
\end{figure}

\begin{figure}[ht]
\centering{
\subfigure[]{\includegraphics[width=0.30\columnwidth]{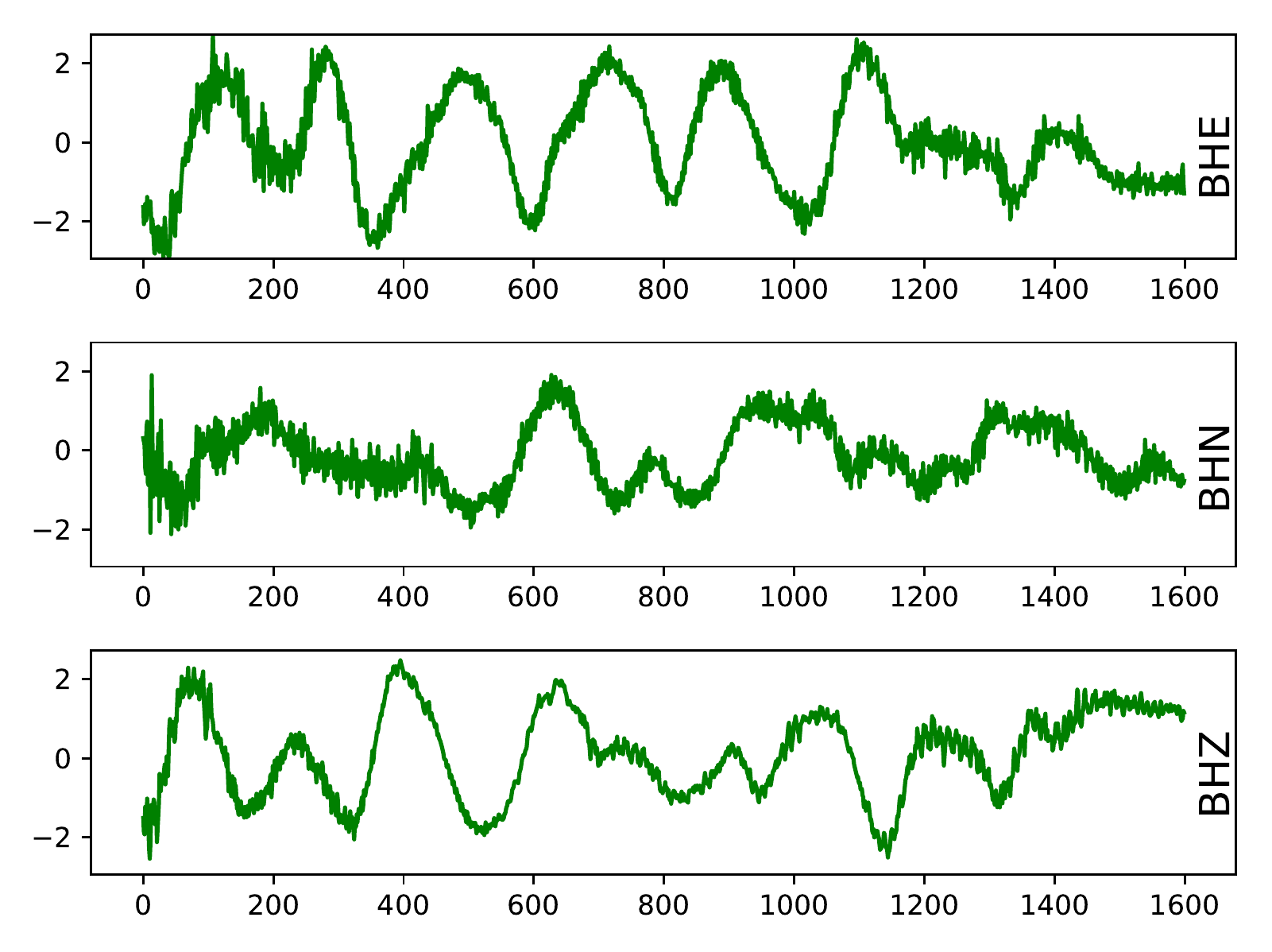}}
\subfigure[]{\includegraphics[width=0.30\columnwidth]{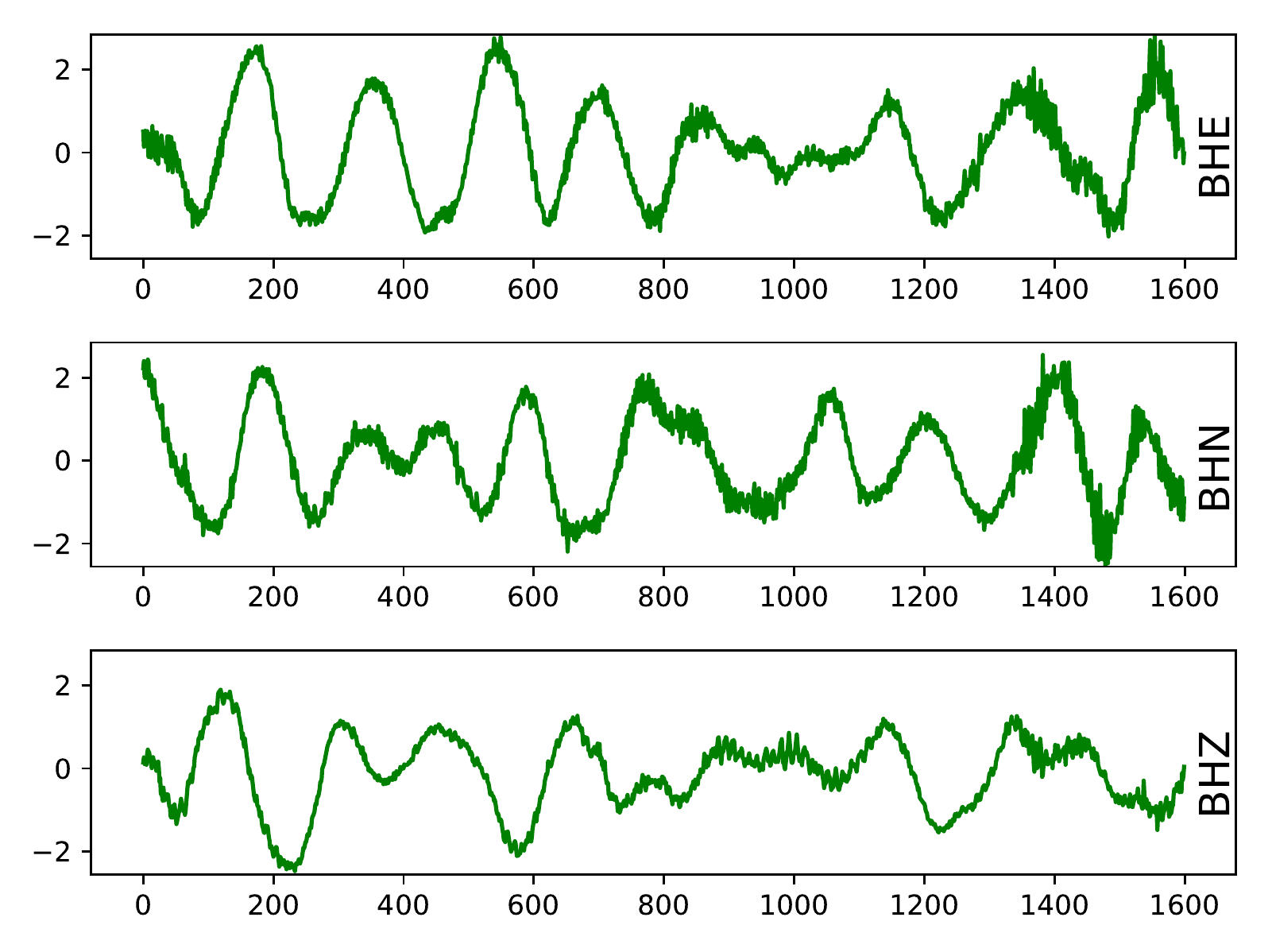}}
\subfigure[]{\includegraphics[width=0.30\columnwidth]{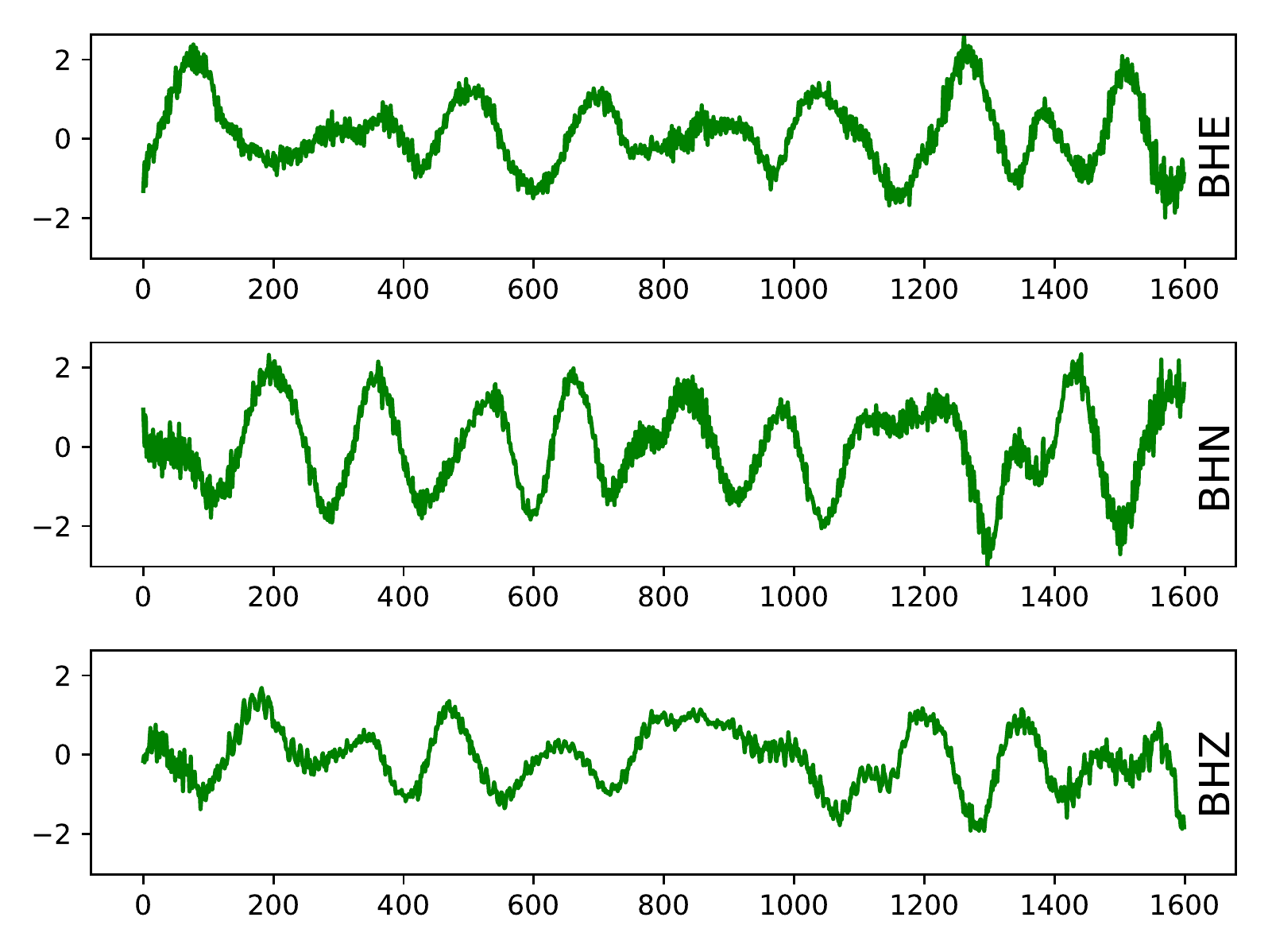}}}
\centering{
\subfigure[]{\includegraphics[width=0.30\columnwidth]{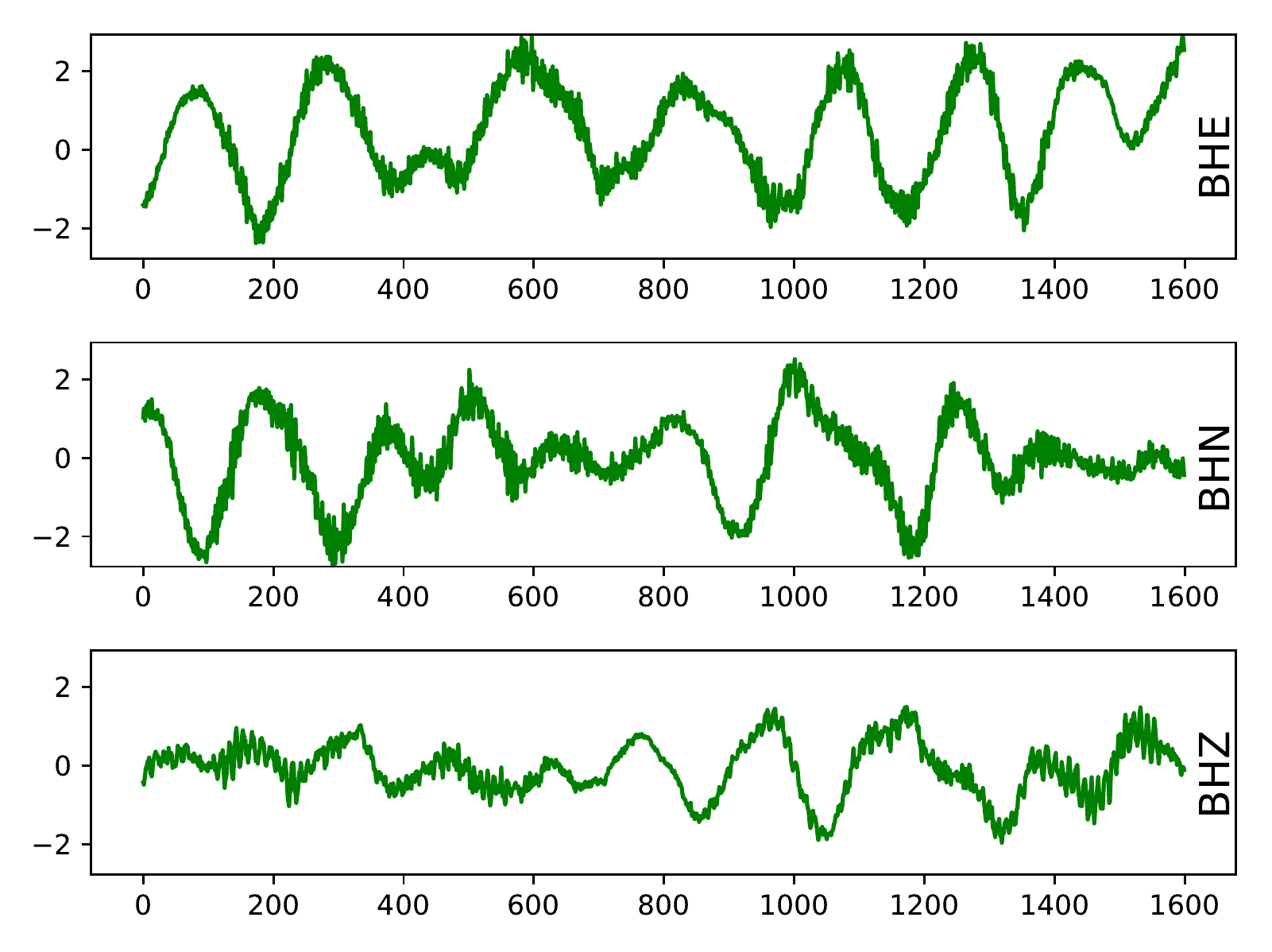}}
\subfigure[]{\includegraphics[width=0.30\columnwidth]{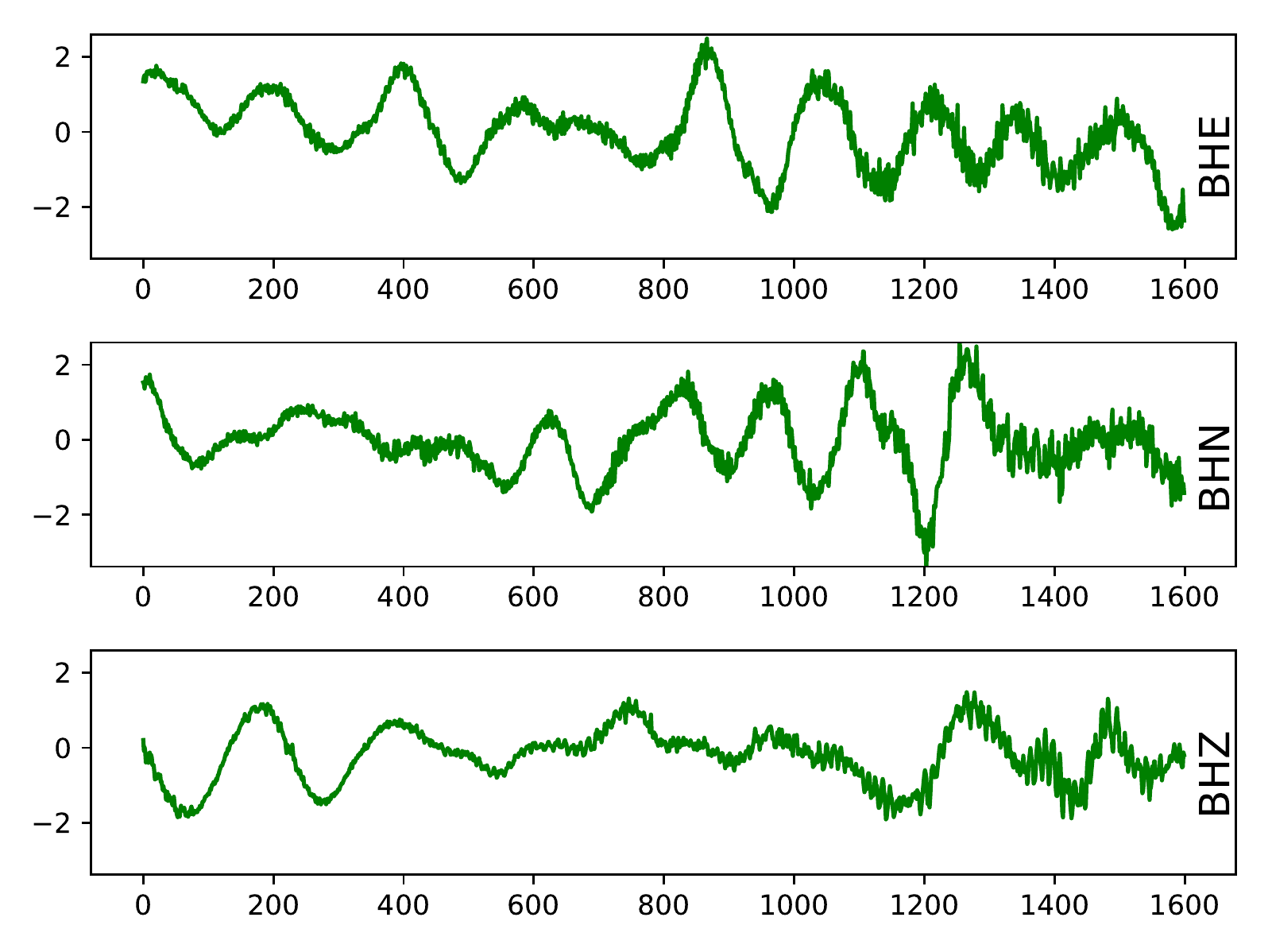}}
\subfigure[]{\includegraphics[width=0.30\columnwidth]{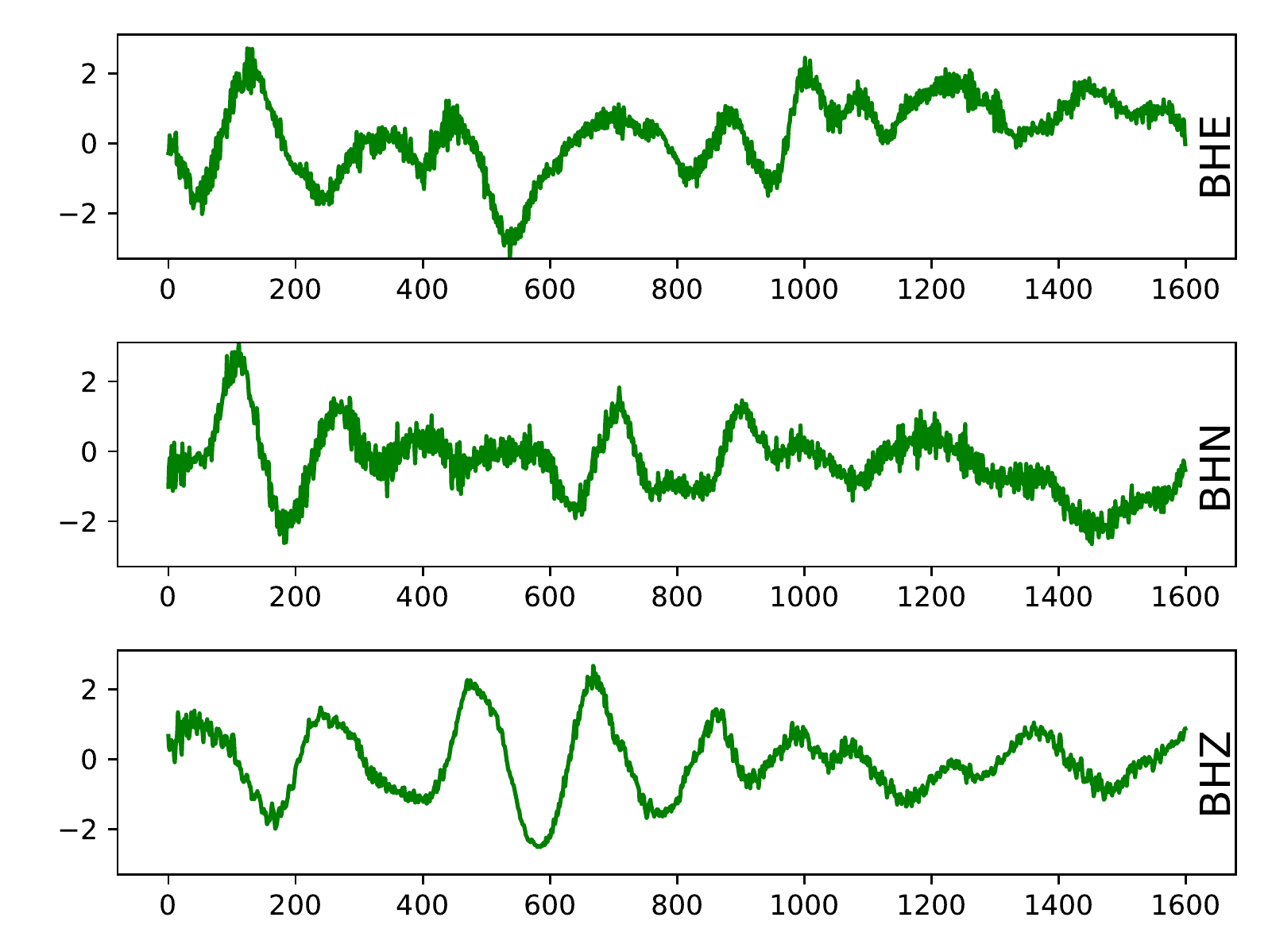}}}
\caption{Illustration of six synthetic raw negative samples generated by our model.}
\label{fig:generatedsample_neg_raw}
\end{figure}

\begin{figure}[ht]
\centering{
\subfigure[]{\includegraphics[width=0.30\columnwidth]{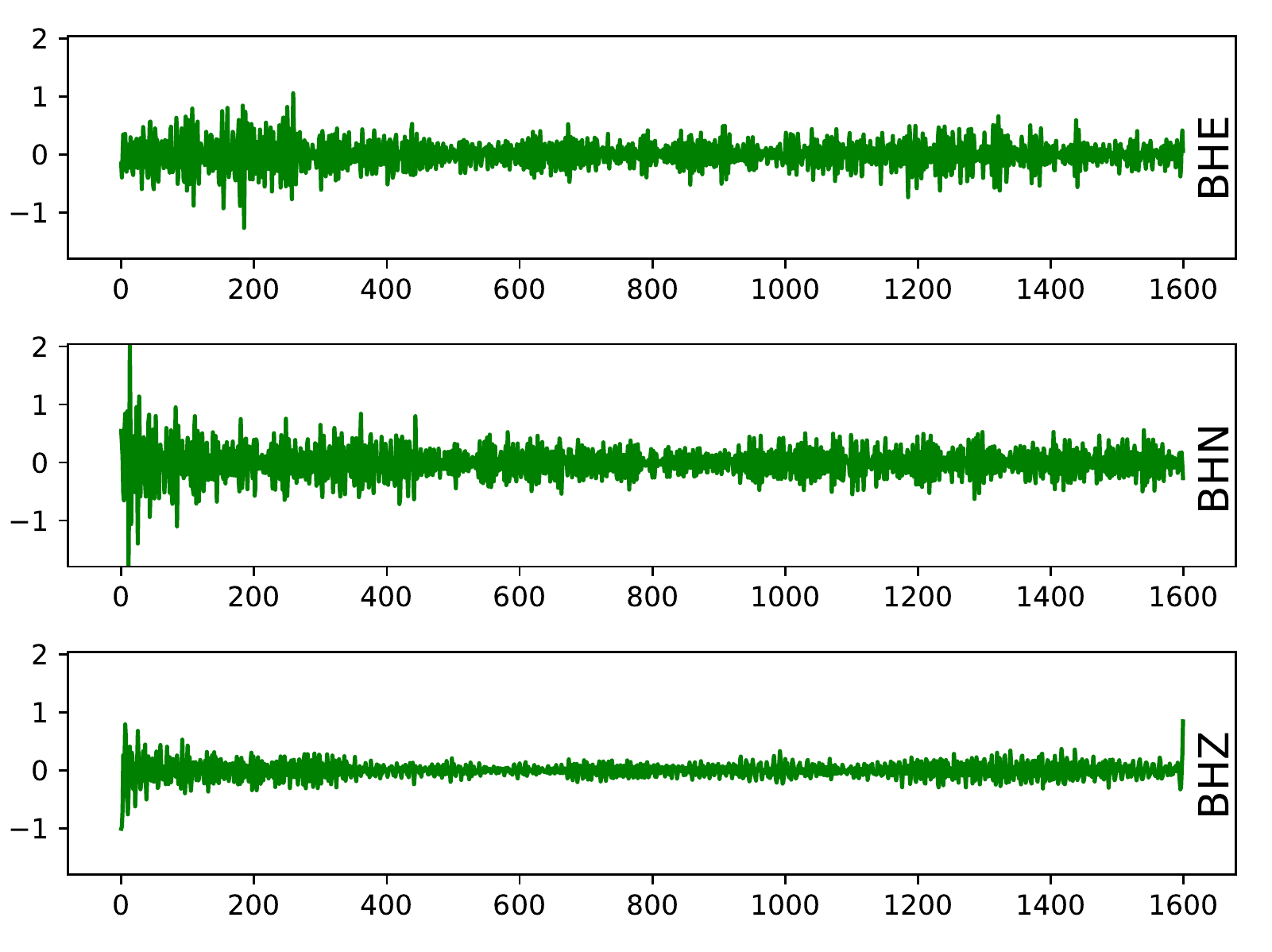}}
\subfigure[]{\includegraphics[width=0.30\columnwidth]{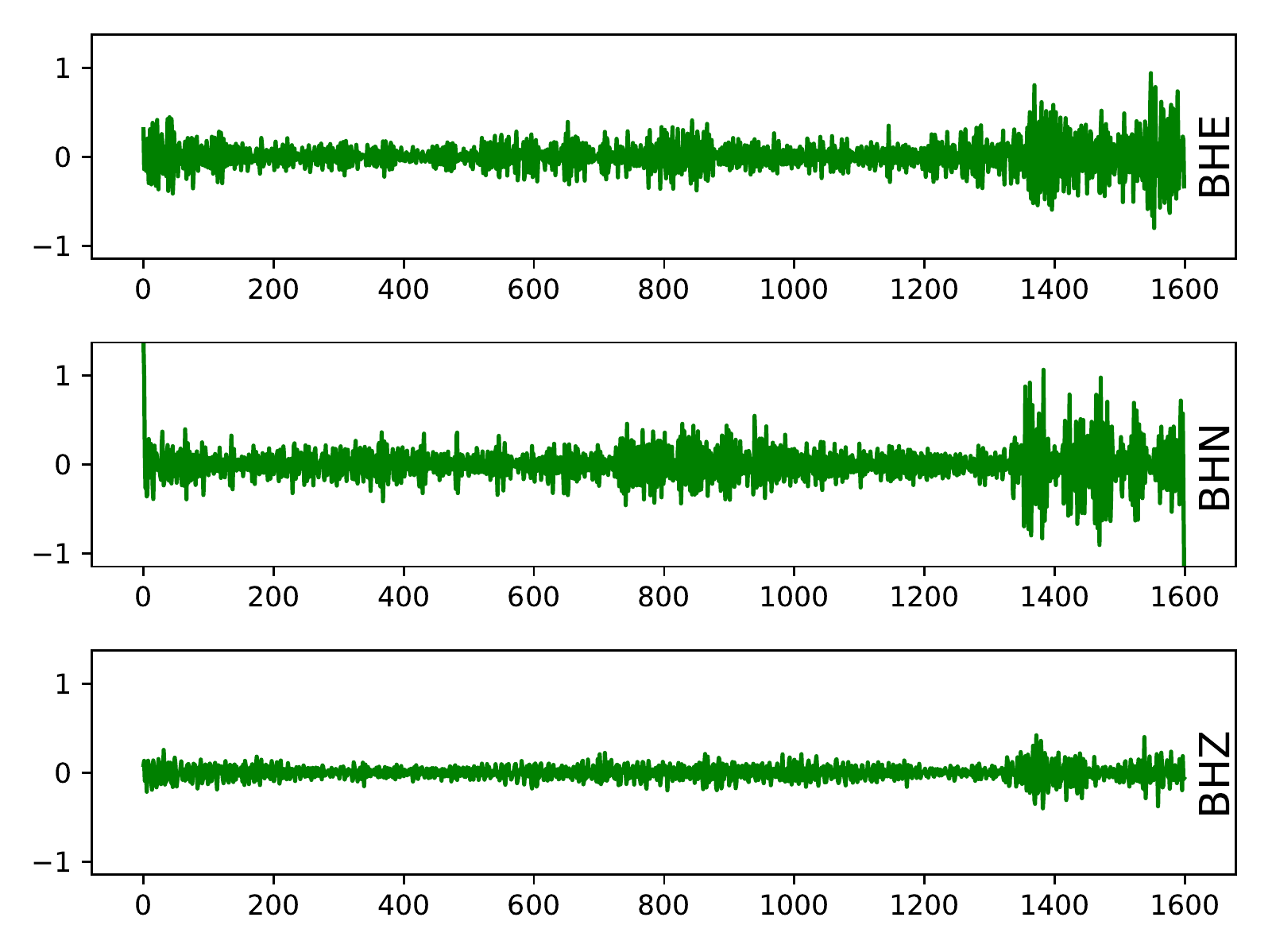}}
\subfigure[]{\includegraphics[width=0.30\columnwidth]{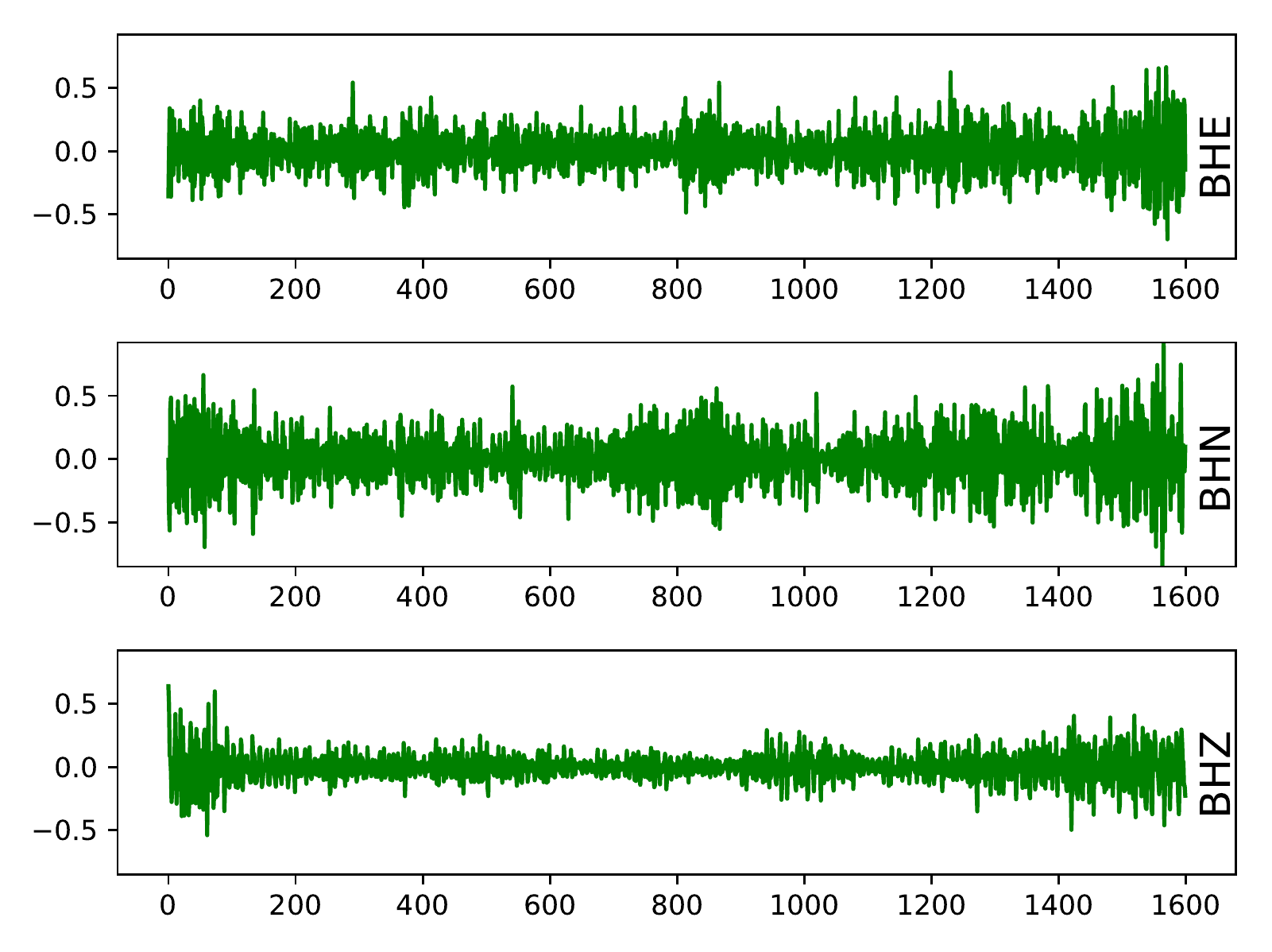}}}
\centering{
\subfigure[]{\includegraphics[width=0.30\columnwidth]{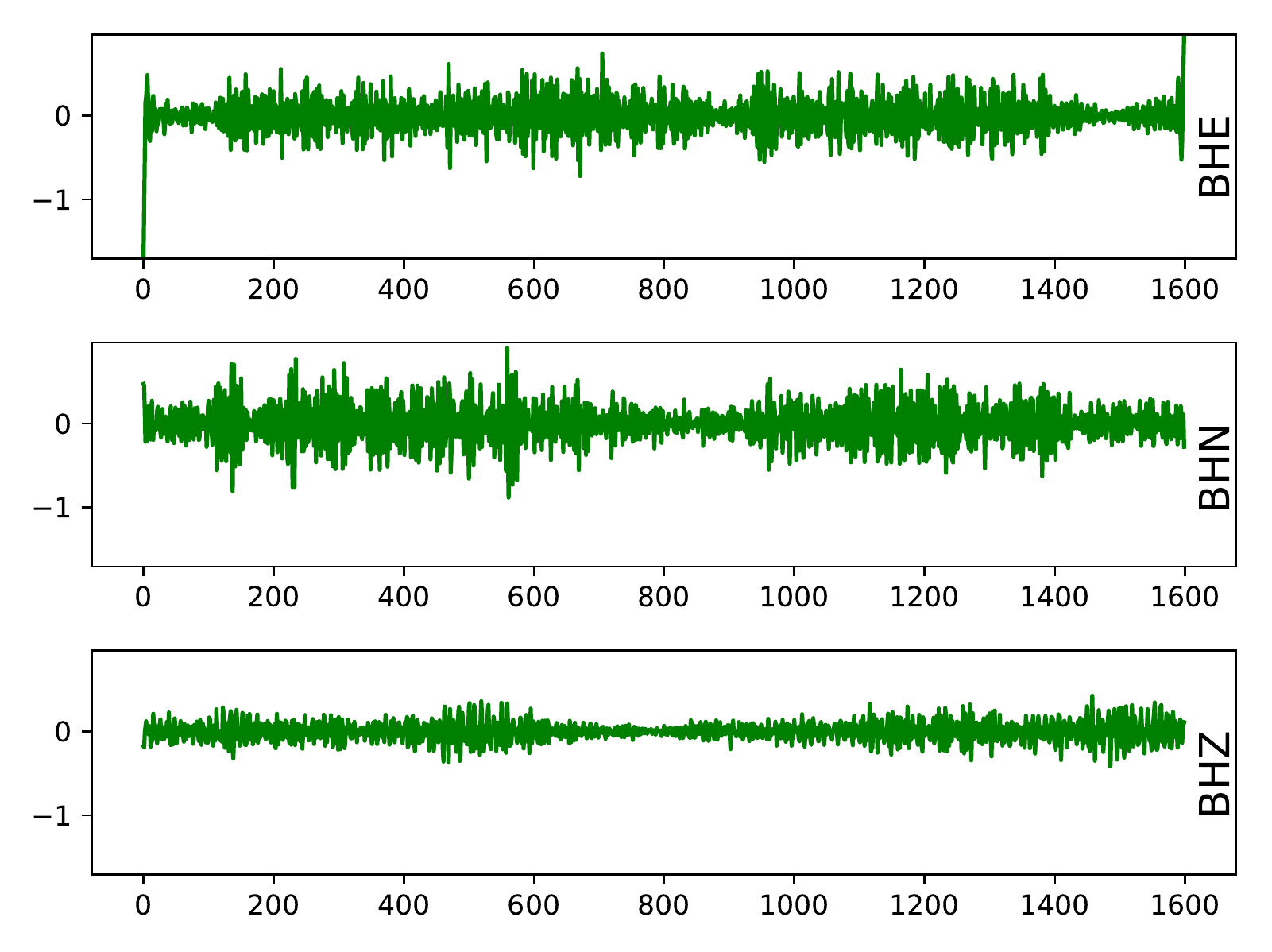}}
\subfigure[]{\includegraphics[width=0.30\columnwidth]{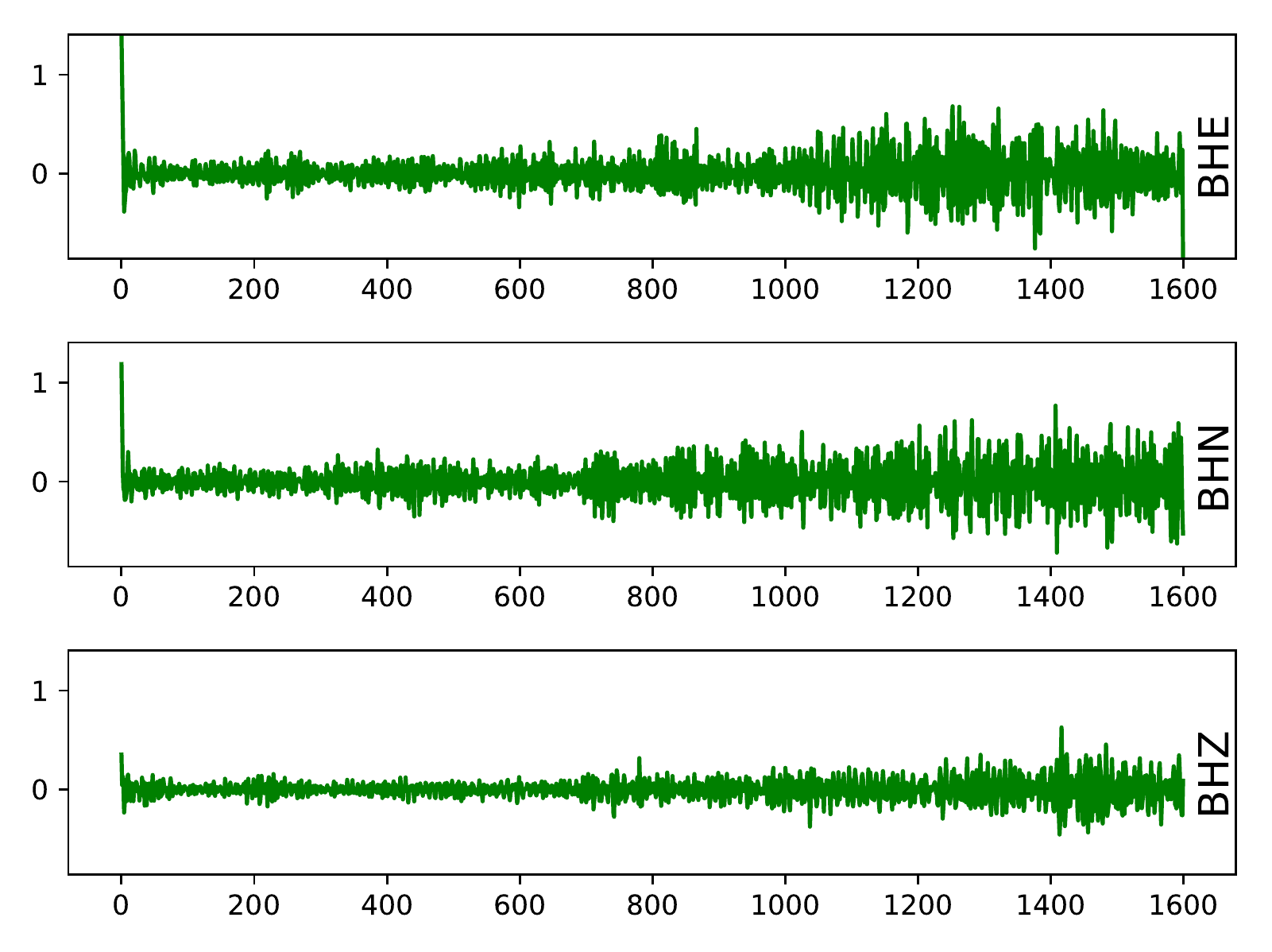}}
\subfigure[]{\includegraphics[width=0.30\columnwidth]{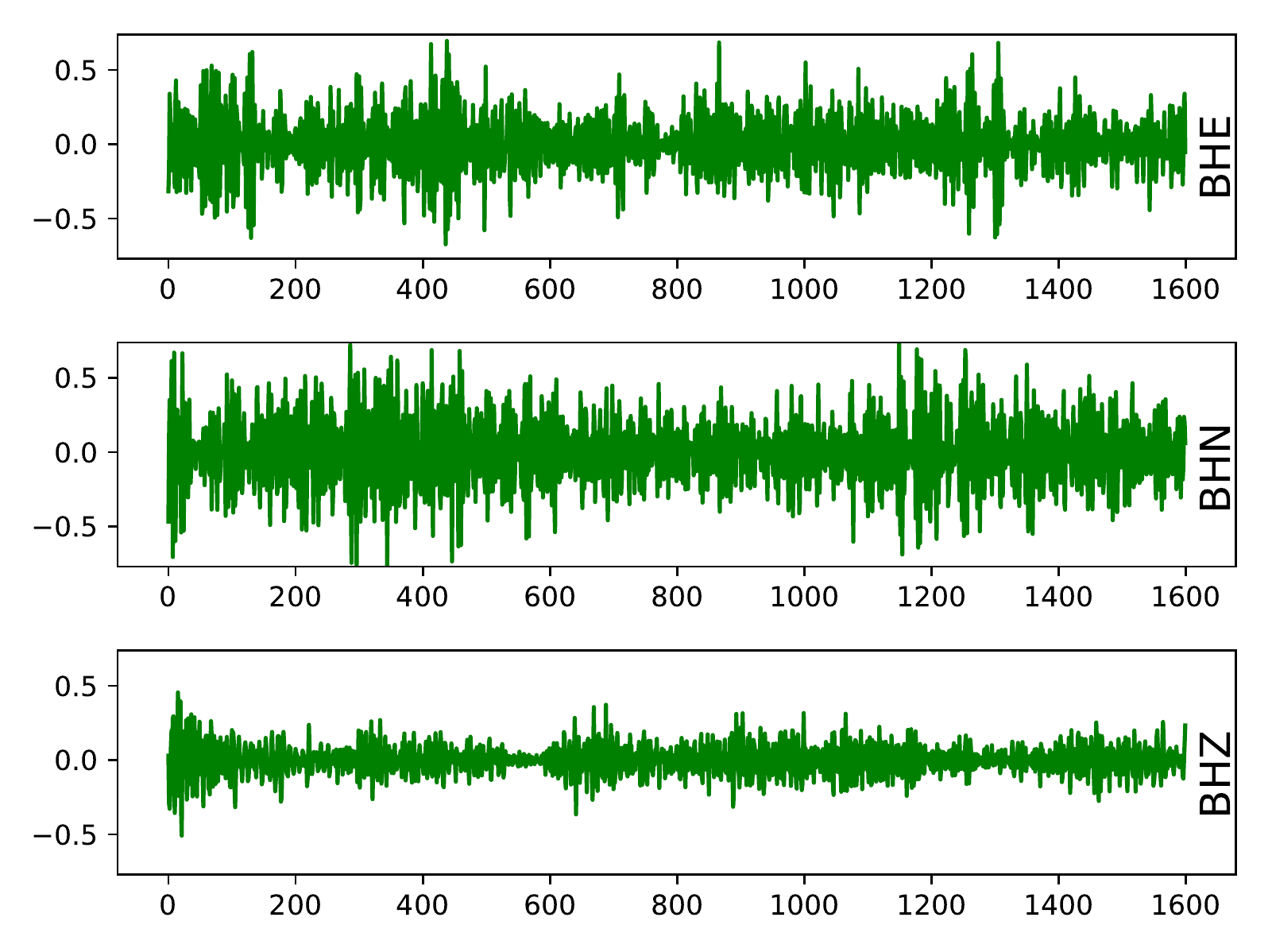}}}
\caption{Illustration of six synthetic filtered negative samples generated by our model.}
\label{fig:generatedsample_neg_filtered}
\end{figure}

Figures~\ref{fig:generatedsample_pos_raw} and \ref{fig:generatedsample_pos_filtered} show synthetic data generated through by our GAN model in raw and filtered form. The positive synthetic samples share similar characteristics to those of the real positive samples in Figure ~\ref{fig:real sample Pos}. While P-wave and S-wave arrivals are apparent on all three channels, the later arriving S-wave is larger in amplitude, especially on the BHE and BHN channels. Coda waves that extend the wavetrain after the direct arrivals are also visible. We also provide five examples negative sythentic waveforms in (raw and filtered) in Figures ~\ref{fig:generatedsample_neg_raw} and ~\ref{fig:generatedsample_neg_filtered}. Comparing to the real negative samples shown in Figure ~\ref{fig:real sample Neg}, these synthesized negative waveforms are highly similar from their visual appearance.

\subsubsection{Comparison Study}
\label{sec:baseline design}

To validate the effectiveness of our generative model, we provide a comprehensive comparison study to baseline models that vary key aspects of our generative model. The seven different baselines are listed in Table.~\ref{tab:BaselineSummary}. A detailed discussion of each baseline model and its corresponding results are provided below.  

\begin{table}[h!]
\centering
 \begin{tabular}{| c || c | c |} 
 \hline
 Baseline ID & Model & Result  \\
\hline
 Baseline~1 & DCGAN~\cite{DCGANimages}  & Figure~\ref{fig:singlepipe}  in the Supporting Information  \\ 
 \hline
 Baseline~2 & Independent Generator      & Figure~\ref{fig:indeppipe}  in the Supporting Information  \\
  \hline
 Baseline~3 to 6 & Varying Kernel Sizes      & Figures~\ref{fig:gk4} to \ref{fig:dk128}  in the Supporting Information\\
  \hline
   Baseline~7 & Spectrum Decomposition      & Figure~\ref{fig:nohp}  in the Supporting Information \\
  \hline
\end{tabular}
\caption{Summary of baseline methods and the corresponding results.}
\label{tab:BaselineSummary}
\end{table}

\textbullet{~Baseline~1 - Direct Deployment of DCGAN}

Most existing generative models based on GAN are targeting on image synthesis~\cite{DCGAN, DCGANimages}, with comparatively few focusing on applying GAN for generating 1D time series like those of seismic waveform data. As an first baseline test of our model to other, well-established techniques in the literature, we select the widely-used deep convolutional generative adversarial networks~(DCGAN)~\cite{DCGAN} due to its popularity. Here we adapt a network structure similar to the one in \citeA{DCGANimages}, which can be seen as a single pipeline variant of our model. Based on this structure, we provide some synthetic positive and negative sample in Figure~\ref{fig:singlepipe} in the Supporting Information. As shown in the figure, for either positive or negative synthetic samples, the waveforms of all three components become almost identical, which indicates the inappropriateness of the direct application of the DCGAN network structure to earthquake detection problems.  

% Our first baseline test is to compare our results to that generated by a network with a structure similar to the one in \citeA{DCGANimages}, which can be seen as a single pipeline variant of our model. Based on this structure, we provide some synthetic positive and negative sample in Figure~\ref{fig:singlepipe} in the Supporting Information. As shown in the figure, for either positive or negative synthetic samples, the waveforms of all three components become almost identical. 

\textbullet{~Baseline~2 - Independent Generators}

It is important to use a shared input for three pipelines in our generator. To demonstrate this, we design a baseline model by feeding each pipeline with independent input pair of $z$ and label feature vector augmented from $\hat{y}$. We show the corresponding synthetic positive and negative samples in Figure~\ref{fig:indeppipe} in the Supporting Information. The synthetic data on all three components maintains some realistic features but are no longer correlated, both in their temporal structure and in the general characteristics of the wavepacket. For an instance, the arrivals of S wave in BHE and BHN components are not strictly correlated in time. This is due to the fact that the three components are generated by three independent generators and no information is shared among them.  More synthetic waveform samples generated by Baseline~2 are included in Supporting Information.

\textbullet{~Baseline~3, 4, 5 and 6 - Kernel Size}

Kernel size can be an important hyperparameter in the design of network structures~\cite{largeKernelMatters, kernelSizeTip}. We design four baseline models (Baseline~3, 4, 5, and 6) to illustrate its effect on both the generator and discriminator. Specifically, we change the generator kernel size from $128$ to $4$ in baseline~3 and from $128$ to $32$ in Baseline~4, respectively. We show the corresponding positive and negative samples in Figures~\ref{fig:gk4} and~\ref{fig:gk32}, respectively, where it clear from visual inspection that the resulting synthetics no longer resemble real waveforms of event and noise classes. Similarly, we change the discriminator kernel size from $16$ to $4$ in Baseline~5 and from $16$ to $128$ in Baseline~6, respectively. Results are provided in Figures~\ref{fig:dk4} and~\ref{fig:dk128}, respectively, where it becomes apparent that neither of Baseline~3 or 4 are capable of learning effective features to generate earthquake events. Particularly, in Figure ~\ref{fig:dk4}, the abrupt arrivals of P- or S-wave are not generated. In Figures~\ref{fig:dk128}, the high frequency component of positive samples are not realistic comparing those real samples in Figures~\ref{fig:real sample Pos} and \ref{fig:real sample Neg}.

\textbullet{~Baseline~7 - Fourier transform removed}

Signal decomposition can be helpful in producing representative feature vectors in our discriminator. To validate this, we design a Baseline~7. In this baseline model, instead of decomposing temporal signal as in Eqs.~\eqref{eq:low_temporal} and \eqref{eq:high_temporal}, we simply duplicate the input temporal signal, and feed them to the  pipelines, respectively. We provide the synthetic positive and negative samples using Baseline~7 in Figure ~\ref{fig:nohp} in the Supporting Information. Visually, Baseline~7 yields better results than aforementioned six baseline models. However, comparing to the real earthquake events in Figure ~\ref{fig:real sample Pos}, there are still some generated samples which are quite visually distinct. More synthetic waveform samples generated by Baseline~7 are included in Supporting Information.

\subsection{Test 2: Synthetic Earthquake Evaluation via Classification}
\label{sec:numerical evaluation}

Now that we have evaluated the quality of our synthetic samples via visualization, in this test we provide a more quantitative evaluation of our synthetic samples. To do this, we use our conditional GAN model to produce synthetic data, and train an independent classifying algorithm on these synthetics. We employ three widely used classification metrics (accuracy, precision and recall) to evaluate the performance. The definitions of the metrics are provided below
\begin{equation}
\mathrm{Accuracy}=\frac{\mathrm{TP} + \mathrm{TN}}{\mathrm{Total}}, \;\;\; \mathrm{Precision}=\frac{\mathrm{TP}}{\mathrm{TP} + \mathrm{FP}}, \;\;\; \mathrm{Recall}=\frac{\mathrm{TP}}{\mathrm{TP} + \mathrm{FN}},
\end{equation}
where ``TP'', ``TN'', ``FP'' and ``FN'' refer to the numbers of true positive, true negative, false positive and false negative, respectively. ``Total'' refers to the total number of samples in the test set, which is $2150$ in V34A and $2,432$ in V35A. In this test, we examine the classification accuracy of models trained on real data with those trained on synthetic data. In both instances, we use the adaptive filtering classification model due to its demonstrated performance in earthquake detection~\cite{zhang2019adaptive}.

 We use datasets from both V34A and V35A for this test. Specifically, take V35A dataset as example, we divide the 6,432 sample dataset into a real training set size of $4,000$ and a testing set size of $2,432$. The ratios of the positive versus negative samples in both training and test sets are $1:1$. With the classification model and the dataset selected, we proceed the test on each of the generative models as in the following four steps:

\begin{enumerate}
\item Train the generative model with $4,000$ sample real training set.
\item Based on the trained generative model, produce additional $4,000$ synthetic samples, which become the synthetic training set.
\item Train the adaptive filtering classifier with the synthetic training set.
\item Test the trained classifier on the test set and report the accuracy.
\end{enumerate}

Intuitively, the performance of the classifier using real training set will be better than the one trained on synthetic set. Hence, we provide the performance of the classifier~(denoted as $C_{R}$) trained on real waveform data for comparison purposes. Similarly, we denote the classifiers trained on synthetic data as $C_{S}$. The higher the classification metric of $C_{S}$, the better the quality of the synthetic samples that are used to train the classifier. We provide the classification results in Table ~\ref{tab:num eval}. 

As expected, $C_R$ yields the best performance among all classifiers. The classifier $C_{S0}$ based on our generative model produces the second-best classification accuracy, with classifiers trained on synthetics from baseline models 1 through 7 lagging well behind. This is consistent with our visual evaluation results reported in Section \ref{sec:baseline design}. Through this test, we verify that our generative model can effectively learn the key features from real seismic time series so that its synthetic samples may be as helpful as the real data for the classification task. It is interesting to notice however that there can be some inconsistency in between visual evaluation and classification accuracy. As an example, The results of baseline~2  as shown in Figure~\ref{fig:indeppipe} can be easily identified as unrealistic samples by human experts. However, when these synthetic samples are used to train a classifier, we obtain accuracy as high as $95.02\%$ and $95.35\%$ based on our two dataset. This inconsistency is due to the fact that adaptive filtering classifier favors local features while human are capable of capturing both local and global features.

\begin{table*}[h!]
\centering
 \begin{tabular}{| c || c || c | c | c | c | c | c | c | c |} 
 \hline
Classifier & $C_{R}$ & $C_{S0}$       & $C_{S1}$ & $C_{S2}$ & $C_{S3}$ & $C_{S4}$ & $C_{S5}$ & $C_{S6}$ & $C_{S7}$ \\
\hline
V34A       & 98.56   & \textbf{97.11} &  49.58   & 95.02    & 68.09    & 90.51    & 95.95    & 94.60    & 94.27    \\
\hline
V35A       & 98.48   & \textbf{96.96} & 40.81    & 95.35    & 52.06    & 53.99    & 91.60    & 95.04    & 93.45    \\ 
\hline
\end{tabular}
\caption{Classification results using classifier trained on real training set ($C_{R}$ in Col.~2) and those trained on synthetic training set ($C_{S}$ in Cols.~3 to 10).  Specifically, $C_{S0}$ is based on our model, and $C_{S1} \sim C_{S7}$ are based on baseline~1 to 7, respectively.}
\label{tab:num eval}
\end{table*}

\subsection{Test 3: Robustness of Our Generative Model }

Our generative model is trained on labeled datasets. Because in practice it may be difficult to obtain high quality labels, it is worthwhile to study the robustness of our generative model when the size of the training set is limited. To do this, we design our test to train on data set with sizes varying among $10$, $20$, $40$, $60$, and $80$. We keep the ratio between positive and negative samples to be one in all those limited training sets. Take the training set size of $10$ as an example, we randomly select $5$ positive and $5$ negative real training samples from a seismic station (here, V35A), and combine them as the limited training set size of $10$. We construct four other training sets sizes of $20$, $40$, $60$, and $80$ in a similar approach. With those five training sets being available, we train and obtain five different generative models, namely, $G_{10}$, $G_{20}$, $G_{40}$, $G_{60}$, and $G_{80}$. Using each generative model, we then synthesize a training set size of $4,000$ that consists of  $2,000$ positive and $2,000$ negative synthetic samples. Based on those five synthetic training sets of $4,000$, we independently train five adaptive filtering classifiers and test each of them on the same V35A test set as those used in Test 2. We record the accuracy, precision, and recall of the predictions from each of the five classifiers (Cols.~2 to 6 in Table~\ref{tab:generator eval 35 acc}, \ref{tab:generator eval 35 prec} and \ref{tab:generator eval 35 rec}). As a benchmark, a classifier trained on the real training set is also reported (denoted as ``real'' in Col.~1 of Table~\ref{tab:generator eval 35 acc}, \ref{tab:generator eval 35 prec} and \ref{tab:generator eval 35 rec}) and we use all $4,000$ real samples as the training set.

Not surprisingly, the classifier trained with large amounts of real data the classifier yields the best performance~(Col.~1 in Table~\ref{tab:generator eval 35 acc}). While using synthetic samples only~(Cols.~2 $\sim$ 7 in Table~\ref{tab:generator eval 35 acc}), the classifiers still produce reasonable predictions with accuracy higher than 75\%, and exceeding 92\% when the training dataset is 80. This indicates the robustness of our generative model with respect to limited training set sizes, which can be further explained using the results of precision and recall. Specifically, as shown in Table~\ref{tab:generator eval 35 prec}, all six classifiers achieve similarly high precision values, which are no less than 96\%. In contrast, we observe that as the training set is augmented from size $10$ to $80$, the recall value of the classifier prediction is rapidly increased from  $52.6\%$ to around $82.0\%$. 

The high precision value of classifier from $G_{10}$ indicates a low number of false positive cases, meant that even with only $10$ training samples, the classifier trained by the model-generated synthetic samples can still recognize most of the negative samples. However, its recall value shows that such classifier mislabels almost half of the positive samples as negative. By increasing the number of training samples from $10$ to $80$, the classifier improves its recall value from 52.63\% to 87.44\% while keeping its high precision value almost unchanged. This shows that when the training set of the generative model is augmented, the classifier is able to recognize more and more positive samples, thus increasing the overall accuracy. We implement a similar robustness test on V34A dataset and report the results in Table~\ref{tab:generator eval 34 acc}, \ref{tab:generator eval 34 prec} and \ref{tab:generator eval 34 rec}. Similar conclusions can be drawn.

In summary, through this test we learn that our generative model can be effective when training set is limited.  This is consistent with the image synthesis task~\cite{gurumurthy2017deligan, marchesi2017megapixel}, where GAN has been proven to be effective on limited datasets. %Despite of the reasonable classification results are obtained in Tables~\ref{tab:generator eval 35} and \ref{tab:generator eval 34}, we will explore the possibility to of further improving accuracy through data augmentation. 

\begin{table}[h!]
\centering
 \begin{tabular}{| c || c | c | c | c | c |} 
 \hline
real  & $G_{10}$ & $G_{20}$ & $G_{40}$ & $G_{60}$ & $G_{80}$ \\
\hline
98.48 & 75.99    & 89.38    & 89.16    & 90.35    & 92.79    \\ 
 \hline
\end{tabular}
\caption{Accuracy of the robustness test on V35A dataset. We provide benchmark accuracy (Col.~1) as well as those results using our generative models based on five different limited training sets (Cols.~2 to 6). Our model yields reasonable robustness with limited training size. }
\label{tab:generator eval 35 acc}
\end{table}

\begin{table}[h!]
\centering
 \begin{tabular}{| c || c | c | c | c | c |} 
 \hline
real  & $G_{10}$ & $G_{20}$ & $G_{40}$ & $G_{60}$ & $G_{80}$ \\
\hline
98.36 & 98.78    & 96.31    & 97.59    & 99.14    & 97.92    \\ 
 \hline
\end{tabular}
\caption{Precision of the robustness test result on V35A dataset.}
\label{tab:generator eval 35 prec}
\end{table}

\begin{table}[h!]
\centering
 \begin{tabular}{| c || c | c | c | c | c |} 
 \hline
real  & $G_{10}$ & $G_{20}$ & $G_{40}$ & $G_{60}$ & $G_{80}$ \\
\hline
98.60 & 52.63    & 82.02    & 80.40    & 81.41    & 87.44    \\ 
 \hline
\end{tabular}
\caption{Recall of the robustness test result on V35A dataset.}
\label{tab:generator eval 35 rec}
\end{table}

\begin{table}[h!]
\centering
 \begin{tabular}{| c || c | c | c | c | c |} 
 \hline
real  & $G_{10}$ & $G_{20}$ & $G_{40}$ & $G_{60}$ & $G_{80}$ \\
\hline
98.56 & 66.59    & 83.43    & 93.04    & 92.64    & 93.75    \\ 
 \hline
\end{tabular}
\caption{Accuracy of the robustness test result on V34A dataset. Similar with result from V35A dataset, our model yields reasonable robustness with limited training size.}
\label{tab:generator eval 34 acc}
\end{table}

\begin{table}[h!]
\centering
 \begin{tabular}{| c || c | c | c | c | c |} 
 \hline
real  & $G_{10}$ & $G_{20}$ & $G_{40}$ & $G_{60}$ & $G_{80}$ \\
\hline
98.24 & 94.41    & 93.16    & 99.18    & 99.21    & 97.97    \\ 
 \hline
\end{tabular}
\caption{Precision of the robustness test result on V34A dataset.}
\label{tab:generator eval 34 prec}
\end{table}

\begin{table}[h!]
\centering
 \begin{tabular}{| c || c | c | c | c | c |} 
 \hline
real  & $G_{10}$ & $G_{20}$ & $G_{40}$ & $G_{60}$ & $G_{80}$ \\
\hline
98.88 & 65.86    & 82.65    & 94.65    & 93.07    & 92.33    \\ 
 \hline
\end{tabular}
\caption{Recall of the robustness test result on V34A dataset.}
\label{tab:generator eval 34 rec}
\end{table}

\subsection{Test 4: Data Augmentation using Our Model}

Data augmentation is a commonly-used technique in machine learning to expand the amount of data available for training. Such a technique can be valuable for earthquake detection tasks using machine learning due to the difficulty in obtaining high-quality, labeled waveform data. However, traditional data augmentation techniques such as cropping, padding, or flipping are limited in their effectiveness because they do little to expand the actual diversity of waveform characteristics necessary to train such models. Through our previous tests, we demonstrate that our generative model is capable of synthesizing realistic positive and negative seismic samples. In this test, we further utilize those synthetic samples to augment the training set of real waveforms and evaluate classification performance.  

We design this test based on the same five generative models ($G_{10}$, $G_{20}$, $G_{40}$, $G_{60}$, and $G_{80}$) and their related real training sets of limited sizes ($10$, $20$, $40$, $60$, and $80$) from Test~3. We use those generative models to produce different numbers of the synthetic samples that will be combined with the existing real training set. For the ease of demonstration, we use a variable $r$ to stand for the augmentation ratio of the synthetic samples to be added to the initial, real sample. We choose six different augmentation scenarios including $r_{1} = 1:1$, $r_{10} = 10:1$, $r_{50} = 50:1$, $r_{100} = 100:1$, $r_{200} = 200:1$ and $r_{300} = 300:1$.  Take $G_{10}$ and $r_{50} = 50:1$ as an example, we begin with a training dataset of 10 real samples, 5 positive and 5 negative. We then generate $50 \times 10 = 500$ synthetic samples (250 positive and 250 negative) and combine them with the existing limited real training set size of 10 to give an augmented training set size of 510. We then train an adaptive filtering classifier using this augmented training dataset. We report our classification accuracy, precision and recall using the V35A test set in Tables~\ref{tab:main result 35 acc}, \ref{tab:main result 35 prec} and \ref{tab:main result 35 rec} respectively. As a baseline, we also include a scenario of non-augmentation test, where the classifier is trained only on real data set and we denote this as $r_{0}$. Following \citeA{zhang2019adaptive}, we use the learning rate $1\times10^{-4}$ for training classifiers. To make a fair comparison, we train each classifier for a total of $1,500$ iterations. 

We observe from Table~\ref{tab:main result 35 acc} that baseline ($r_{0}$) typically yields worse classification accuracy compared to the augmentation scenarios.  For 23 out of 30 cases (bold in Table~\ref{tab:main result 35 acc}), the augmentation of dataset shows improvement on the performance of the classifier. In the best case, the accuracy is increased by over 14\%~(Col.~3 in Table~\ref{tab:main result 35 acc}). Such an improvement can be explained by the improvement of both precision and recall results, which can be observed in Tables~\ref{tab:main result 35 prec} and \ref{tab:main result 35 rec}. The rare counterexamples where accuracy does not increase occur in small sample-size regimes ($r_1$ and $G_{10}$) where sample-to-sample variability becomes important.

We proceed similar tests on the V34A test set and report the results in Tables~\ref{tab:main result 34 acc}, \ref{tab:main result 34 prec} and \ref{tab:main result 34 rec}. Similar performance improvement can be observed, where 27 out 30 cases result in improvement and the largest increase in classification accuracy is over 17\% (Col.~3 in Table~\ref{tab:main result 34 acc}). In Tables~\ref{tab:main result 34 prec} and \ref{tab:main result 34 rec}, we also observe a similar improvement of both precision and recall values as those in V35A test set. Through this test, we conclude that the synthetic samples generated by our generative model can improve the performance of the classifier by data augmentation. 

\begin{table}[h!]
\centering
 \begin{tabular}{|c || c | c | c | c | c | c |} 
 \hline
           &       $G_{10}$ &       $G_{20}$ &       $G_{40}$ &       $G_{60}$ &       $G_{80}$ \\
 \hline
 $r_{0}$   &         71.15  &         76.73  &         84.41  &         89.36  &         94.38  \\ 
 \hline
 $r_{1}$   &         65.63  &         70.52  &         79.56  &         86.05  &         92.96  \\ 
 \hline
 $r_{10}$  &         67.32  &         70.74  & \textbf{92.15} & \textbf{92.64} & \textbf{96.03} \\ 
 \hline
 $r_{50}$  & \textbf{73.35} & \textbf{87.36} & \textbf{92.42} & \textbf{93.04} & \textbf{95.56} \\ 
 \hline
 $r_{100}$ & \textbf{78.27} & \textbf{89.10} & \textbf{92.12} & \textbf{92.54} & \textbf{95.77} \\ 
 \hline
 $r_{200}$ & \textbf{79.54} & \textbf{90.74} & \textbf{91.65} & \textbf{92.98} & \textbf{95.31} \\ 
 \hline
 $r_{300}$ & \textbf{80.72} & \textbf{90.80} & \textbf{91.41} & \textbf{92.80} & \textbf{95.37} \\ 
 \hline
\end{tabular}
\caption{Detection accuracy using classifiers trained on augmented training set from V35A dataset. Entries marked in bold provide improved performance over the baseline $r_0$ (first row) where no data augmentation is performed. }
\label{tab:main result 35 acc}
\end{table}

\begin{table}[h!]
\centering
 \begin{tabular}{|c || c | c | c | c | c | c |} 
 \hline
           &       $G_{10}$ &       $G_{20}$ &       $G_{40}$ &       $G_{60}$ &       $G_{80}$ \\
 \hline
 $r_{0}$   &         74.97  &         81.71  &         90.00  &         93.35  &         94.47  \\ 
 \hline
 $r_{1}$   & \textbf{84.22} & \textbf{83.81} & \textbf{92.16} &         92.96  & \textbf{96.71} \\ 
 \hline
 $r_{10}$  & \textbf{90.07} & \textbf{92.71} & \textbf{97.93} & \textbf{97.62} & \textbf{97.97} \\ 
 \hline
 $r_{50}$  & \textbf{96.38} & \textbf{97.32} & \textbf{97.72} & \textbf{97.25} & \textbf{97.59} \\ 
 \hline
 $r_{100}$ & \textbf{97.06} & \textbf{97.45} & \textbf{97.54} & \textbf{96.69} & \textbf{97.53} \\ 
 \hline
 $r_{200}$ & \textbf{97.27} & \textbf{96.99} & \textbf{97.22} & \textbf{96.67} & \textbf{97.03} \\ 
 \hline
 $r_{300}$ & \textbf{97.25} & \textbf{97.32} & \textbf{97.55} & \textbf{96.45} & \textbf{96.61} \\ 
 \hline
\end{tabular}
\caption{Precision values using classifiers trained on augmented training set from V35A dataset.}
\label{tab:main result 35 prec}
\end{table}

\begin{table}[h!]
\centering
 \begin{tabular}{|c || c | c | c | c | c | c |} 
 \hline
           &       $G_{10}$ &       $G_{20}$ &       $G_{40}$ &       $G_{60}$ &       $G_{80}$ \\
 \hline
 $r_{0}$   &         63.56  &         69.19  &         77.18  &         84.65  &         94.29  \\ 
 \hline
 $r_{1}$   &         38.62  &         51.23  &         64.64  &         77.98  &         88.95  \\ 
 \hline
 $r_{10}$  &         38.95  &         45.01  & \textbf{86.12} & \textbf{87.43} &         94.01  \\ 
 \hline
 $r_{50}$  &         48.52  & \textbf{76.85} & \textbf{86.88} & \textbf{88.59} &         93.43  \\ 
 \hline
 $r_{100}$ &         58.33  & \textbf{80.30} & \textbf{86.42} & \textbf{88.10} &         93.92  \\ 
 \hline
 $r_{200}$ &         60.84  & \textbf{84.08} & \textbf{85.76} & \textbf{89.05} &         93.49  \\ 
 \hline
 $r_{300}$ &         63.26  & \textbf{83.90} & \textbf{84.97} & \textbf{88.87} &         94.07  \\ 
 \hline
\end{tabular}
\caption{Recall values using classifiers trained on augmented training set from V35A dataset.}
\label{tab:main result 35 rec}
\end{table}

\begin{table}[h!]
\centering
 \begin{tabular}{|c || c | c | c | c | c |} 
 \hline
           &       $G_{10}$ &       $G_{20}$ &       $G_{40}$ &       $G_{60}$ &       $G_{80}$ \\
 \hline
 $r_{0}$   &         64.18  &         73.53  &         90.12 &          91.69  &         93.59  \\ 
 \hline
 $r_{1}$   &         61.94  &         68.49  & \textbf{91.24} &         88.64  & \textbf{94.87} \\ 
 \hline
 $r_{10}$  & \textbf{66.47} & \textbf{81.72} & \textbf{95.90} & \textbf{95.06} & \textbf{96.21} \\ 
 \hline
 $r_{50}$  & \textbf{73.91} & \textbf{90.51} & \textbf{96.57} & \textbf{95.86} & \textbf{96.12} \\ 
 \hline
 $r_{100}$ & \textbf{76.49} & \textbf{90.00} & \textbf{96.77} & \textbf{95.69} & \textbf{96.18} \\ 
 \hline
 $r_{200}$ & \textbf{78.27} & \textbf{90.21} & \textbf{96.72} & \textbf{95.58} & \textbf{95.94} \\ 
 \hline
 $r_{300}$ & \textbf{78.22} & \textbf{90.84} & \textbf{96.40} & \textbf{95.63} & \textbf{96.11} \\ 
 \hline
\end{tabular}
\caption{Detection accuracy using classifiers trained on augmented training set from V34A dataset. Entries marked in bold provide improved performance over the baseline $r_0$ (first row) where no data augmentation is performed. }
\label{tab:main result 34 acc}
\end{table}

\begin{table}[h!]
\centering
 \begin{tabular}{|c || c | c | c | c | c |} 
 \hline
           &       $G_{10}$ &       $G_{20}$ &       $G_{40}$ &       $G_{60}$ &       $G_{80}$ \\
 \hline
 $r_{0}$   &         67.58  &         75.42  &         91.63  &         92.94  &         94.15  \\ 
 \hline
 $r_{1}$   &         71.71  &         72.87  & \textbf{95.16} &         93.01  & \textbf{97.12} \\ 
 \hline
 $r_{10}$  &         88.90  & \textbf{85.87} & \textbf{97.79} & \textbf{98.27} & \textbf{97.79} \\ 
 \hline
 $r_{50}$  &         95.29  & \textbf{95.50} & \textbf{98.25} & \textbf{98.72} & \textbf{97.79} \\ 
 \hline
 $r_{100}$ & \textbf{95.67} & \textbf{96.32} & \textbf{98.45} & \textbf{98.58} & \textbf{97.86} \\ 
 \hline
 $r_{200}$ & \textbf{95.16} & \textbf{96.72} & \textbf{98.45} & \textbf{98.54} & \textbf{97.88} \\ 
 \hline
 $r_{300}$ & \textbf{95.39} & \textbf{96.62} & \textbf{98.31} & \textbf{98.44} & \textbf{97.89} \\ 
 \hline
\end{tabular}
\caption{Precision values using classifiers trained on augmented training set from V34A dataset.}
\label{tab:main result 34 prec}
\end{table}

\begin{table}[h!]
\centering
 \begin{tabular}{|c || c | c | c | c | c |} 
 \hline
           &       $G_{10}$ &       $G_{20}$ &       $G_{40}$ &       $G_{60}$ &       $G_{80}$ \\
 \hline
 $r_{0}$   &         55.29  &         69.82  &         88.36  &         90.28  &         92.89  \\ 
 \hline
 $r_{1}$   &         43.79  &         58.88  & \textbf{86.90} &         83.58  &         92.49  \\ 
 \hline
 $r_{10}$  &         37.09  & \textbf{75.83} & \textbf{93.93} & \textbf{91.72} & \textbf{94.55} \\ 
 \hline
 $r_{50}$  &         50.23  & \textbf{85.02} & \textbf{94.83} & \textbf{92.93} & \textbf{94.38} \\ 
 \hline
 $r_{100}$ & \textbf{55.39} & \textbf{83.19} & \textbf{95.02} & \textbf{92.73} & \textbf{94.42} \\ 
 \hline
 $r_{200}$ & \textbf{59.48} & \textbf{83.25} & \textbf{94.92} & \textbf{92.53} & \textbf{93.91} \\ 
 \hline
 $r_{300}$ & \textbf{59.22} & \textbf{84.63} & \textbf{94.43} & \textbf{92.73} & \textbf{94.25} \\ 
 \hline
\end{tabular}
\caption{Recall values using classifiers trained on augmented training set from V34A dataset.}
\label{tab:main result 34 rec}
\end{table}

\section{Discussion and Future Work}

In this work, we have demonstrated how a machine learning approach based on the conditional Generative Adversarial Network (GAN) can be used to generate realistic seismic waveforms that sample either earthquake or non-earthquake classes. A generative model of this type may have multiple use cases in seismology. The focus of this paper is on data augmentation, where we have shown that synthetic waveforms can be used to expand the amount of available training data and thereby improve the classification accuracy of machine learning algorithms when applied to real datasets. A potential related use case would be the application of synthetics of this type to test the robustness of detection algorithms. A particularly salient example would be in the field of earthquake early warning, where distinguishing between earthquake and non-earthquake events is of fundamental importance \cite{meier_reliable_2019}. In other instances, having a means to generate both earthquake and non-earthquake records, and combine them in superposition, may help test the sensitivity of seismic methodology to varying levels of signal-to-noise ratio.

The techniques we outline in this manuscript do have limitations that are important to be aware of. Perhaps the most obvious is that the model is constrained based on training records from only two stations, both in Oklahoma. Because of this, the model learns what earthquake and non-earthquake waveforms tend to look like at these stations, and is capable of reproducing these basic features in its generative model. However, the model is unlikely to generalize (without additional training) to other stations, where both the noise characteristics of the non-earthquake seismic record and the earthquake arrivals may differ. And while the channel-to-channel temporal correlations in the synthetics are realistic, the model has no understanding of the expected moveout of waveforms across a seismic network that are crucial in many seismic applications. Thus, the model we present here should be view more as a proof of principle that the methodology is promising, rather than a finished machine learning product ready for widespread deployment.

Moreover, our approach is fundamentally data-driven. We train our model on data, which we posit is sufficient to learn the details of the task at hand. However, real earthquakes, and the seismic waves they broadcast, obey the physical constraints of the governing equations and constitutive laws of dynamic rupture and seismic wave propagation. Incorporating aspects of the known, underlying physical theory in the form of hybrid, physics-informed machine learning models is an active area of research. We hope that in future work, we can improve our generative modeling framework by adopting a more holistic, physics-informed approach.

\section{Conclusions}

We develop a generative model that can produce realistic, synthetic seismic waveforms of either earthquake or non-earthquake (noise) classes. Our machine learning model is in essence a conditional generative adversarial network designed to operate on three-component waveforms at a single seismic station. To verify the efficacy of our generative model, we apply it to seismic field data collected at Oklahoma. Through a sequence of qualitative and quantitative tests and benchmarks, we show that our model can generate high-quality synthetic waveforms. We further demonstrate that performance of machine learning based detection algorithms can be improved by using augmented training sets with both synthetic and real samples. Our generative model has several potential use cases across seismology, but our focus in this work is on the earthquake detection problem.

\section*{Acknowledgments, Samples, and Data}

The authors declare no conflicts of interest. This work was supported by the Center for Space and Earth Science (CSES) at Los Alamos National Laboratory (LANL). The experiment was performed using supercomputers of LANL's Institutional Computing Program. We also would like to acknowledge Dr.~Jake Walter from Oklahoma Geological Survey and University of Oklahoma for providing revised catalog. Thanks to USGS for providing the raw seismic data from Transportable Array (network code TA). Both the seismic datasets collected at Stations V34A and V35A used in this work were downloadable from the openly accessible Data management Center managed by IRIS~(\url{http://ds.iris.edu/ds/nodes/dmc/}). For all the training sets that are used for our model can be downloaded from the Gitlab repo~(\url{https://gitlab.com/huss8899/seismogramgen}). All the results generated using our generative model as well as all the baseline models have also been shared in the report for any potential interests from readers.

\bibliography{reference}

\end{document}